\documentclass[11pt]{article}

\usepackage{amsmath}

\usepackage{graphicx}
\usepackage{amssymb,amsmath}
\usepackage{diagbox}
\usepackage{enumerate}
\usepackage{bm}

\usepackage{xcolor}
\usepackage{amsmath}
\usepackage{amssymb}
\usepackage{latexsym}
\usepackage{graphicx}
\usepackage{graphics}
\usepackage{epstopdf}
\usepackage{subfigure}
\usepackage{appendix}
\usepackage{color}
\usepackage{ulem}
\usepackage{cancel}
\usepackage{diagbox}
\usepackage{textcomp,booktabs}
\usepackage{colortbl}
\usepackage{rotating}
\usepackage{bm}
\usepackage{siunitx}
\usepackage{multirow}
\usepackage{url}
\usepackage{hyperref}
\usepackage{doi}
\usepackage[numbers]{natbib}

\oddsidemargin
-0.2in \topmargin -.60in
\begin{document}

\title{Real-time Forecast Models for TBM Load Parameters Based on Machine Learning Methods}

\author{
Xianjie~Gao$^{1}$,\quad Xueguan~Song$^{2}$,\quad Maolin~Shi$^{2}$,\quad Chao~Zhang$^{3}\footnote{Corresponding author}$,\quad Hongwei~Zhang$^{3}$\\
$^1$Department of Basic Sciences,\\
Shanxi Agricultural University, Taigu, Shanxi 030801, P.R. China\\
\texttt{xjgao@sxau.edu.cn}\\
$^2$School of Mechanical Engineering,\\
Dalian University of Technology, Dalian, Liaoning, 116024, P.R. China\\
\texttt{sxg@dlut.edu.cn; shl5985336@126.com} \\
$^{3}$School of Mathematical Sciences,\\
Dalian University of Technology, Dalian, Liaoning, 116024, P.R. China\\
\texttt{chao.zhang@dlut.edu.cn; hwzhang@dlut.edu.cn}
%
%
}
\date{}



\maketitle

\begin{abstract}%

Because of the fast advance rate and the improved personnel safety, tunnel boring machines (TBMs) have been widely used in a variety of tunnel construction projects. The dynamic modeling of TBM load parameters (including torque, advance rate and thrust) plays an essential part in the design, safe operation and fault prognostics of this complex engineering system. In this paper, based on in-situ TBM operational data, we use the machine-learning (ML) methods to build the real-time forecast models for TBM load parameters, which can instantaneously provide the future values of the TBM load parameters as long as the current data are collected. To decrease the model complexity and improve the generalization, we also apply the least absolute shrinkage and selection (Lasso) method to extract the essential features of the forecast task. The experimental results show that the forecast models based on deep-learning methods, {\it e.g.}, recurrent neural network and its variants, outperform the ones based on the shallow-learning methods, {\it e.g.}, support vector regression and random forest. Moreover, the Lasso-based feature extraction significantly improves the performance of the resultant models.

\end{abstract}


\section{Introduction}

With the increasing demand of long and deep tunnel construction, the traditional tunneling techniques ({\it e.g.,} drilling and blasting) have not satisfied the construction requirement of high efficiency, small environmental damage and low construction risk. Instead, tunnel boring machines (TBMs) have become a primary method in various underground excavation projects ({\it e.g.,} mountain tunnel \cite{miura2003design}, city subway \cite{edalat2010choosing} and coal mine \cite{cigla2001application}) with its predominance of high advance rate, environmental friendliness and security. The widespread use of TBMs benefits from the experiences gained from different conditions and the new technological advances. However, there are still many challenges in the machinery design and the project application of TBMs. For example, under complicated construction conditions, TBMs' excavation process will be faced with high risk caused by an unanticipated presence of water leakage, rock spalling, geological faulting, weak regions, break outs of the tunnel wall and subsidence of work area. To improve TBMs' reliability and reduce unscheduled downtime, some research works have paid attention to TBMs' risk evaluation \cite{hong2009quantitative}, design and R$\&$D \cite{cho2014design}, and safety decision \cite{wu2015dynamic}.

One main research concern of TBM is to develop a health monitoring and prognostics system that can automatically find out the anomalies and the provides the fault prognostic during the TBM operation \cite{kothamasu2006system}. An essential part of such a system is the real-time forecast of TBM load parameters (including torque, advance rate and thrust), which are important indicators to evaluate TBMs' machine health condition by observing whether their forecast values exceed the pre-defined threshold or not \cite{wang2015disc,talebi2015modeling}. However, due to the complex geological condition, it is still challenging to achieve the accurate real-time forecast of TBM load parameters
({\it cf.} \cite{javad2010application,delisio2013analysis,gong2009development,ates2014estimating,wang2019sensitivity,8850794,zhao2019data}).

In general, the existing methods for forecasting TBM load parameters can be categorized into three kinds:
\begin{enumerate}[(i)]
\item Empirical methods aim to developing the empirical equations, which describe the relationship between the geological information and some important operation parameters, {\it e.g.,} cutter force, thrust and cutter power, to forecast TBM load parameters \cite{krause1976geologische,yagiz2017new,hamidi2010performance,hassanpour2010tbm}. In order to improve the forecast accuracy, there have also been some attempts to combine the empirical methods with the cutting experiments ({\it cf.} \cite{gertsch2007disc,entacher2014tunnel,xue2009soft}).

\item Under the assumption that one stratum or different stratums are distributed on the excavation face, rock-soil mechanics methods explore the relationship between the rock-soil mechanics characteristics of the stratums and TBMs' operation state and then provide the estimate of TBM load parameters \cite{shi2011determination,wang2012modeling,zhang2016modeling}.

\item The numerical simulation methods employ some differential equation models to describe the dynamic variation of TBMs' operation parameters, and then provide the estimate of these parameters with the aid of computer simulation, {\it e.g.,} the three-dimensional finite element simulation model \cite{kasper2006influence} and the 3D FEM model \cite{su2011analysis}.
\end{enumerate}
However, the practical applications of these methods have the following limitations: 1) we need the considerable specialty knowledge or the participation of relevant professional technicians; 2) the forecast results are lack of generality and cannot be (at least cannot be directly) generalized to other construction conditions; and 3) these methods are unsuitable (at least cannot be directly applied) to the real-time forecast, which requires that the future forecast results should be instantaneously provided as long as the current operational data are collected.


In this paper, we adopt the machine-learning (ML) methods to develop the real-time forecast models for TBM load parameters based on TBM operational data collected from the sensors equipped on a TBM in the construction process. We mainly concern with two kinds of ML models: the deep-learning models including recurrent neural networks (RNN) and its variants ({\it e.g.} long-short term memory network (LSTM) and gated recurrent unit network (GRU)); and the shallow-learning models including support vector regression (SVR), random forest (RF) and feed-forward neural network (FNN). To capture the temporal information encoded in the original data, the input of these models is a length of time series data containing the previous and the current data, and the output is the estimate of the future values of the TBM load parameters including torque, advance rate and thrust.

Although the existing work \cite{gao2019recurrent} has discussed the prediction for TBM operational parameters, the models given in \cite{gao2019recurrent} can only provide the prediction value of one single operational parameter once a time and thus has a low efficiency in the prediction task of multiple operational parameters. Moreover, the training difficulty of multiple-output models is relatively higher than the single-output models because the former have more undetermined weights. Therefore, we introduce the Lasso method to extract essential features from the original TBM operational data to simplify the model structure and then to improve the training efficiency accordingly. The experimental results support the effectiveness of the ML-based models for the real-time forecast task.

The rest of this paper is organized as follows. In Section \ref{sec:ai}, we give the necessary preliminaries on the ML methods considered in this paper. In Section \ref{3B}, we introduce the ML-based models of the real-time forecast of the TBM load parameters. Section \ref{sec:exp} presents experimental results and the last section concludes the paper. In the appendix, we list some important features appearing in the in-situ TBM operational data.

\section{Preliminaries on ML Methods}\label{sec:ai}

In this section, we give a brief introduction of the main ML methods used in this paper including FNN, RNN, LSTM, GRU and Lasso. We will show that, because of the existence of loop structure, RNN and its variants can maintain the previous input information for the future forecast, and thus are more suitable to the real-time forecast. The details of SVR and RF are referred to \cite{bishop2006pattern}.

\subsection{Feed-forward Neural Networks}

\begin{figure}[htbp]
  \centering
  \includegraphics[width=3in]{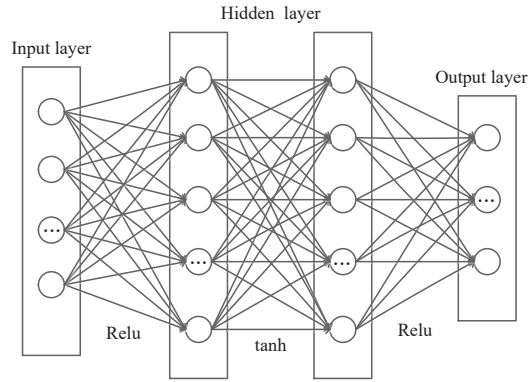}
  \caption{Structure of FNN with Two Hidden Layers}\label{fig:fnn}
\end{figure}

Artificial neural networks (ANNs) are the typical representatives of connectionism models. By using the weighted directed lines to connect neurons squashed by non-linear functions, ANNs can accurately approximate arbitrary continuous functions \cite{cybenko1989approximation}. Because of this strong non-linear mapping capability, ANNs have been widely used in various kinds of AI-related problems \cite{jain1996artificial}. One of the most frequently used ANNs is the feed-forward neural network (FNN) that stacks multiple-layer neurons and the adjacent layers are connected by the unidirectional lines such that the information flow is unidirectional from the input layer to the output layer ({\it cf.} Fig. \ref{fig:fnn}).

Take FNN with one hidden layer as an example. Let ${\bf V} = [{\bf v}_1, {\bf v}_2,\cdots, {\bf v}_L]\in\mathbb{R}^{J\times L}$ be the weight matrix between the output layer and the hidden layer, where ${\bf v}_l$ ($1\leq l\leq L$) is the weight vector connecting the hidden layer with the $l$-th output node. Denote ${\bf W} = [{\bf w}_1, {\bf w}_2,\cdots, {\bf w}_J]\in\mathbb{R}^{I\times J}$ as the weight matrix between the input layer and the hidden layer, where ${\bf w}_j$ ($1\leq j\leq J$) is the weight vector connecting the input layer with the $j$-th hidden node.
Let $\sigma : \mathbb{R} \rightarrow \mathbb{R}$ be a sigmoid function. Given an $I$-dimensional input ${\bf x}=(x^{(1)},\cdots,x^{(I)})^T\in\mathbb{R}^I$, the output ${\bf y}\in\mathbb{R}^L$ of the FNN can be computed by:
\begin{equation}\label{eq:ann}
{\bf y} = \bm{\sigma} ({\bf V}'{\bf h}) =  \left(
  \begin{array}{c}
         \sigma (\langle {\bf v}_1, {\bf h})\rangle \\
          \sigma (\langle{\bf v}_2, {\bf h}) \rangle\\
          \vdots\\
          \sigma (\langle{\bf v}_L, {\bf h})\rangle
 \end{array}
 \right)
= \left(
  \begin{array}{c}
         \sigma \big( \langle{\bf v}_1, \bm{\sigma} ({\bf W}'{\bf x}\rangle\big) \\
          \sigma \big(\langle{\bf v}_2 , \bm{\sigma} ({\bf W}'{\bf x})\rangle\big) \\
          \vdots\\
          \sigma \big(\langle {\bf v}_L , \bm{\sigma} ({\bf W}'{\bf x})\rangle\big)
 \end{array}
 \right),
\end{equation}
where $\bm{\sigma}(\cdot)$ is the vector-valued function that is component-wise w.r.t. the scalar-valued function $\sigma(\cdot)$. One popular FNN training algorithm is the stochastic gradient descent method which updates the weights ${\bf W}$ and ${\bf V}$ in the following way: given a sample set $\{({\bf x}_t, {\bf y}_t)  \}_{t=1}^T$, randomly select $t$ from $\{1,2,\cdots,T\}$ and then
\begin{align*}
{\bf W}_{k+1} =& {\bf W}_k - \eta\frac{\partial \mathcal{L}_t^k({\bf V}_k,{\bf W}_k)}{\partial {\bf h}_t^k(\bf W)} \frac{\partial {\bf h}_t^k(\bf W)}{\partial {\bf W}_k};\nonumber\\
{\bf V}_{k+1} =& {\bf V}_k - \eta\frac{\partial \mathcal{L}_t^k({\bf V}_k,{\bf W}_k)}{\partial {\bf h}_t^k(\bf W)}
\end{align*}
with $\mathcal{L}_t^k ({\bf V},{\bf W})=\|{\bf y}_{t}^k-\widehat{{\bf y}}_{t}^k\|_2^{2}$, where $\eta>0$ is a learning rate and $\widehat{{\bf y}}_{t}^k$ is the network output w.r.t. the input ${\bf x}_t$ at the $k$-th iteration.

\subsection{Recurrent Neural Networks}

\begin{figure}[htbp]
  \centering
  \includegraphics[width=3.5in]{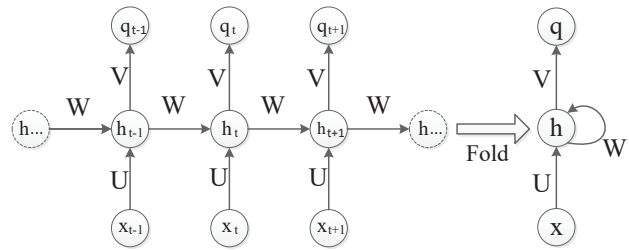}
  \caption{Structure of Recurrent Neural Networks}\label{fig2}
\end{figure}

Recurrent neural networks (RNNs) are a class of neural networks that have the loop structures to maintain the previous learning information and then combine it with the current input to generate the future network outputs \cite{williams1986learning}.
In this manner, the output of a RNN contains the information of the current and the previous inputs. This characteristics makes RNNs quite suitable to dynamically modeling the sequential behavior of time series. As shown in Fig. \ref{fig2}, given a current input ${\bf x}_{t+1}$, the current output of the hidden node ${\bf h}_{t+1}$ is computed by
\begin{equation*}
{\bf h}_{t+1}= {\bf tanh} ({\bf U}{\bf x}_{t+1}+{\bf W}{\bf x}_t),
\end{equation*}
where ${\bf tanh}(\cdot)$ is the vector-valued function that is component-wise w.r.t. the scalar-valued function $\tanh(\cdot)$; ${\bf U}$ is the weight matrix between the current input ${\bf x}_{t+1}$ and the current hidden output ${\bf h}_{t +1}$; and ${\bf W}$ is the weight matrix between the current hidden output ${\bf h}_{t +1}$ and the previous hidden output ${\bf h}_{t}$. Then, the fully-connected node gives the following output:
\begin{equation*}
{\bf q}_{t +1}= {\bf tanh} ({\bf V}{\bf h}_{t+1}),
\end{equation*}
where ${\bf V}$ is the weight matrix between the current hidden output ${\bf h}_{t +1}$ and the current output ${\bf q}_{t+1}$.

RNNs perform well in modeling short sequences, while the training for the long sequences will fail because of vanishing gradients or explosive gradients. The former means that the gradient signal is too small to effectively update weights; and the latter refers to the phenomenon that the magnitude of gradients becomes so large that the training process diverges \cite{pascanu2013difficulty}. To overcome this limitation, some specific gate operations are imposed into RNN's hidden nodes, and then there arise some variants of RNN, {\it e.g.,} long-short term memory network (LSTM) and gated recurrent unit network (GRU).

\subsection{Long-short Term Memory Networks}

In RNNs, the hidden nodes implement the inner-product operation and then are squashed by a nonlinear activation function. For a long time series data, this simple structure cannot efficiently capture the useful sequential characteristic or eliminate the reductant information. Instead of the inner-product operation, the long-short term memory (LSTM) networks impose three kinds of adaptive and multiplicative gate operations into RNN's hidden nodes including the input gate, the forget gate and the output gate \cite{hochreiter1997long}. The input and the output gates respectively control the information flowing in and out; and the forget gate is used to eliminate the redundancy from time series data and meanwhile to maintain the useful information.

\begin{figure}[htbp]
  \centering
  \includegraphics[width=3in]{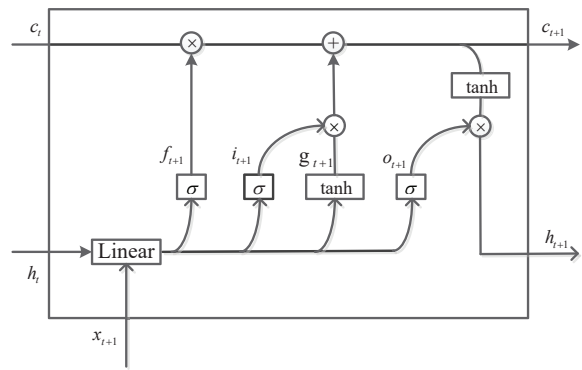}
  \caption{Structure of LSTM's Hidden Node}\label{fig3}
\end{figure}

In Fig. \ref{fig3}, we display a typical structure of LSTMs' hidden nodes, where $i$, $f$ and $o$ stand for the input gate, the forget gate and the output gate respectively. Denote $c$ and $g$ as the memory cell and the new memory cell respectively. Given the previous hidden output ${\bf h}_{t}$ and the current input ${\bf x}_{t+1}$, the forward propagation equations of LSTM are given as follows:
\begin{align*}
{\bf i}_{t+1}  = &\bm{\sigma}({\bf U}^{[i]}{\bf x}_{t+1}+{\bf W}^{[i]}{\bf h}_{t}); \nonumber\\
{\bf f}_{t+1}=&\bm{\sigma}({\bf U}^{[f]}{\bf x}_{t+1}+{\bf W}^{[f]}{\bf h}_t);\nonumber\\
{\bf o}_{t+1}=&\bm{\sigma}({\bf U}^{[o]}{\bf x}_{t+1}+{\bf W}^{[o]}{\bf h}_{t});\\
{\bf g}_{t+1}=&{\bf tanh}({\bf U}^{[g]}{\bf x}_{t+1}+{\bf W}^{[g]}{\bf h}_{t});\nonumber\\
{\bf c}_{t+1}=&{\bf c}_{t}\odot {\bf f}_{t+1}+{\bf g}_{t+1}\odot {\bf i}_{t+1};\nonumber\\
{\bf h}_{t+1}=&{\bf tanh}({\bf c}_{t+1})\odot {\bf o}_{t+1},\nonumber
\end{align*}
where $\odot$ is the Hadamard product, and ${\bf U}^{[\cdot]}$, ${\bf W}^{[\cdot]}$ stand for the weight matrices for the corresponding gates.


\subsection{Gated Recurrent Unit Networks}

\begin{figure}[htbp]
  \centering
  \includegraphics[width=3.5in]{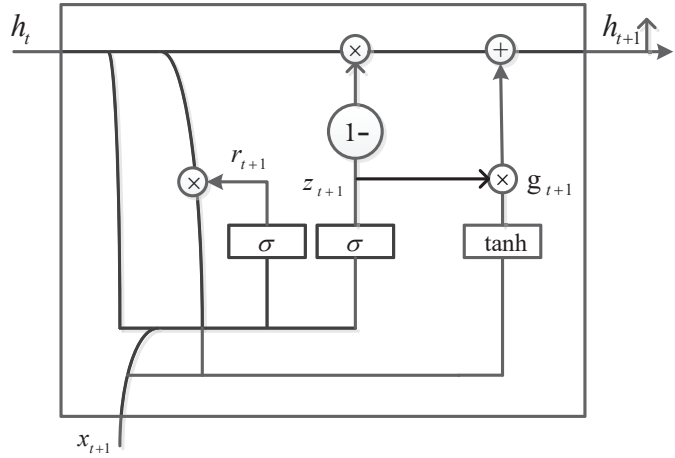}
  \caption{Structure of GRU's Hidden Node}\label{fig4}
\end{figure}

In some problems, the three-gate structure of LSTM's hidden node is likely to cause the overfitting, and thus Cho {\it et al.} \cite{cho2014learning} proposed gated recurrent unit networks (GRUs) , which integrate the input and the forget gates in LSTMs' hidden nodes into one gate ({\it cf.} Fig. \ref{fig4}). Because of the relatively simpler structure, GRUs have a higher training efficiency but also perform better than RNNs and LSTMs in many learning tasks \cite{cho2014learning}. The forward propagation equations of GRU are given as follows:
\begin{align}
{\bf z}_{t+1}= &\bm{\sigma}({\bf U}^{[z]}{\bf x}_{t+1}+{\bf W}^{[z]}{\bf h}_t); \nonumber\\
{\bf r}_{t+1}=&\bm{\sigma}({\bf U}^{[r]}{\bf x}_{t+1}+{\bf W}^{[r]}{\bf h}_t);\\
{\bf g}_{t+1}=&{\bf tanh}\big({\bf U}^{[g]}{\bf x}_{t+1}+{\bf W}^{[g]}({\bf h}_{t}\odot {\bf r}_{t+1})\big);\nonumber\\
{\bf h}_{t+1}=&({\bf 1}-{\bf z}_{t+1})\odot {\bf g}_{t+1}+{\bf z}_{t+1}\odot {\bf h}_{t},\nonumber
\end{align}
where ${\bf x}_{t+1}$, ${\bf h}_{t+1}$, ${\bf z}_{t+1}$ and ${\bf r}_{t+1}$ are the input, the output, the update gate and the reset gate vectors respectively; and ${\bf U}^{[\cdot]}$, ${\bf W}^{[\cdot]}$ stand for the weight matrices for the corresponding gates.

\subsection{Least Absolute Shrinkage and Selection Operator}\label{sec:lasso}

Given a sample set $D=\{({\bf x}_1,y_1),({\bf x}_1,y_1),\ldots,({\bf x}_N,y_N)\}$, where ${\bf x}\in {\mathbb R}^I$, $y\in {\mathbb R}$, consider the following linear regression model:
\begin{equation*}
  \min_{\bm{\beta}} \frac{1}{2} \sum_{n=1}^N \left(  y_i - (\langle {\bf x}_n , \bm{\beta}    \rangle + \beta_0)   \right)^2,
\end{equation*}
where $\langle\cdot,\cdot\rangle$ stands for the inner product, $\bm{\beta}\in\mathbb{R}^I$ is the weight vector and $\beta_0\in\mathbb{R}$ is the bias. This model easily leads to overfitting and has a low tolerance for noisy information. To overcome these shortcomings, one way is to add the regularizer into the objective function:
\begin{equation*}
  \min_{\bm{\beta}} \frac{1}{2} \sum_{n=1}^N \left(  y_i - (\langle {\bf x}_n , \bm{\beta}    \rangle + \beta_0)   \right)^2 + \lambda \Omega(\bm{\beta}),
\end{equation*}
where $\lambda>0$ is the regularization coefficient and $\Omega(\bm{\beta})$ is the regularizer w.r.t. the weight vector $\bm{\beta}$. Especially, when $\Omega(\bm{\beta}) = \|\bm{\beta}  \|_2^2$, the model is called as the ridge regression ({\it cf.} \cite{tikhonov1977solutions}); and when $\Omega(\bm{\beta}) = \|\bm{\beta}  \|_1$, the model is called as the least absolute shrinkage and selection operator (Lasso) ({\it cf.} \cite{tibshirani1996regression}).

Lasso has become one of the most frequently used feature-extraction methods and has been widely used in many engineering problems ({\it e.g.} \cite{lai2019sparse,bartkowiak2014dimensionality}). We note that, one common way to use Lasso method for the feature extraction is to remove the features whose Lasso coefficients are close to {\it zero} and maintain the rest as the input features for the subsequent learning process.




\section{Real-time Forecast of TBM Load Parameters}\label{3B}

In this section, we present the models of the real-time forecast of TBM load parameters. First, we give the problem setup of the real-time forecast.
Then, we discuss two output-modes of the forecast models. Finally, we introduce the Lasso-based feature extraction for TBM operational data.

\subsection{Real-time Forecast}

The real-time forecast requires that models should instantaneously generate the future values of the TBM load parameters during the next period as long as the current operational data are collected. Therefore, a desired forecast model should be able to dynamically capture the dynamic nature of time series data. One reasonable way to handle this issue is to take a time-continuous sequence from the whole time series as the input of the models.

Given a time series $\{({\bf x}_t,{\bf y}_t)\}_{t=1}^\infty$ with ${\bf x}_t =  (x_t^{(1)},\cdots,x_t^{(I)})^T\in\mathbb{R}^I$ and ${\bf y}_t =  (y_t^{(1)},\cdots,y_t^{(J)})^T\in\mathbb{R}^J$, the real-time forecast task aims to find a function $f : \underbrace{\mathbb{R}^I \times \cdots \times \mathbb{R}^I }_\tau\rightarrow \mathbb{R}^J $ such that its output
\begin{equation*}
\widehat{{\bf y}}_{t+1} := f({\bf x}_{t-\tau+1},{\bf x}_{t-\tau+2},\cdots,{\bf x}_{t-1},{\bf x}_{t})
\end{equation*}
can approximate the actual output ${\bf y}_{t+1}$ accurately, where $\tau>0$ is the window width that controls the length of input sequences. In general, the Euclidean distance $\|{\bf y}_t-\widehat{{\bf y}}_t\|_2^2$ is used to measure the difference between ${\bf y}_t$ and $\widehat{{\bf y}}_t$. Especially, if the output vector ${\bf y}_t$ is composed of some specific components of ${\bf x}_t$, {\it i.e.,} ${\bf y}_t := (x_t^{(i_1)},\cdots,x_t^{(i_J)} )^T$ ($J<I$), it is the so-called self-forecast problem.

In practical applications, given a $ T$ length of time series data $\{({\bf x}_t,{\bf y}_t)\}_{t=1}^T$, we can draw $(T- \tau)$ sub-sequence $({\bf x}_{t-\tau+1},{\bf x}_{t-\tau+2},\cdots,{\bf x}_{t-1},{\bf x}_{t})$ from it and the length of each sub-sequence is $\tau$. Then, the function $f$ is obtained by minimizing the mean squared error (MSE)
\begin{equation}\label{eq:mse}
\min_{f\in\mathcal{F}} \frac{1}{ T- \tau}\sum_{t=\tau+1}^{T}\|{\bf y}_{t}-f({\bf x}_{t-\tau+1},{\bf x}_{t-\tau+2},\cdots,{\bf x}_{t-1},{\bf x}_{t})\|_2^{2},
\end{equation}
where $\mathcal{F}$ is a function class that can be set as a shallow-learning model ({\it e.g.} SVR, RF or FNN) or a deep-learning model ({\it e.g.} RNN, LSTM or GRU).

\subsection{Output Modes and Lasso-based Feature Extraction}

As addressed above, TBM's load refer to three kinds of parameters: torque, advance rate and thrust. Thus, forecasting these parameters can be achieved in two kinds of output modes: single-output and multiple-output. In the single-output mode, the aforementioned $f$ is a scalar function that only generates the estimate of one of the three parameters each time. Because of the relatively simpler model structure, the single-output models have a higher training efficiency than that of multiple-output models. However, three individual models should be built for the three TBM parameters respectively. In contrast, one multiple-output model is enough to simultaneously forecast the three parameters, and thus it has a higher operating efficiency.

One way to improve the performance of multiple-output models is to decrease the model complexity by extracting useful features from TBM operational data. As addressed in Section \ref{sec:lasso}, Lasso provides a powerful method for extracting features that are strongly related to learning tasks by observing the magnitude of their Lasso coefficients. In this manner, we select the essential features of the forecast task. The subsequent experimental results show that the Lasso-based feature extraction can significantly improve the performance of the forecast models.

\section{Numerical Experiments}
\label{sec:exp}

In this section, we use the in-situ TBM operational data to conduct the numerical experiments for verifying the validity of the proposed models. All experiments were processed by using PyTorch in a computer with Intel$\circledR$Core$^\text{TM}$ i7 CPU at $4$ GHz, $64$ GB RAM and two NVIDIA GTX 1080 graphics cards.

\subsection{Data Description}

The data set is collected from the sensors equipped on an earth pressure balance shield TBM of the subway tunnel project in Shenzhen, a city of China. The TBM has $500T$ total mass and $120$ knifes on its cutterhead. The tunnel length is about $2000m$ and its diameter is $6.3m$. The elevation of ground surface is between $0.2m\sim5.8m$, and the depth of the tunnel floor from the ground surface is between $11.8m\sim25.4m$. There are $44$ features in the TBM operational data ({\it cf.} Tab. \ref{tab:list}).

We choose a $3000$ length of time-continuous TBM operational data $\{{\bf x}_1,{\bf x}_2,\cdots,{\bf x}_{3000}\}$ and then normalize them in the component-wise way:
\begin{equation}
  x_t^{(i)} =  \frac{x_t^{(i)}-\min\limits_{1\leq k\leq 3000}\big\{x^{(i)}_k\big\}}{\max\limits_{1\leq j, k\leq 3000}\big\{|x^{(i)}_k-{x^{(i)}_j|\big\}}},\quad t=1,2,\cdots,3000,\;\; i=1,2,\cdots,44.
\end{equation}
This sequence is split into two time-continuous sub-sequences: $\{{\bf x}_1,{\bf x}_2,\cdots,{\bf x}_{2500}\}$ and $\{{\bf x}_{2501},{\bf x}_{2502},\cdots,{\bf x}_{3000}\}$. The former is used to train models, and the latter serves as the test set to examine whether the trained models can accurately forecast the load parameters during next period.

Moreover, we consider two error criterions: the root of MSE (RMSE) ({\it cf. }\eqref{eq:mse}) and the mean absolute percentage error (MAPE):
\begin{equation*}
  {\rm MAPE}=\frac{100}{T- \tau}\sum_{t=\tau+1}^T\frac{|\widehat{y}_t^{(j)}-y_t^{(j)}|}{y_t^{(j)}}\%,\quad j = \{1,2,3\},
\end{equation*}
where $\widehat{y}_t^{(j)}$ and $y_t^{(j)}$ stand for the prediction and the actual values of the load parameters ``Torque" ($j=1$), ``Advance Rate" ($j=2$) and ``Thrust" ($j=3$).

\subsection{Experimental Settings}

Here, we consider the real-time forecast of TBM load parameters including torque, thrust and advance rate. Note that the three parameters belong to the $44$ features of TBM operational data, and thus it is a self-forecast problem. Set the window width $\tau=5$, {\it i.e.,} generating the current output needs the previous five inputs.

\begin{figure}[htbp]
  \centering
  \includegraphics[width=3.5in]{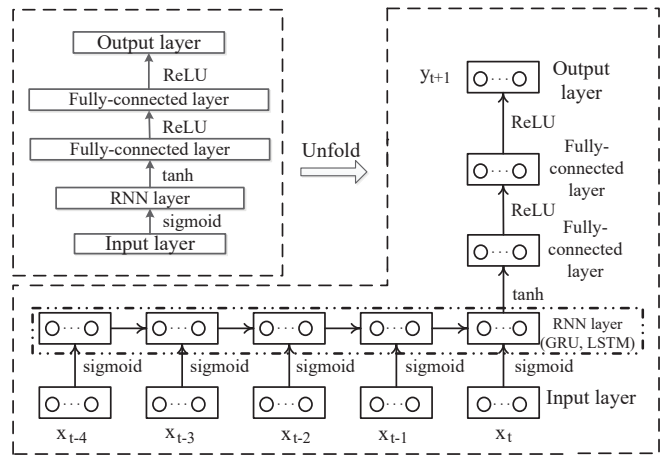}
  \caption{Structure of RNN}\label{fig5}
\end{figure}

The following are the specific settings of the aforementioned models in the experiments:

\begin{itemize}
\item The kernel function of support vector regression (SVR) is selected as ``RBF" with the parameter $\epsilon=0.1$ and the constant $C$ equals to $1$.

\item Random forest (RF) is an ensemble of ten decision trees with the maximal depth being $5$.

\item Feed-forward neural network (FNN) has two hidden layers both of which have $10$ nodes, its input layer contains $220$ nodes and its output layer has one node (resp. three nodes) corresponding to single-output mode (resp. multiple-output mode) ({\it cf.} Fig. \ref{fig:fnn}). The network is trained by using stochastic gradient descent with learning rate $\eta=0.01$ and the maximal training epoch is $100$.

\item The recurrent neural network (RNN) is composed of an input layer with $44$ nodes, one RNN layer with $10$ RNN units, a fully-connected layer composed of two layers both of which have $10$ hidden nodes and an output layer with one or three nodes. The LSTM (resp. GRU) has the same structure as that of RNN except that the RNN units in the RNN layer are replaced with LSTM (resp. GRU) units ({\it cf.} Fig. \ref{fig5}). These models are trained by using RMSprop \cite{tielemandivide} with the learning rate $\eta=0.0005$ and the maximal training epoch is $30000$.
\end{itemize}

The comparative experiments are conducted in the four kinds of settings: single-output mode with (reps. without) Lasso-based feature extraction (briefly denoted as ``s.w.l." (resp. ``s.w/o.l.")) and multiple-output mode with (reps. without) Lasso-based feature extraction (briefly denoted as ``m.w.l." (resp. ``m.w/o.l.")), respectively. In the s.w/o.l. and m.w/o.l. settings, the inputs of these models are the original operation data that have $44$ features. The Lasso-based feature extraction will select the essential features according to their Lasso coefficients whose magnitudes reflects their importance to the forest task. Thus, in the s.w.l. setting, for each one of the three parameters, the input features are extracted from the $44$ features by using the Lasso method.
Combine its own feature and extracted features, we use Lasso-based single-output SVR, RF, FNN amd RNN-based (RNNs, LSTM and GRU) predictors to real-time forecast. We change the number of input nodes according to the number of extracted features. The number of nodes in the input layer of the predicted torque FNN and RNN-based models are $30$ and $6$, respectively. The number of nodes in the input layer of the predicted advance rate FNN and RNN-based models are $35$ and $7$, respectively. The number of nodes in the input layer of the predicted thrust FNN and RNN-based models are $35$ and $7$, respectively. Comprehensive the features to be predicted and all extracted features. We use multi-output SVR, RF, FNN amd RNN-based (RNNs, LSTM and GRU) predictors. We set the number of nodes in the input layer of FNN and RNN-based models to $90$ and $18$, respectively. Note that, in the m.w.l. setting, the input features are the integration of the features extracted by the Lasso method implemented for each of the load parameters.

\subsection{Experimental Results}

To exam the validity of the proposed models, we mainly concern with the following issues:

\begin{itemize}
\item The performance of these models for the real-time forecast task.

\item The performance discrepancy between the single-output mode and the multiple-output mode.

\item The effectiveness of the Lasso-based feature extraction.

\end{itemize}

\begin{figure}[htbp]
\centering
\subfigure[SVR (s.w/o.l.)]{
\begin{minipage}{0.22\textwidth}
\includegraphics[width=1.2\textwidth]{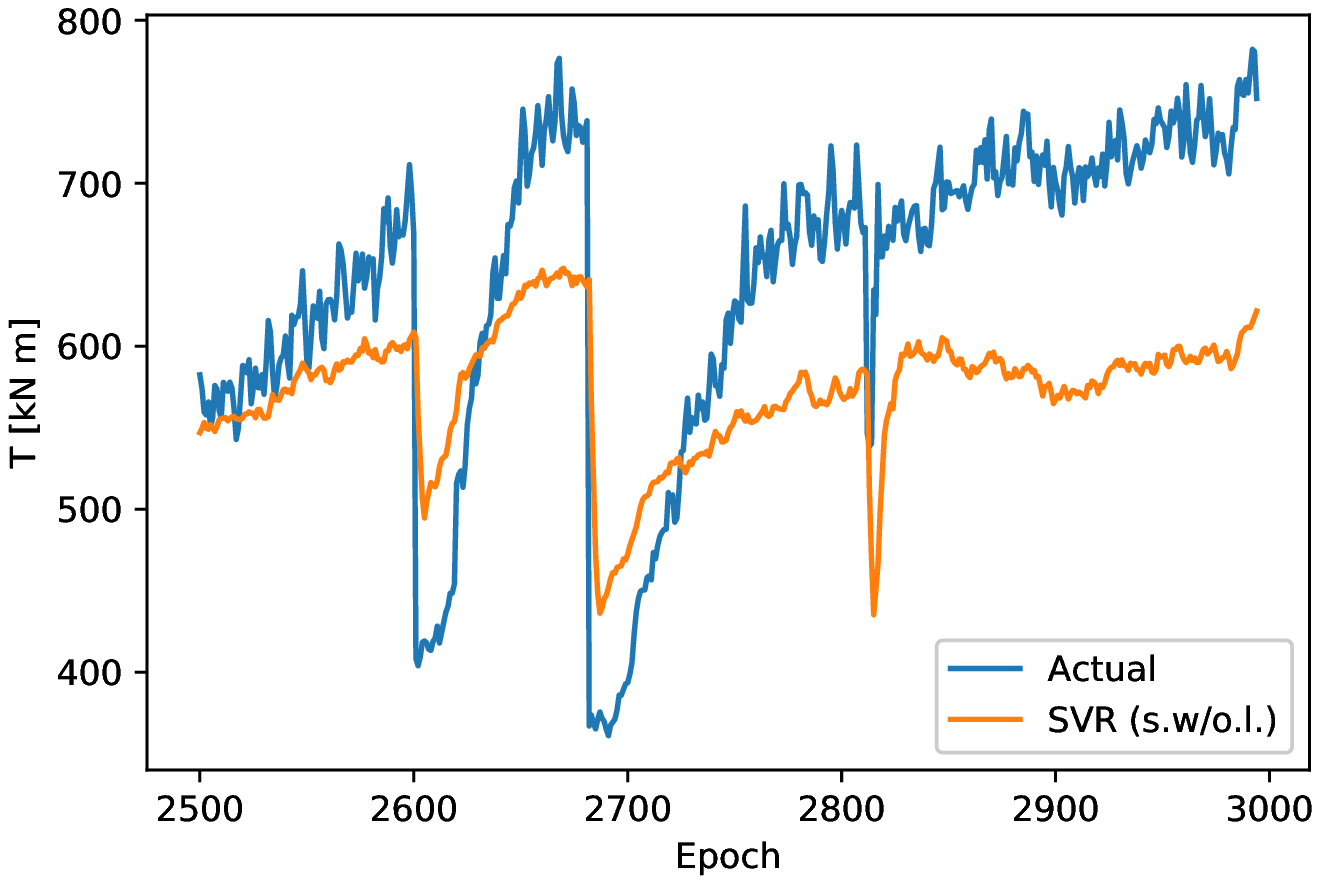}
\end{minipage}
}
\subfigure[SVR (s.w.l.)]{
\begin{minipage}{0.22\textwidth}
\includegraphics[width=1.2\textwidth]{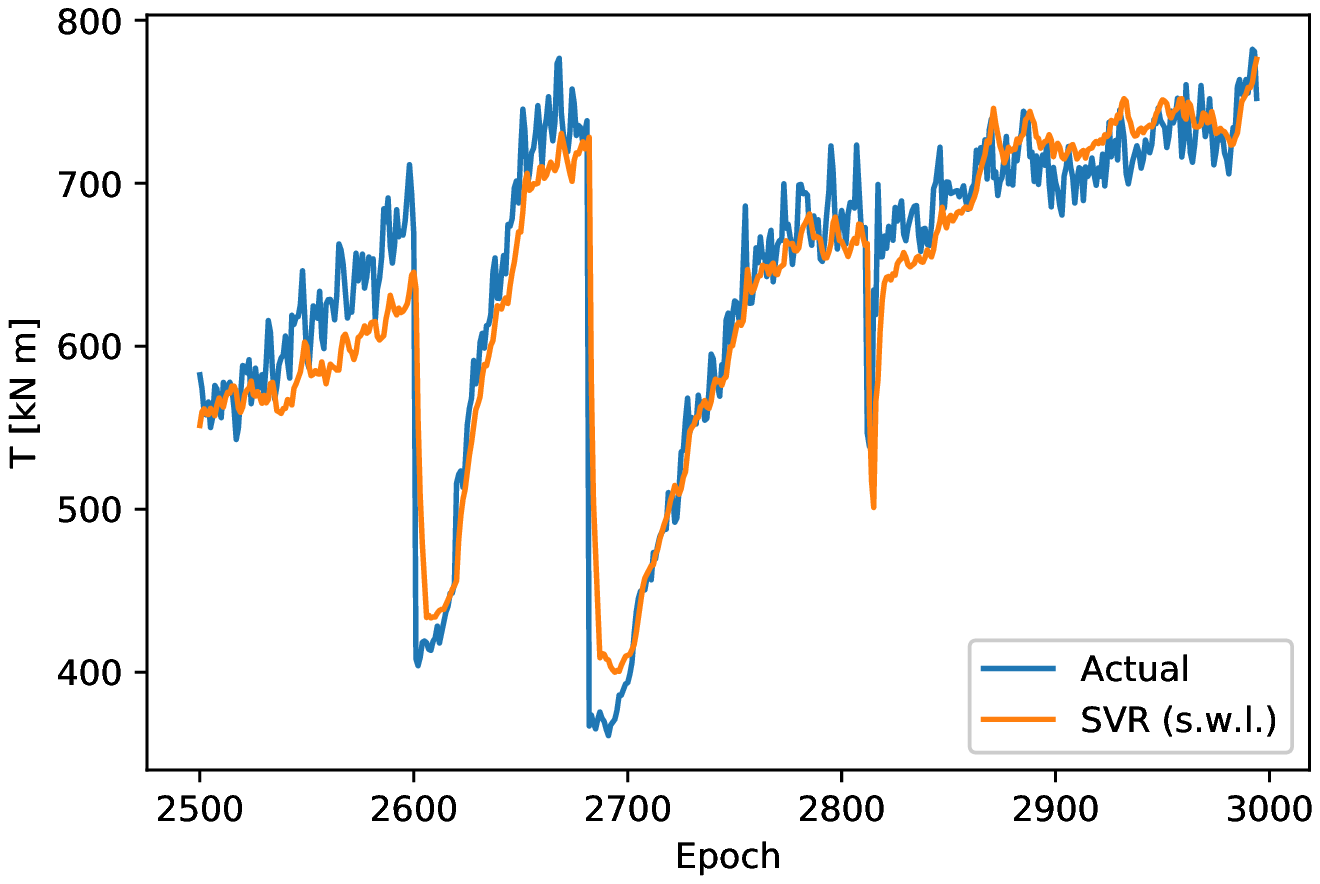}
\end{minipage}
}
\subfigure[SVR (m.w/o.l.)]{
\begin{minipage}{0.22\textwidth}
\includegraphics[width=1.2\textwidth]{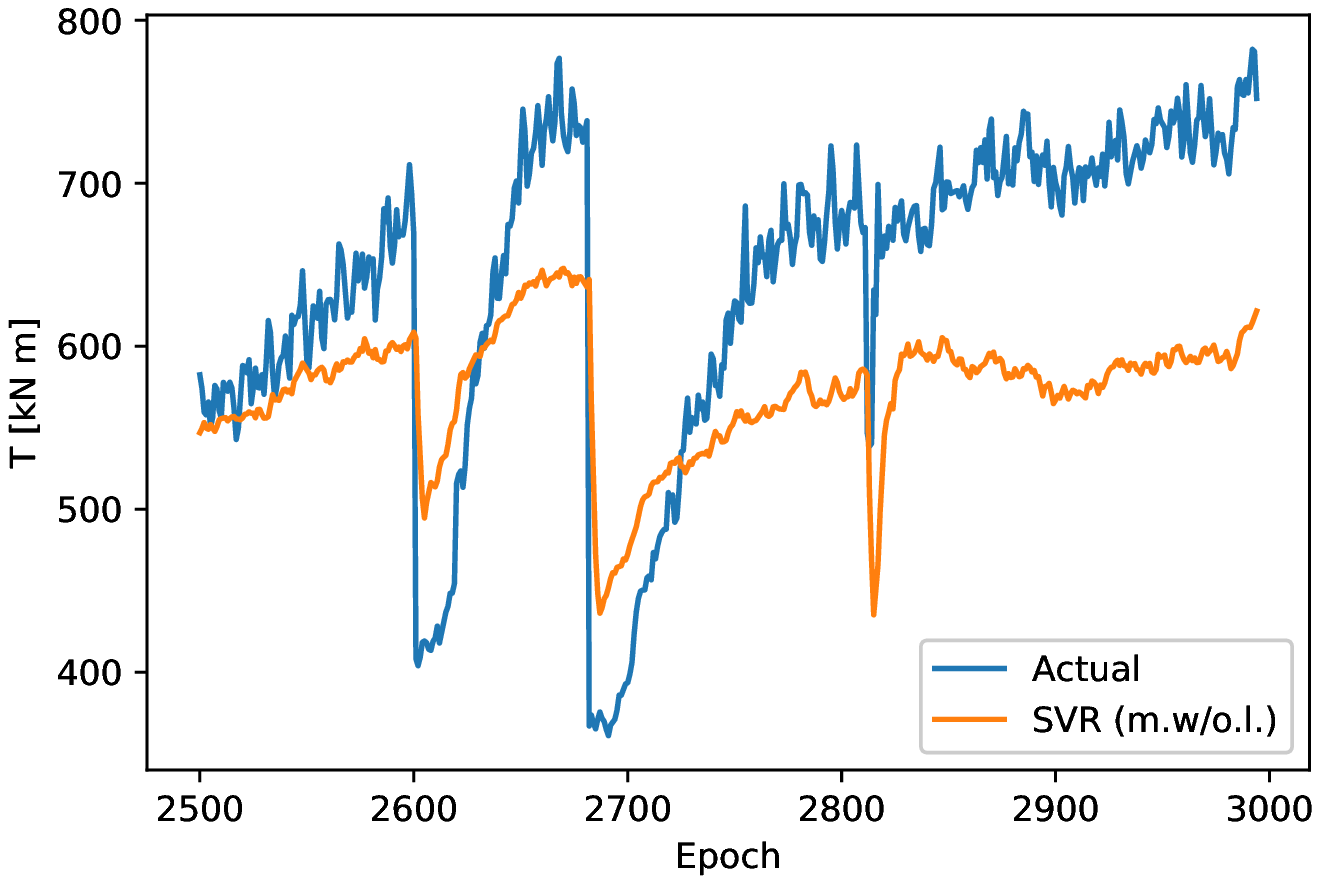}
\end{minipage}
}
\subfigure[SVR (m.w.l.)]{
\begin{minipage}{0.22\textwidth}
\includegraphics[width=1.2\textwidth]{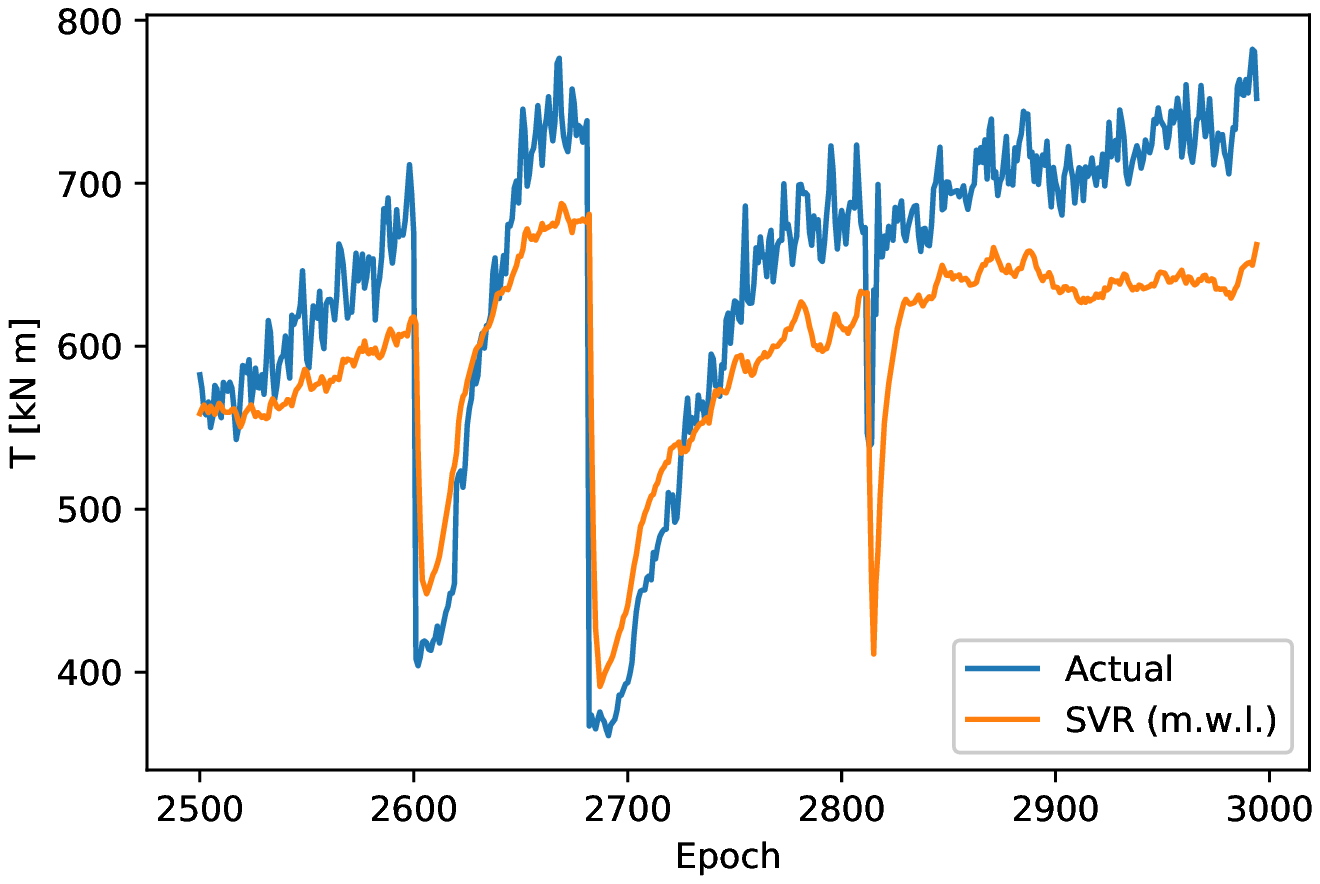}
\end{minipage}
}

\subfigure[RF (s.w/o.l.)]{
\begin{minipage}{0.22\textwidth}
\includegraphics[width=1.2\textwidth]{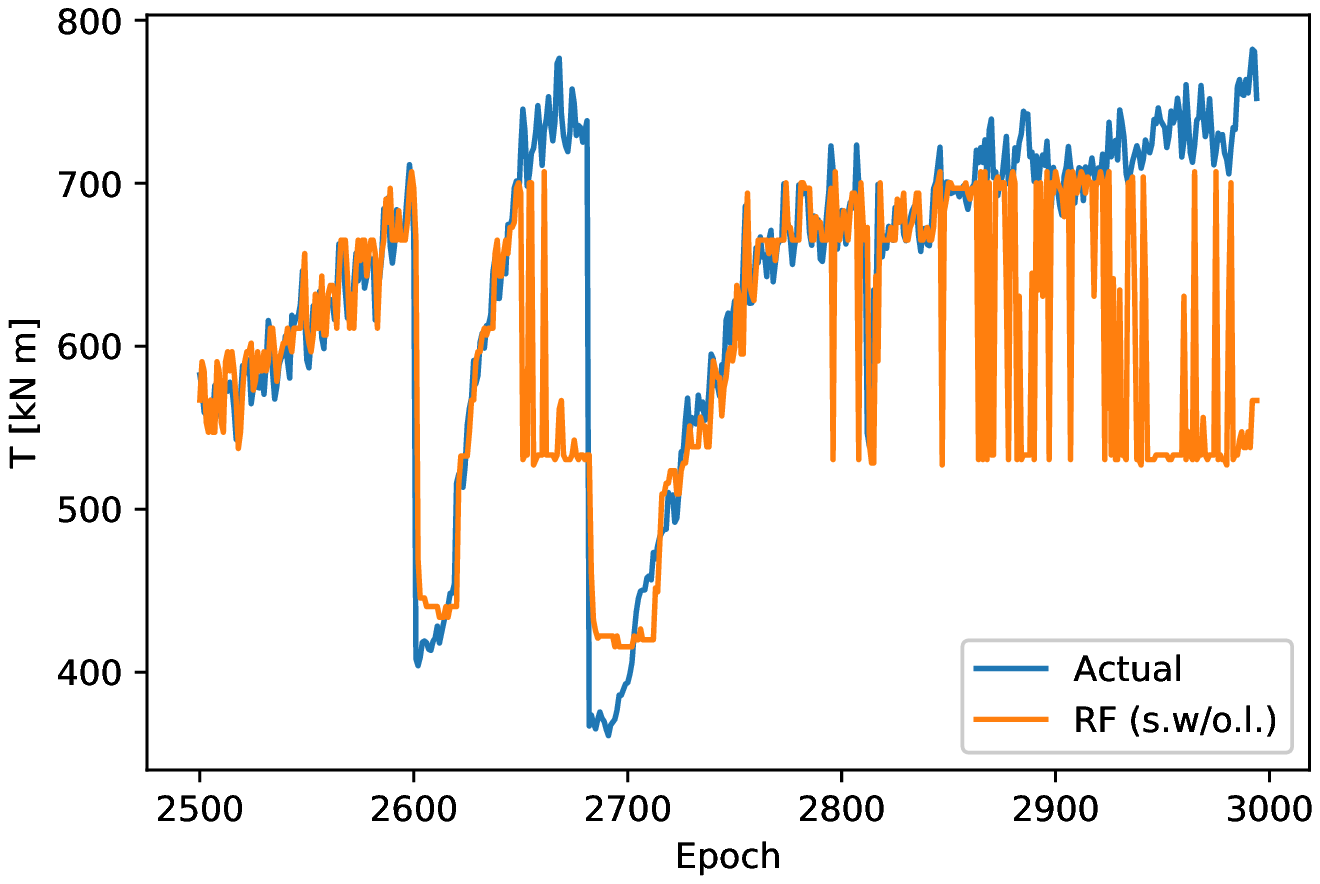}
\end{minipage}
}
\subfigure[RF (s.w.l.)]{
\begin{minipage}{0.22\textwidth}
\includegraphics[width=1.2\textwidth]{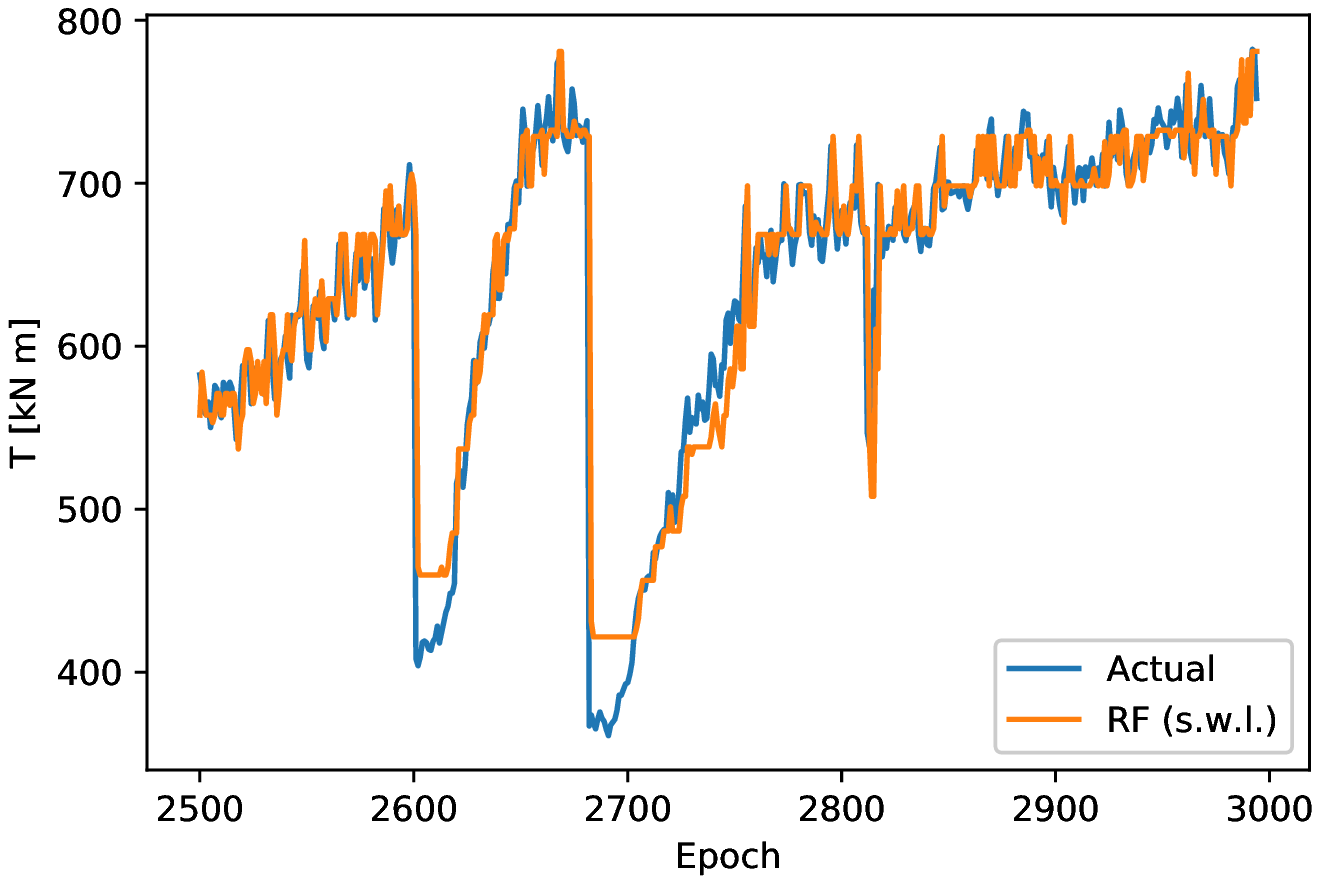}
\end{minipage}
}
\subfigure[RF (m.w/o.l.)]{
\begin{minipage}{0.22\textwidth}
\includegraphics[width=1.2\textwidth]{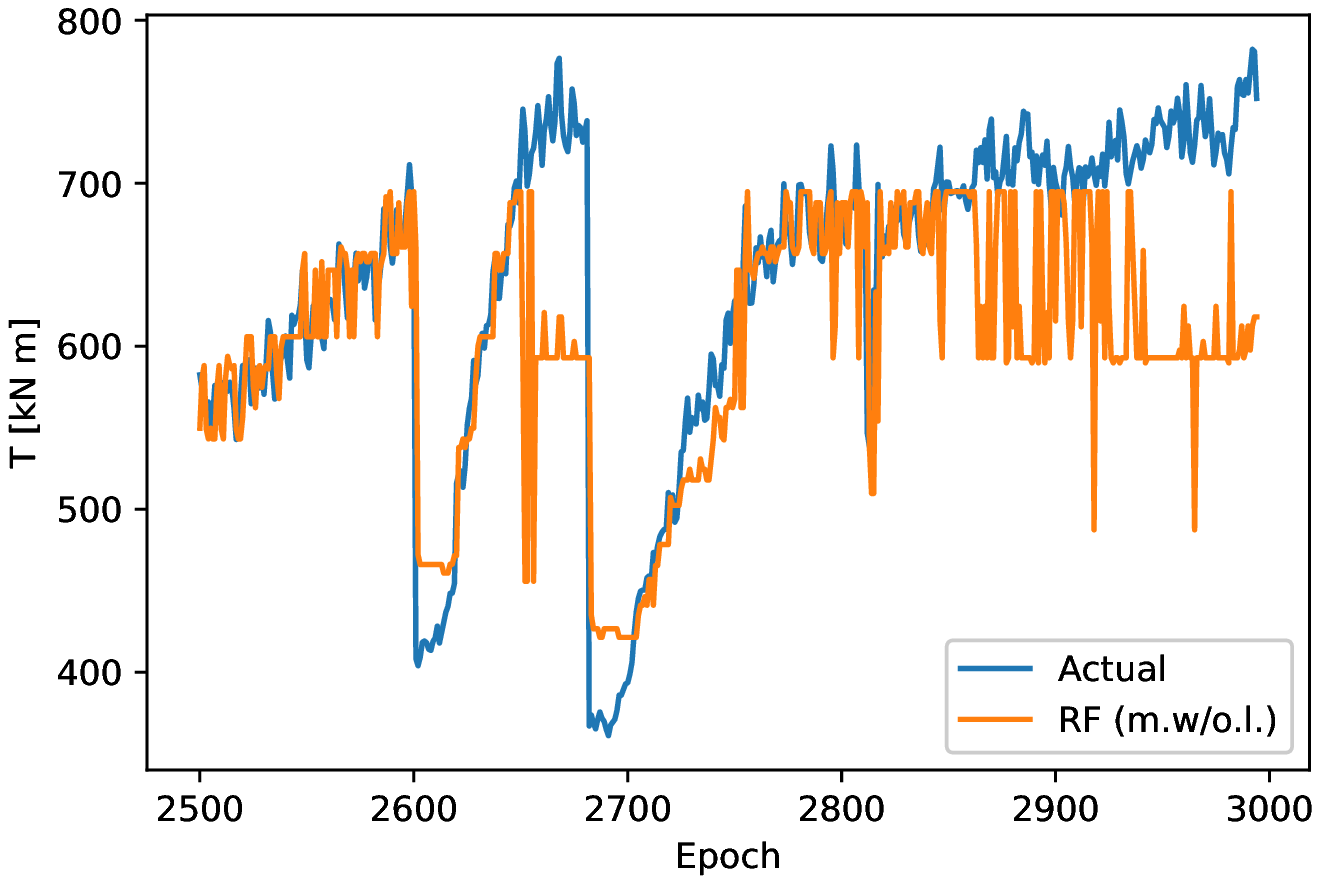}
\end{minipage}
}
\subfigure[RF (m.w.l.)]{
\begin{minipage}{0.22\textwidth}
\includegraphics[width=1.2\textwidth]{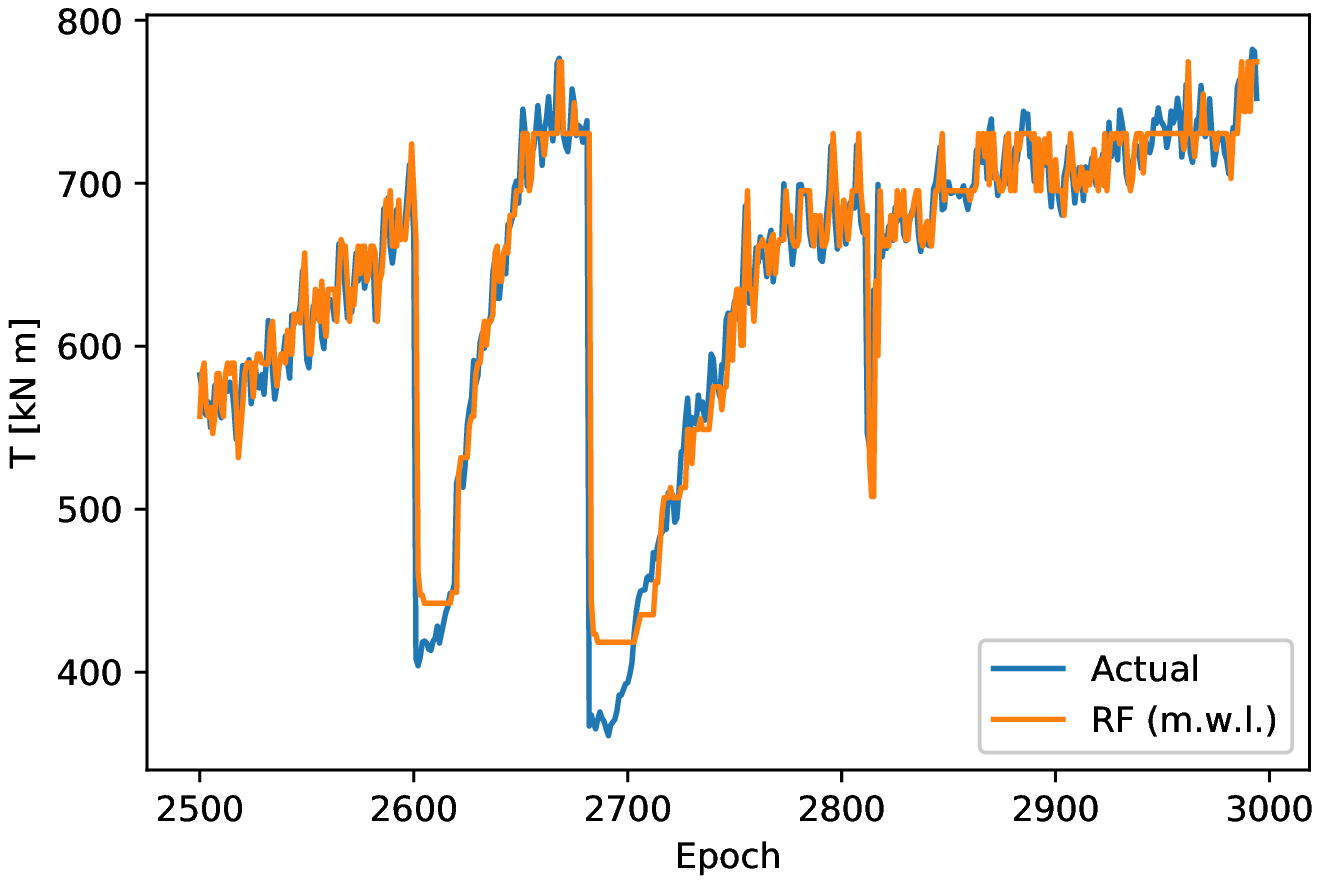}
\end{minipage}
}

\subfigure[FNN (s.w/o.l.)]{
\begin{minipage}{0.22\textwidth}
\includegraphics[width=1.2\textwidth]{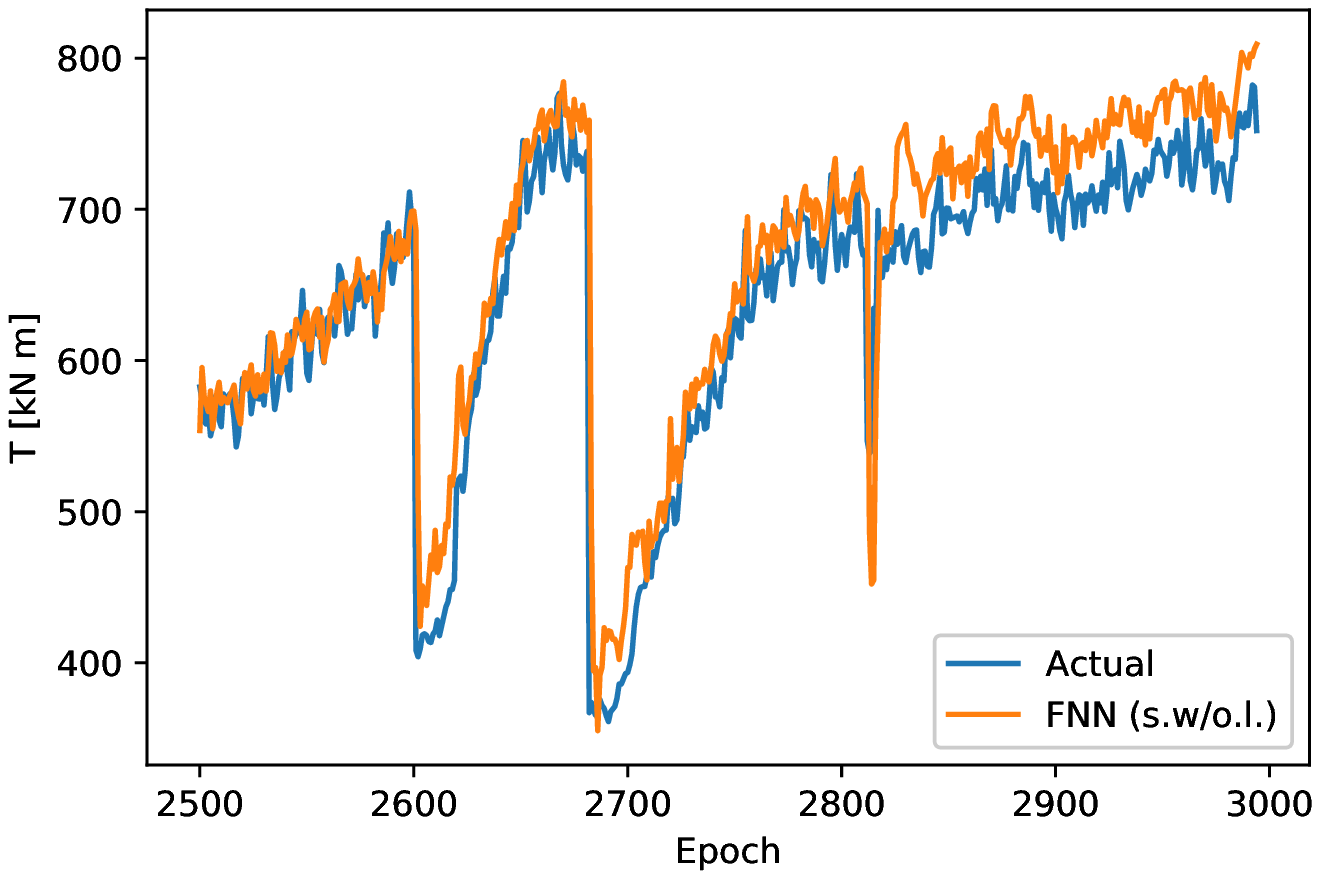}
\end{minipage}
}
\subfigure[FNN (s.w.l.)]{
\begin{minipage}{0.22\textwidth}
\includegraphics[width=1.2\textwidth]{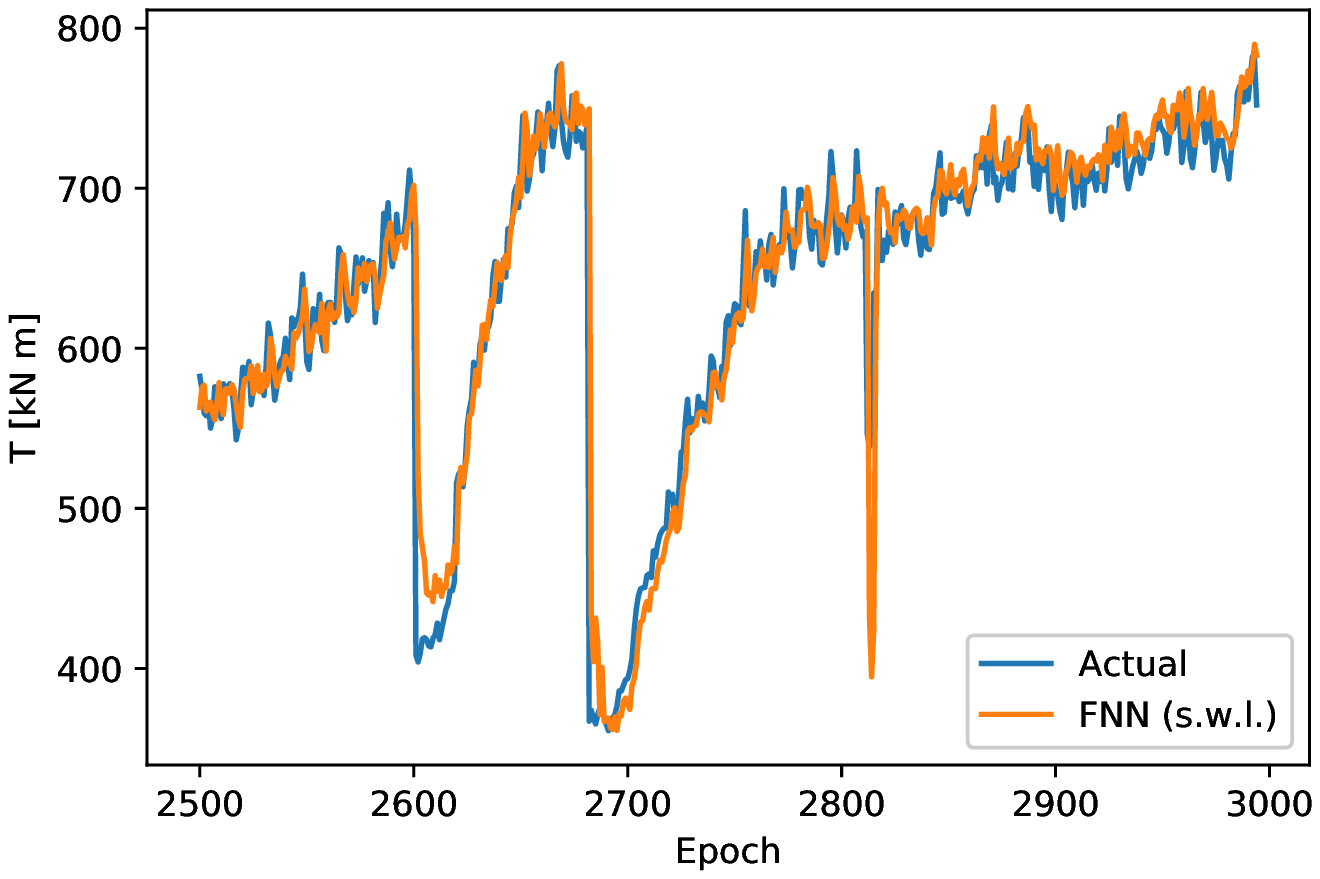}
\end{minipage}
}
\subfigure[FNN (m.w/o.l.)]{
\begin{minipage}{0.22\textwidth}
\includegraphics[width=1.2\textwidth]{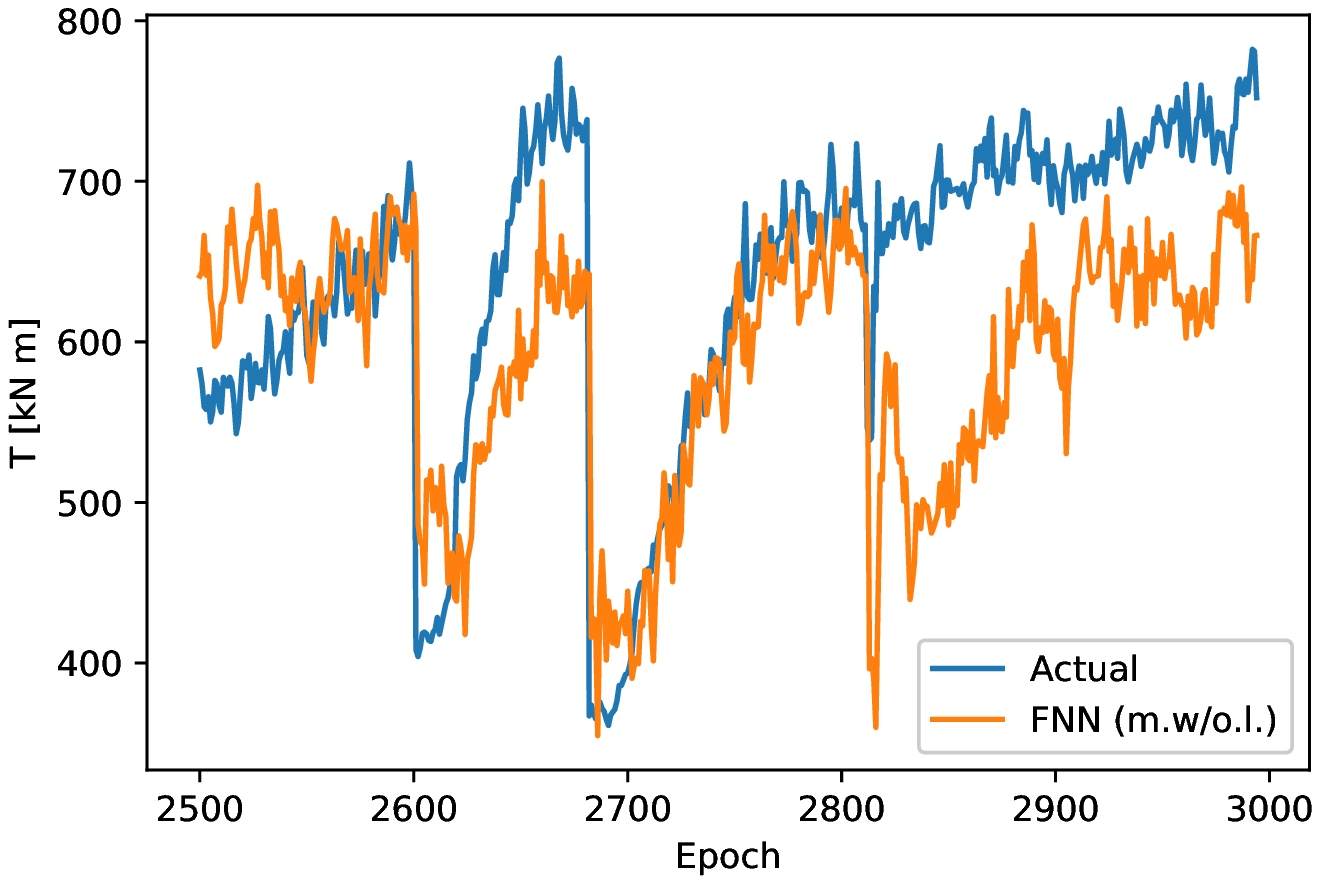}
\end{minipage}
}
\subfigure[FNN (m.w.l.)]{
\begin{minipage}{0.22\textwidth}
\includegraphics[width=1.2\textwidth]{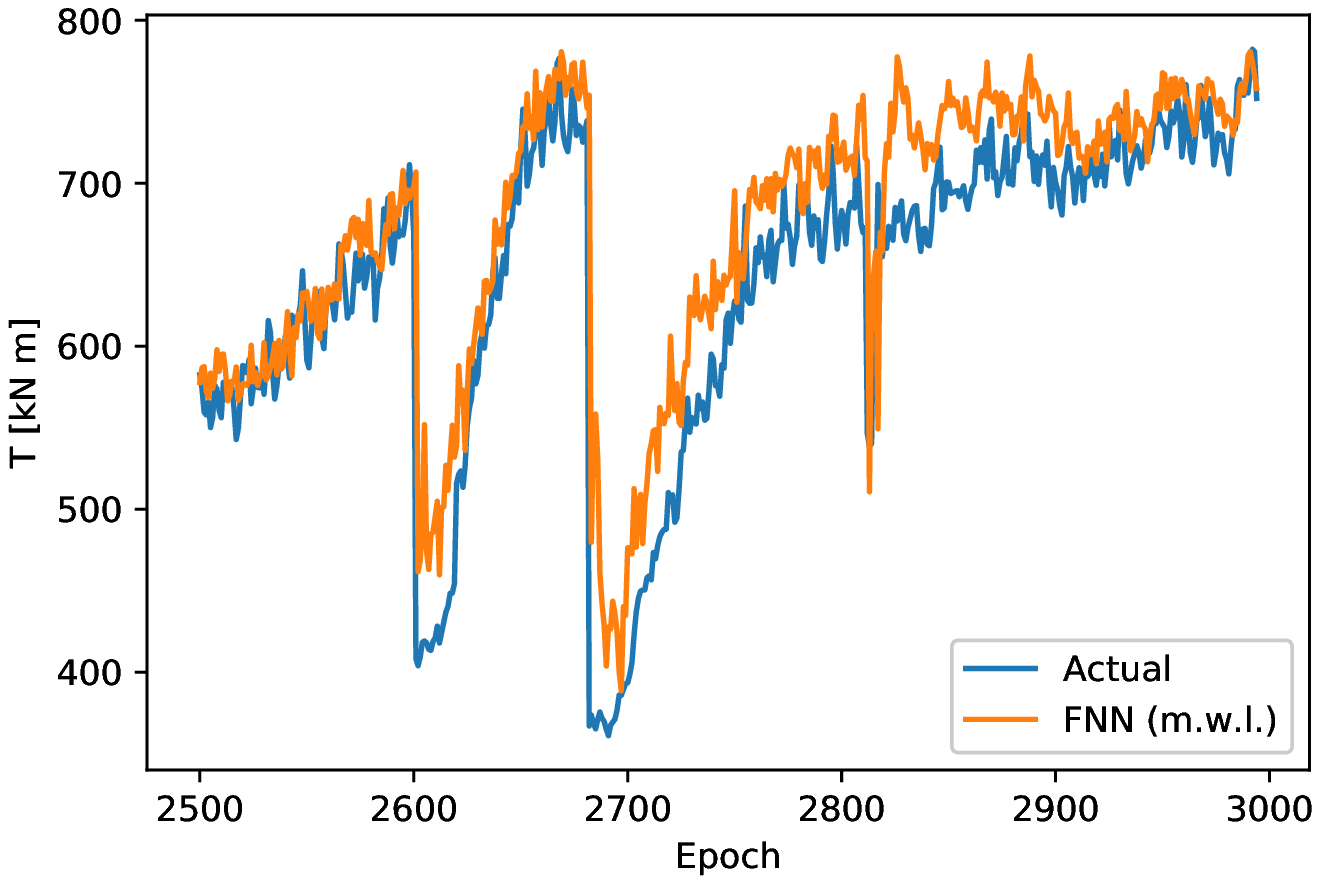}
\end{minipage}
}

\subfigure[RNN (s.w/o.l.)]{
\begin{minipage}{0.22\textwidth}
\includegraphics[width=1.2\textwidth]{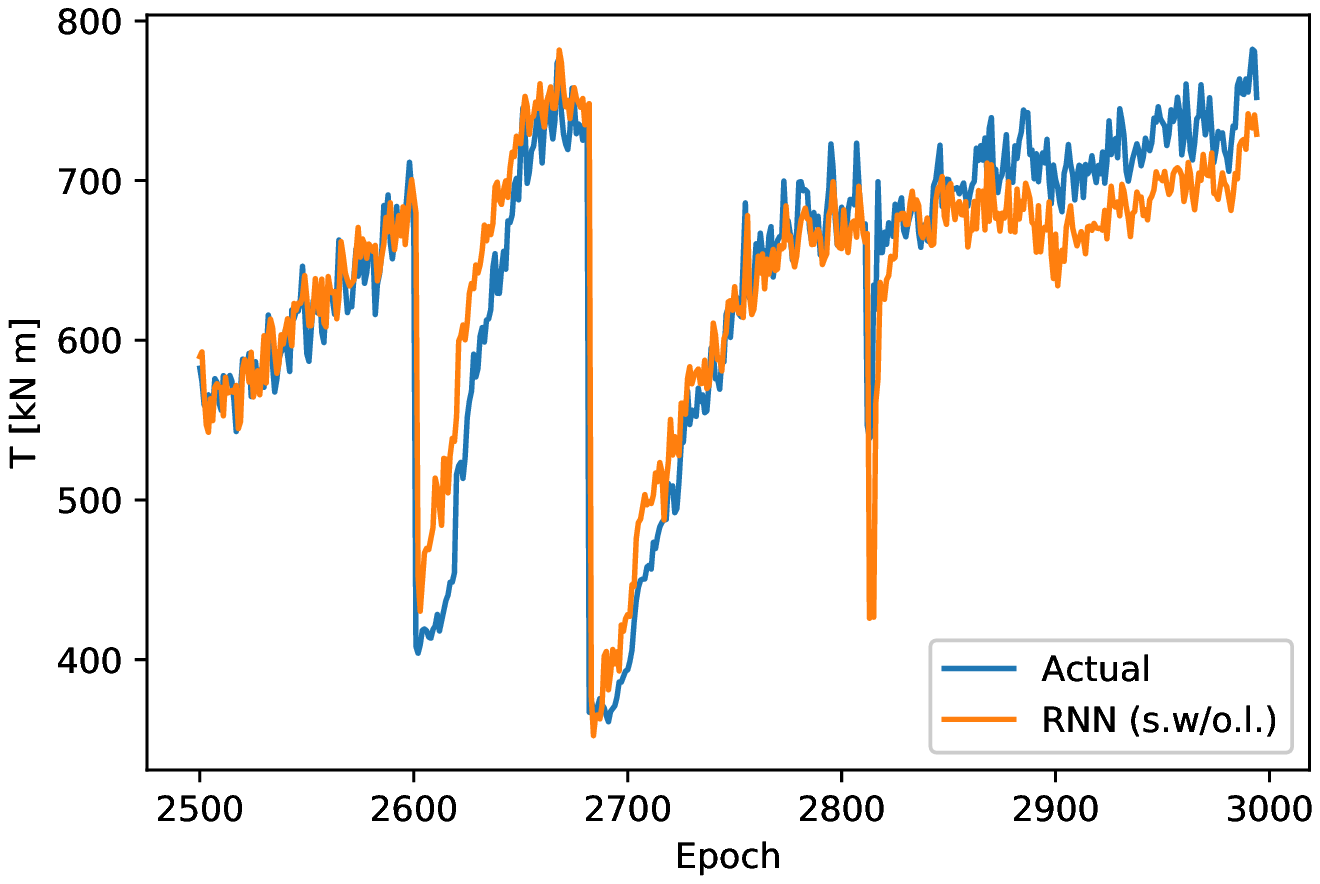}
\end{minipage}
}
\subfigure[RNN (s.w.l.)]{
\begin{minipage}{0.22\textwidth}
\includegraphics[width=1.2\textwidth]{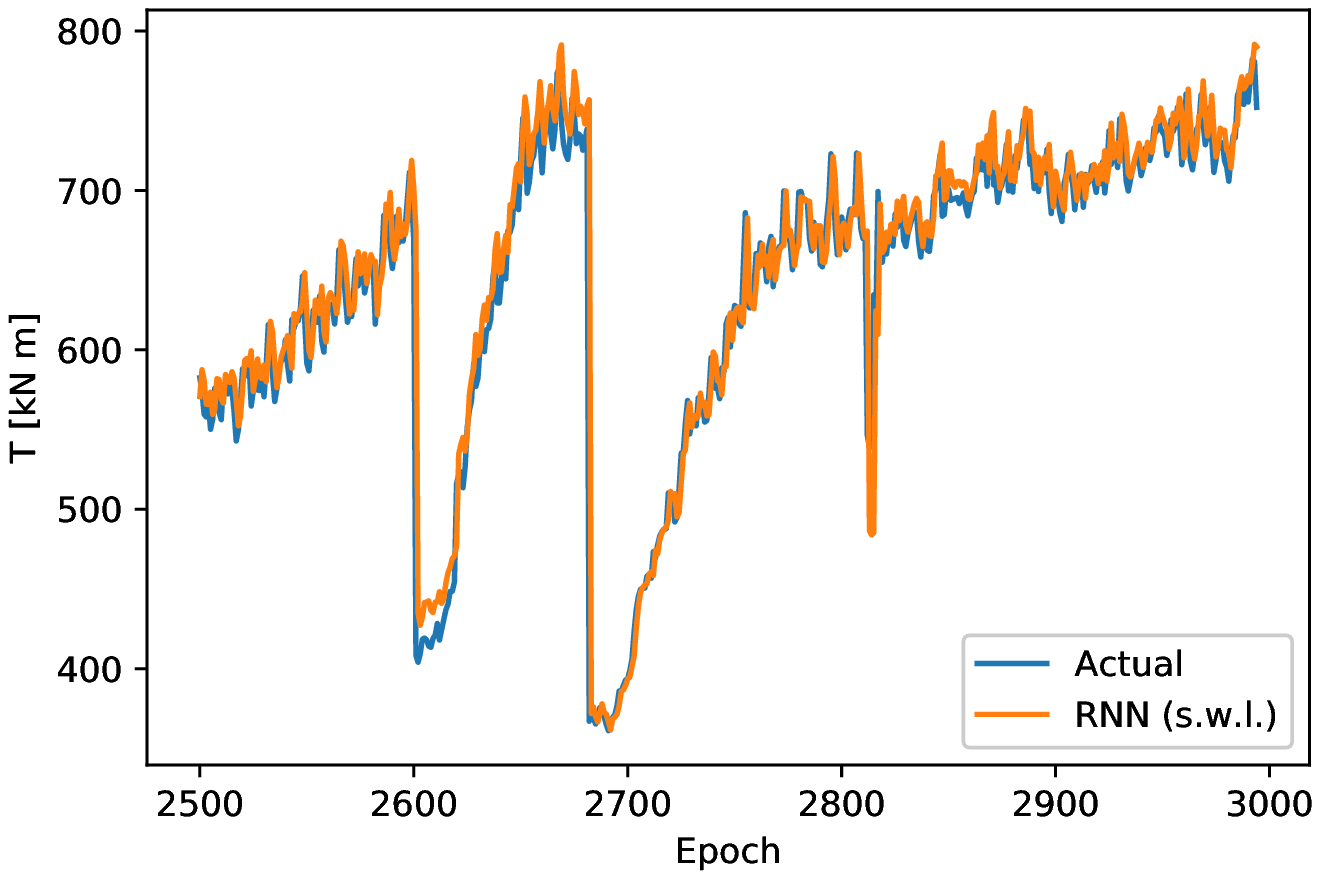}
\end{minipage}
}
\subfigure[RNN (m.w/o.l.)]{
\begin{minipage}{0.22\textwidth}
\includegraphics[width=1.2\textwidth]{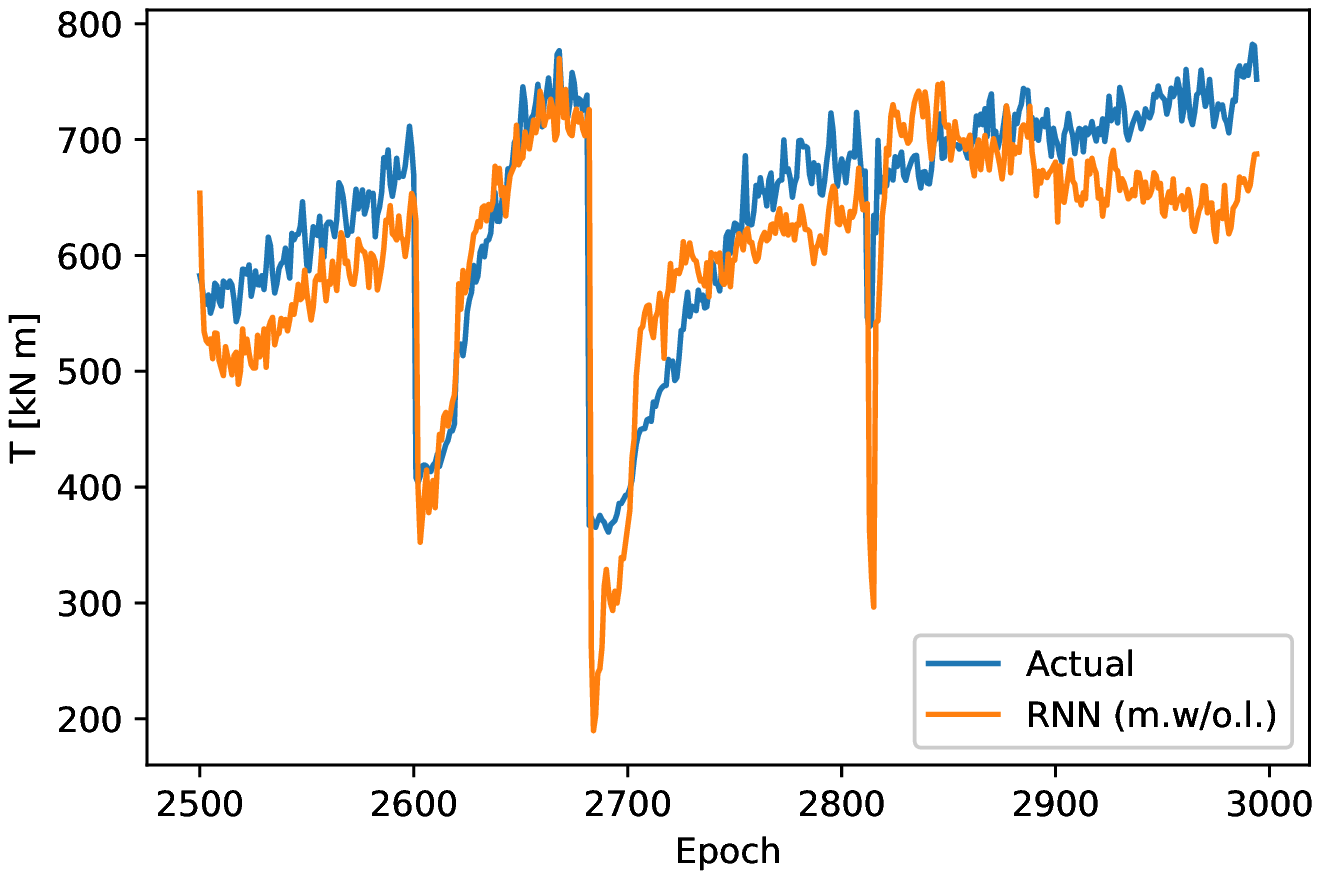}
\end{minipage}
}
\subfigure[RNN (m.w.l.)]{
\begin{minipage}{0.22\textwidth}
\includegraphics[width=1.2\textwidth]{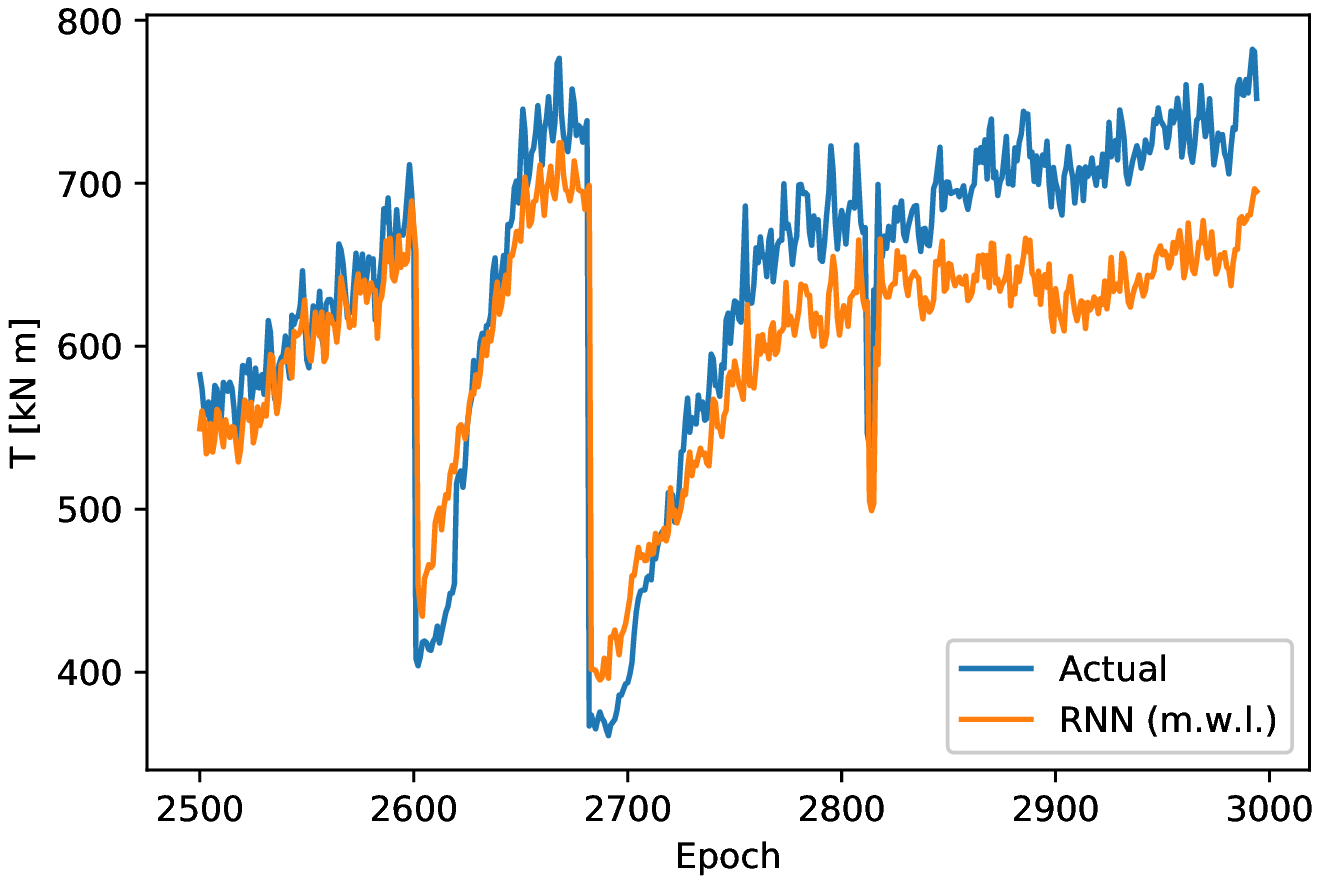}
\end{minipage}
}

\subfigure[GRU (s.w/o.l.)]{
\begin{minipage}{0.22\textwidth}
\includegraphics[width=1.2\textwidth]{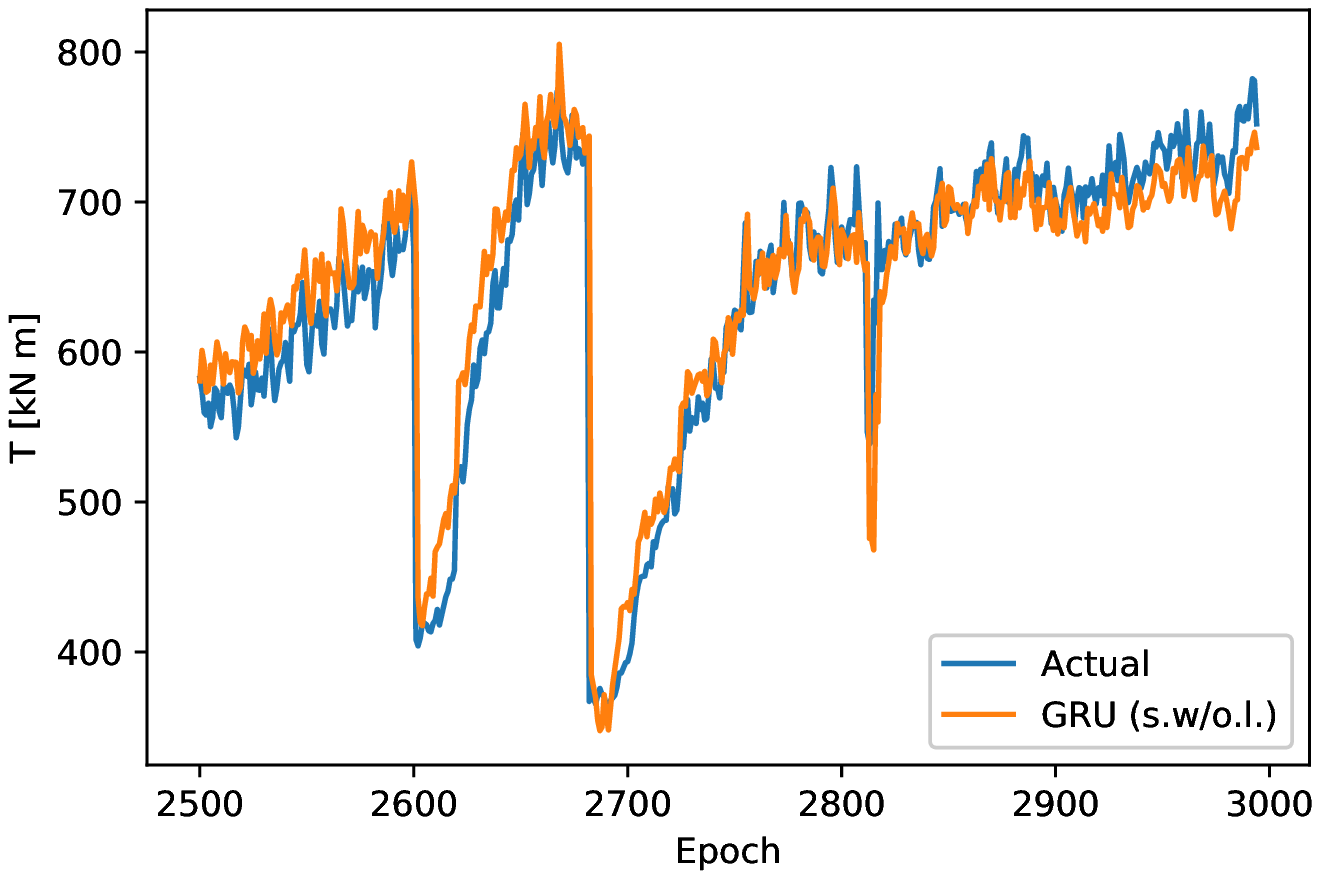}
\end{minipage}
}
\subfigure[GRU (s.w.l.)]{
\begin{minipage}{0.22\textwidth}
\includegraphics[width=1.2\textwidth]{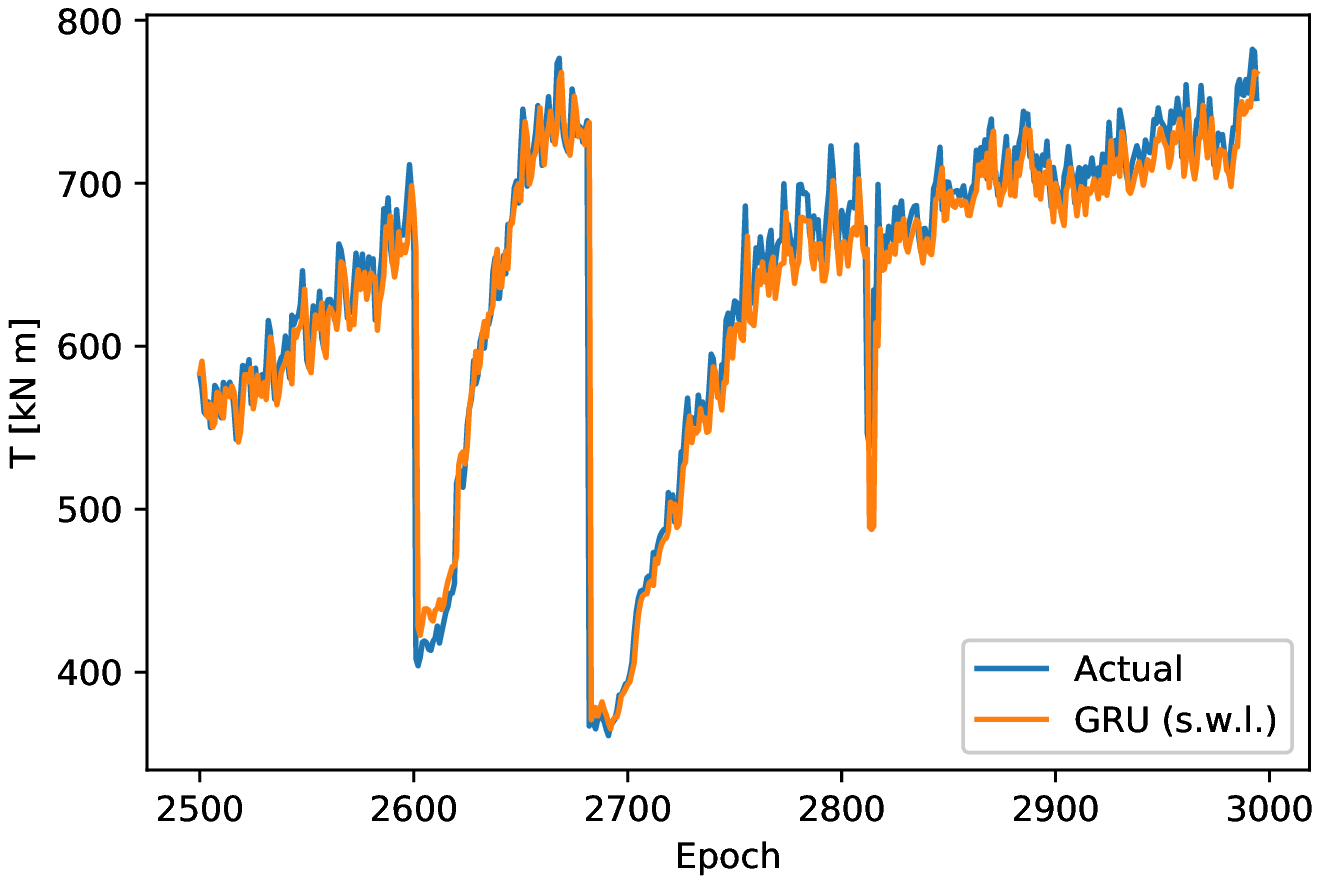}
\end{minipage}
}
\subfigure[GRU (m.w/o.l.)]{
\begin{minipage}{0.22\textwidth}
\includegraphics[width=1.2\textwidth]{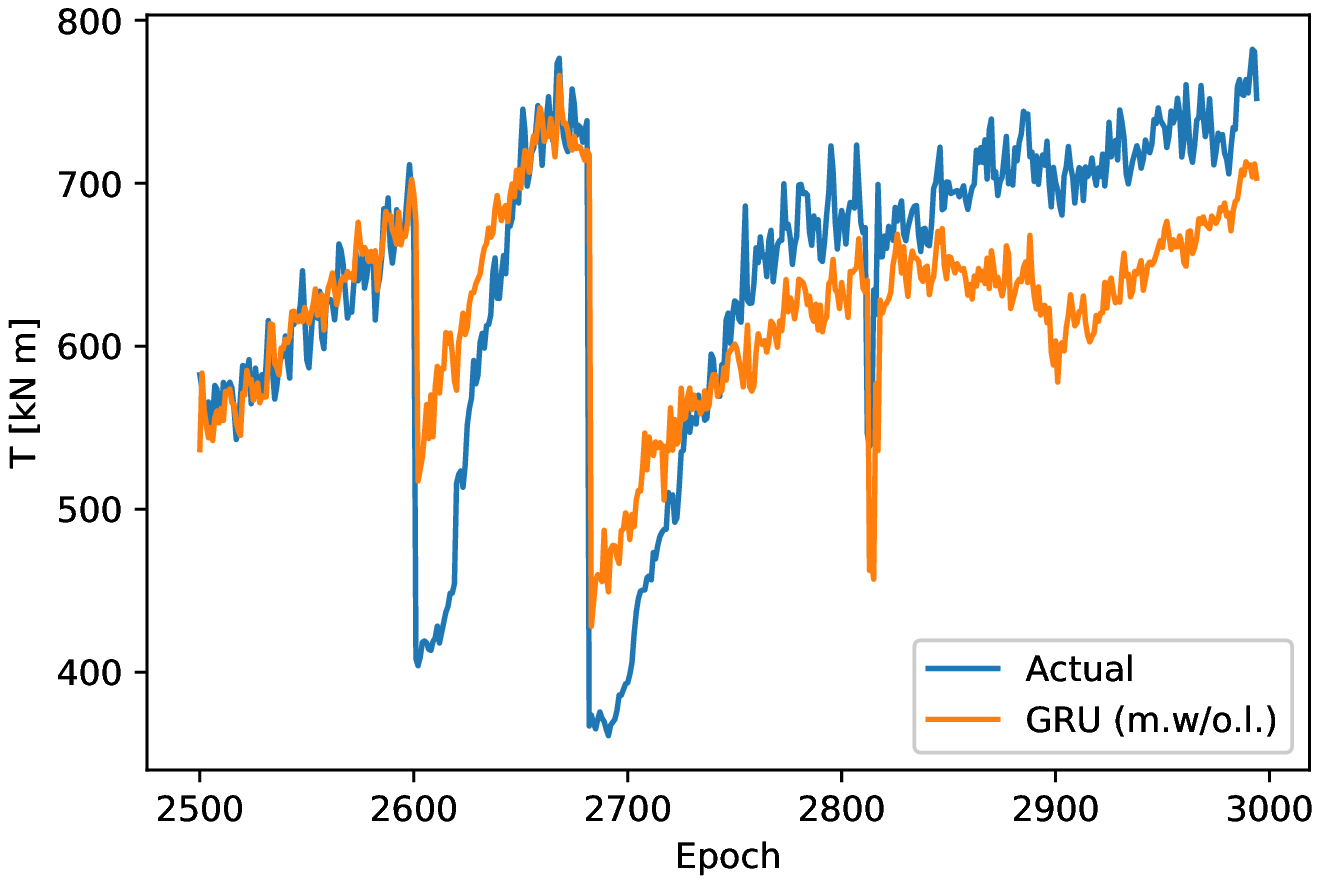}
\end{minipage}
}
\subfigure[GRU (m.w.l.)]{
\begin{minipage}{0.22\textwidth}
\includegraphics[width=1.2\textwidth]{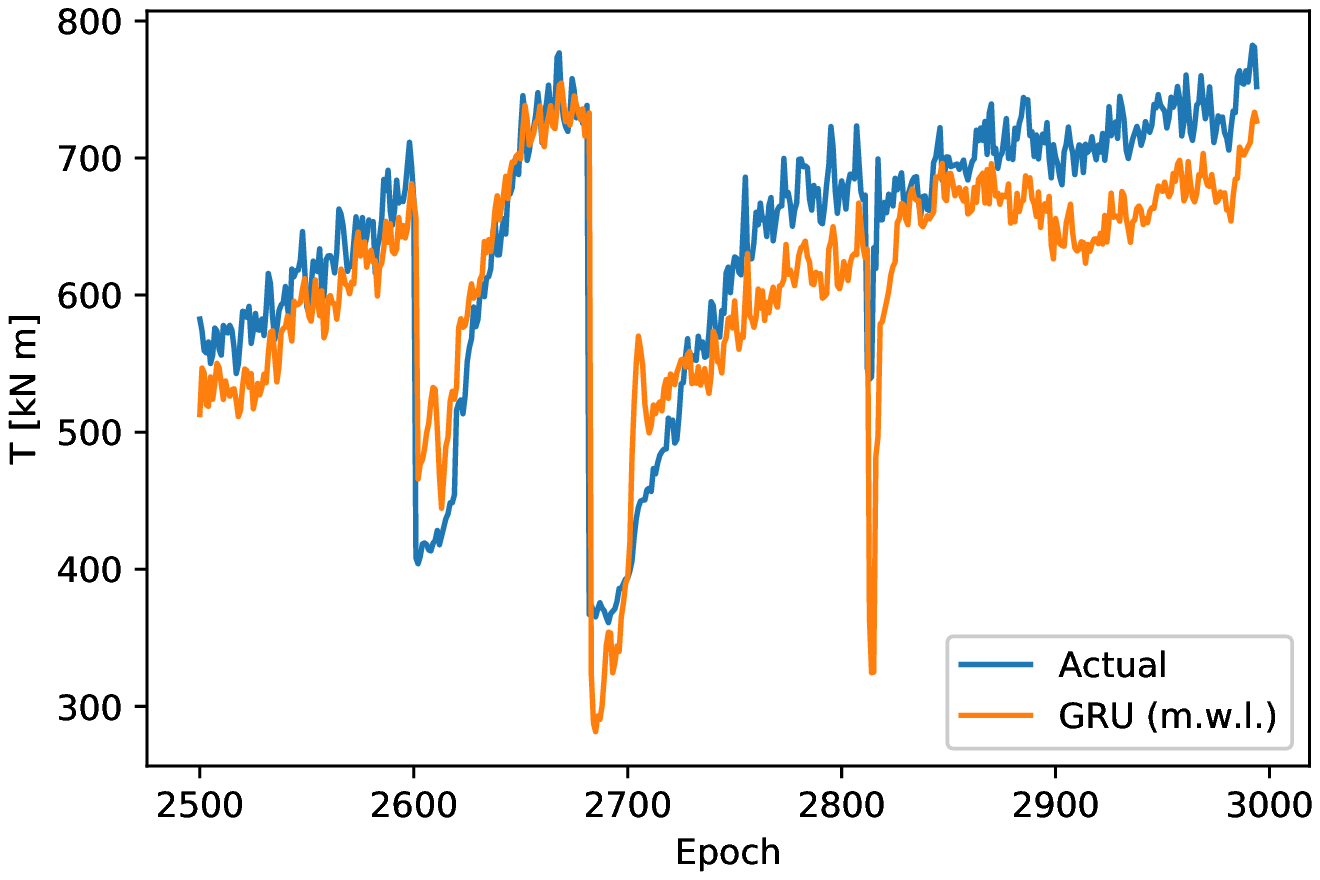}
\end{minipage}
}

\subfigure[LSTM (s.w/o.l.)]{
\begin{minipage}{0.22\textwidth}
\includegraphics[width=1.2\textwidth]{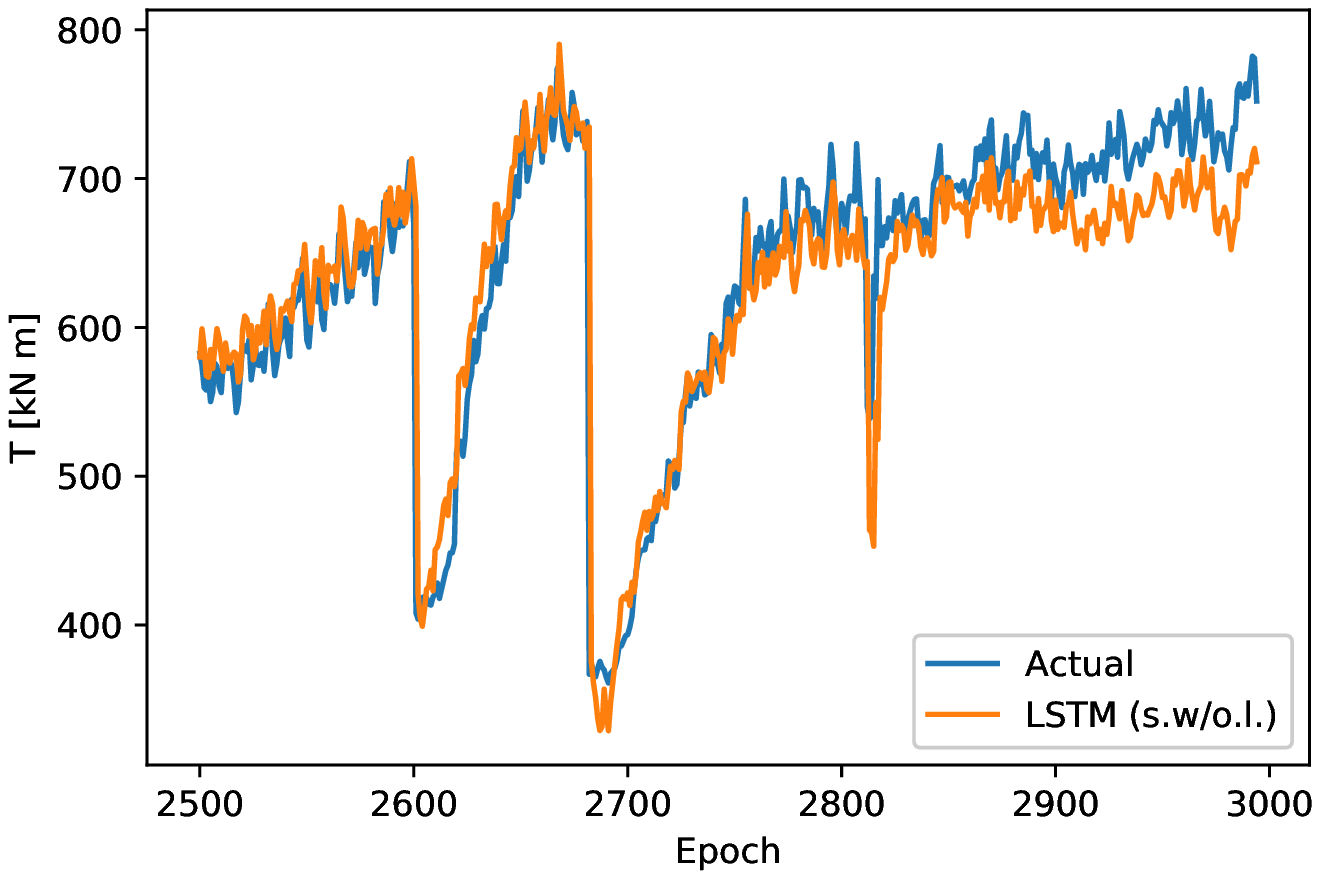}
\end{minipage}
}
\subfigure[LSTM (s.w.l.)]{
\begin{minipage}{0.22\textwidth}
\includegraphics[width=1.2\textwidth]{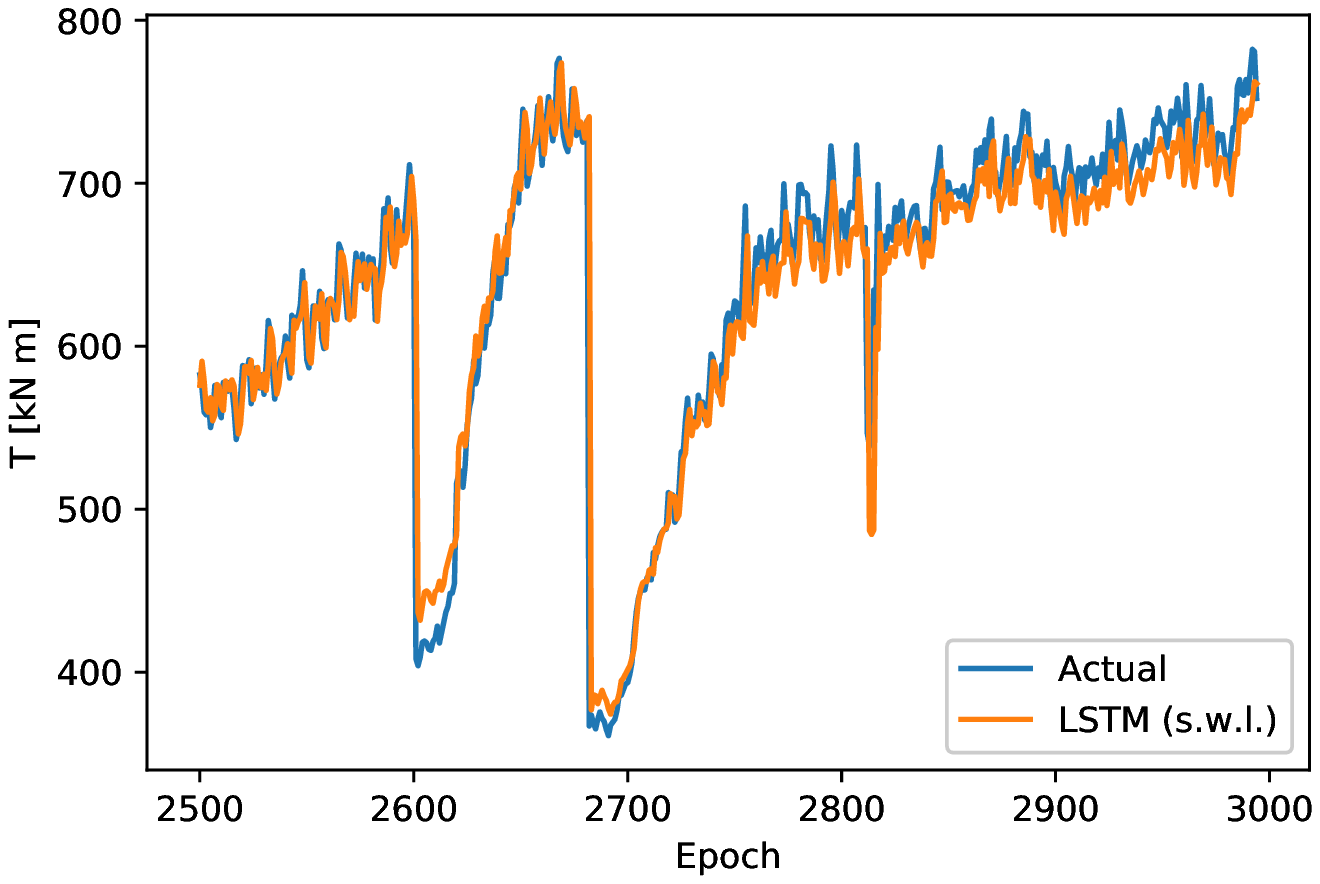}
\end{minipage}
}
\subfigure[LSTM (m.w/o.l.)]{
\begin{minipage}{0.22\textwidth}
\includegraphics[width=1.2\textwidth]{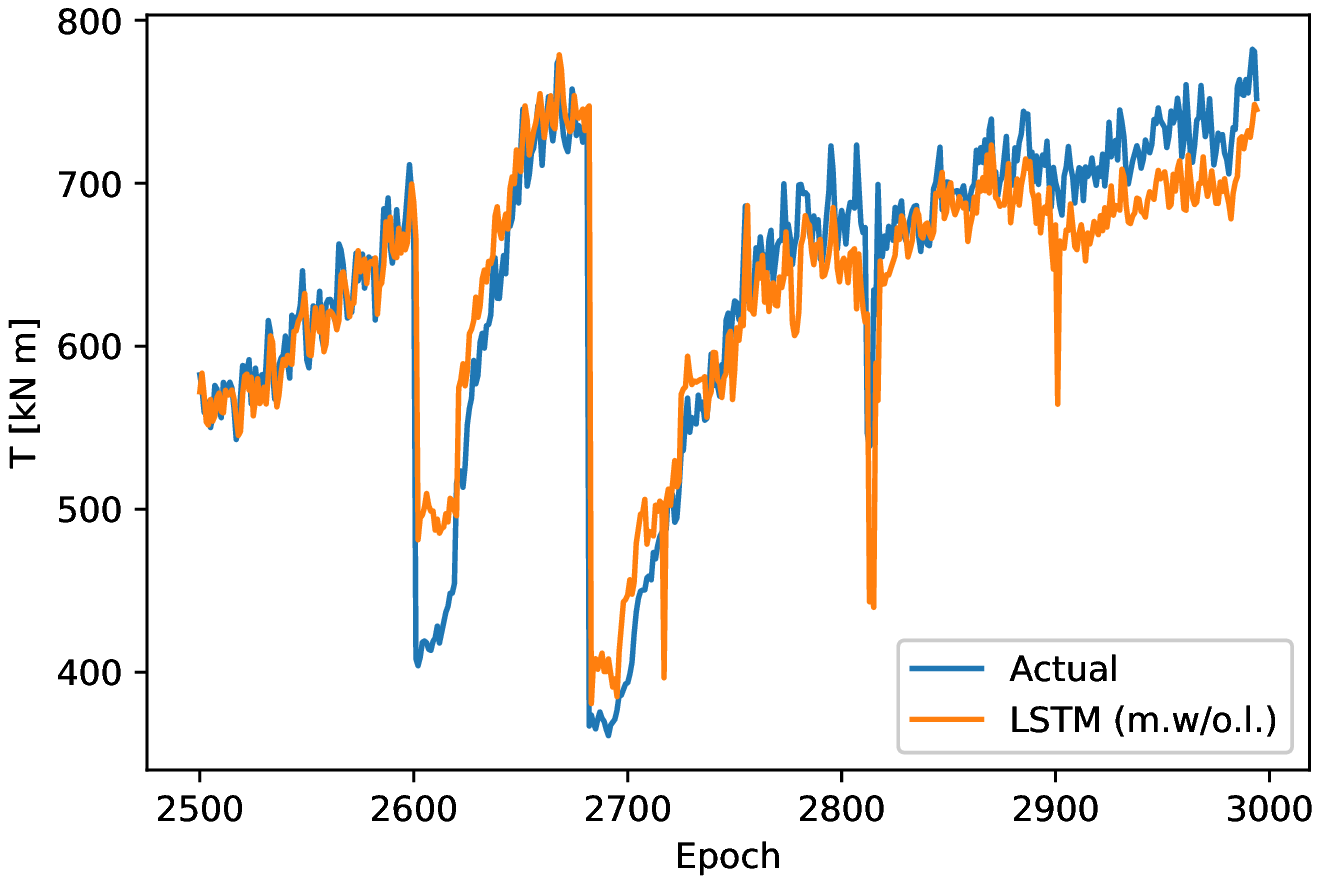}
\end{minipage}
}
\subfigure[LSTM (m.w.l.)]{
\begin{minipage}{0.22\textwidth}
\includegraphics[width=1.2\textwidth]{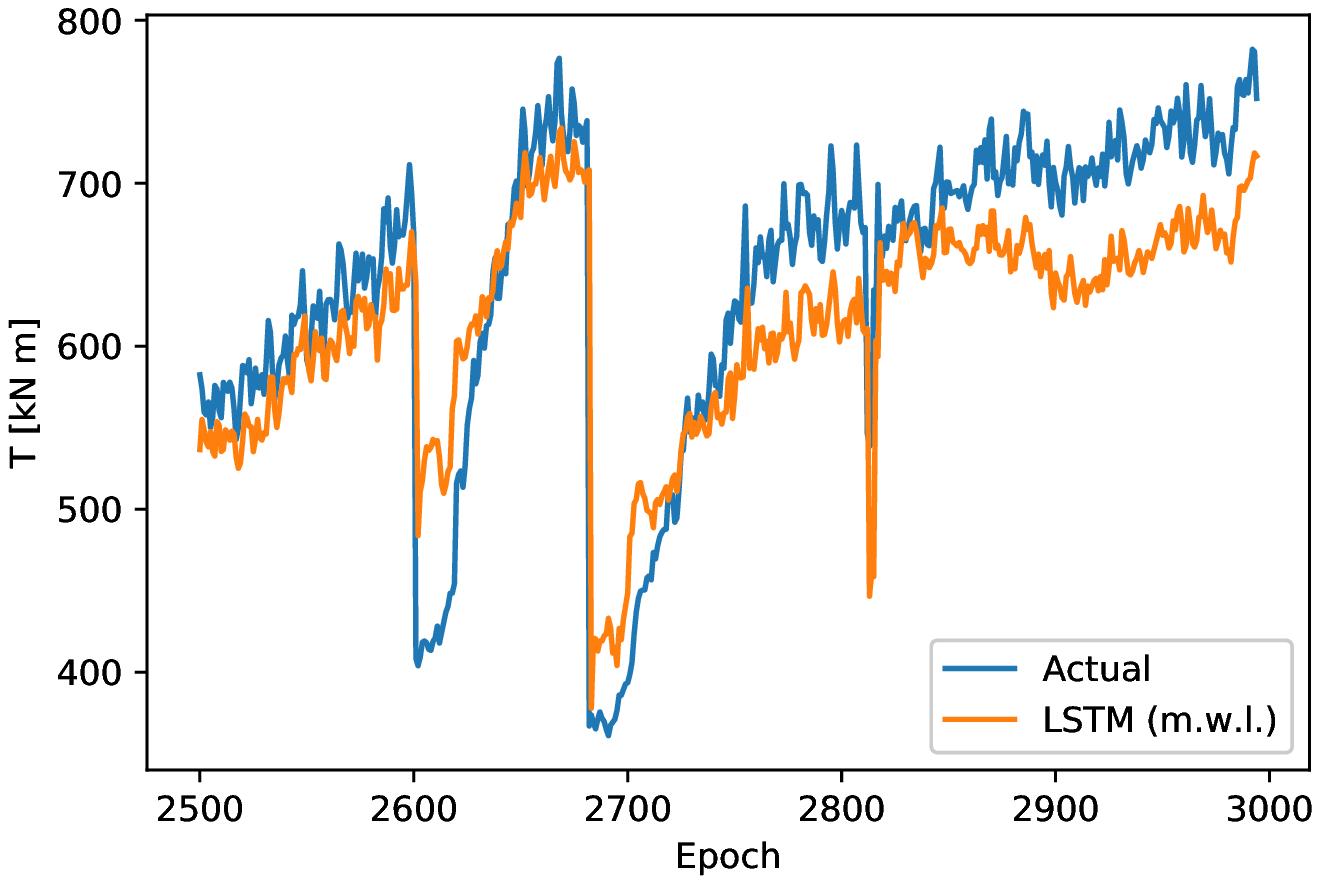}
\end{minipage}
}
\caption{Forecast Results of ``Torque"} \label{fig:torque}
\end{figure}

\begin{figure}[htbp]
\centering
\subfigure[SVR (s.w/o.l.)]{
\begin{minipage}{0.22\textwidth}
\includegraphics[width=1.2\textwidth]{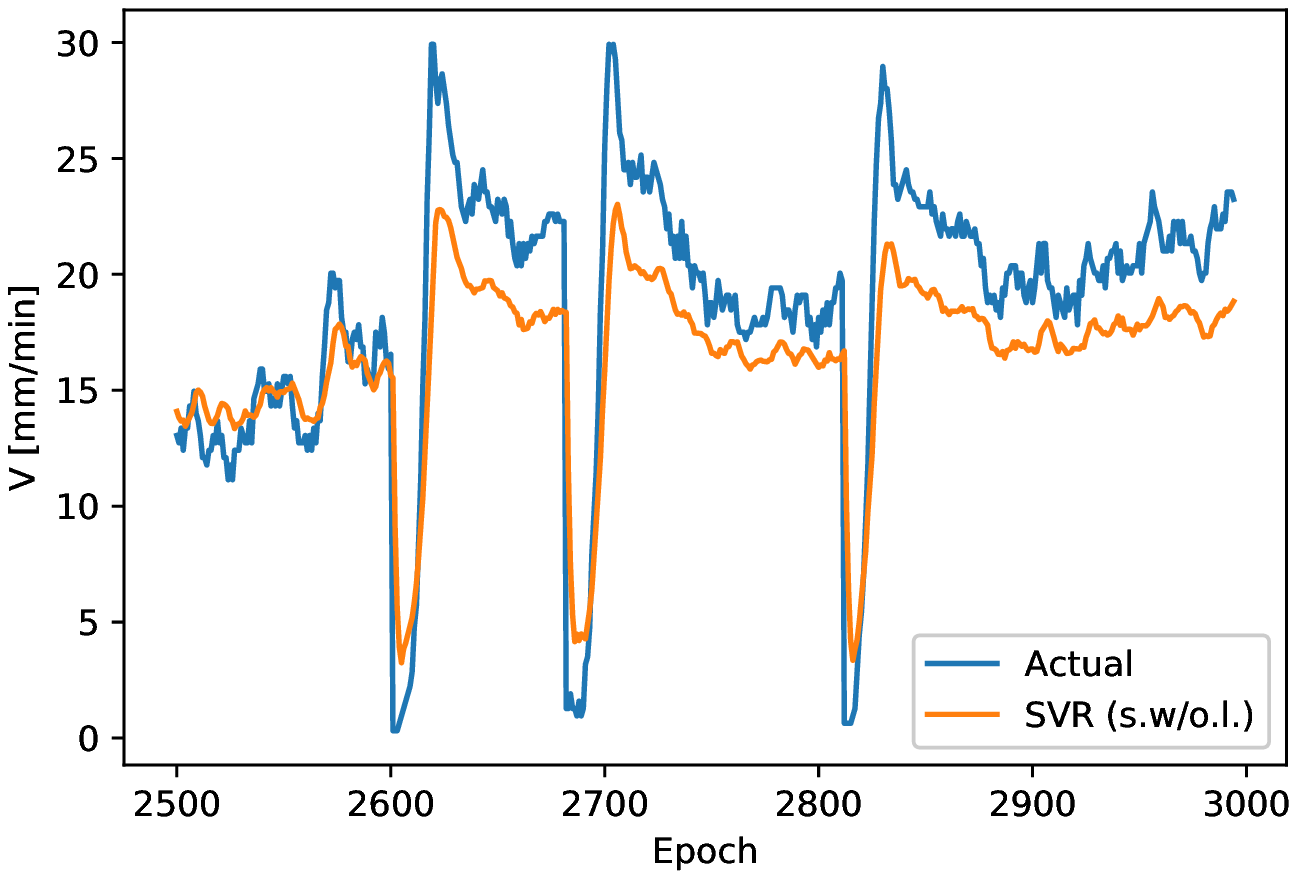}
\end{minipage}
}
\subfigure[SVR (s.w.l.)]{
\begin{minipage}{0.22\textwidth}
\includegraphics[width=1.2\textwidth]{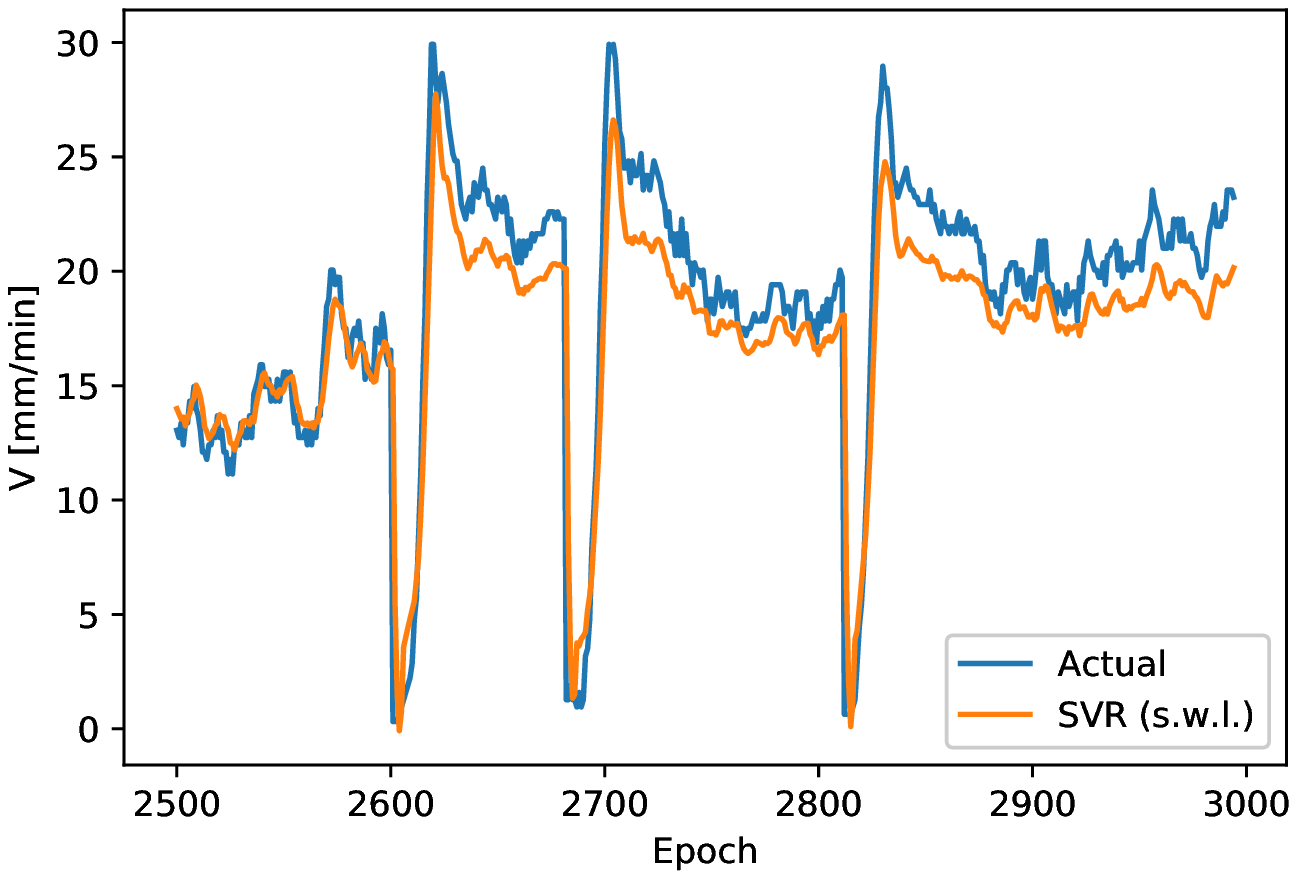}
\end{minipage}
}
\subfigure[SVR (m.w/o.l.)]{
\begin{minipage}{0.22\textwidth}
\includegraphics[width=1.2\textwidth]{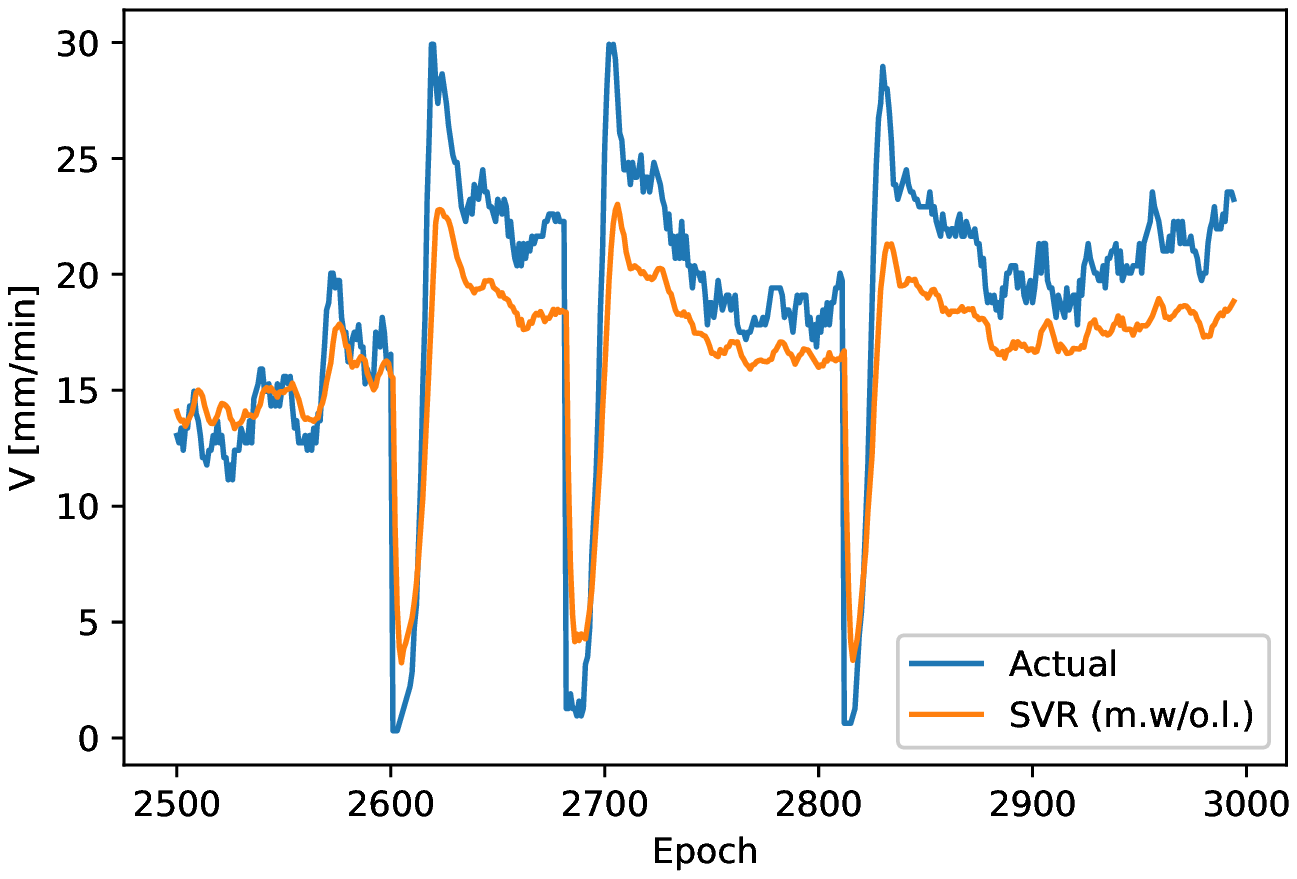}
\end{minipage}
}
\subfigure[SVR (m.w.l.)]{
\begin{minipage}{0.22\textwidth}
\includegraphics[width=1.2\textwidth]{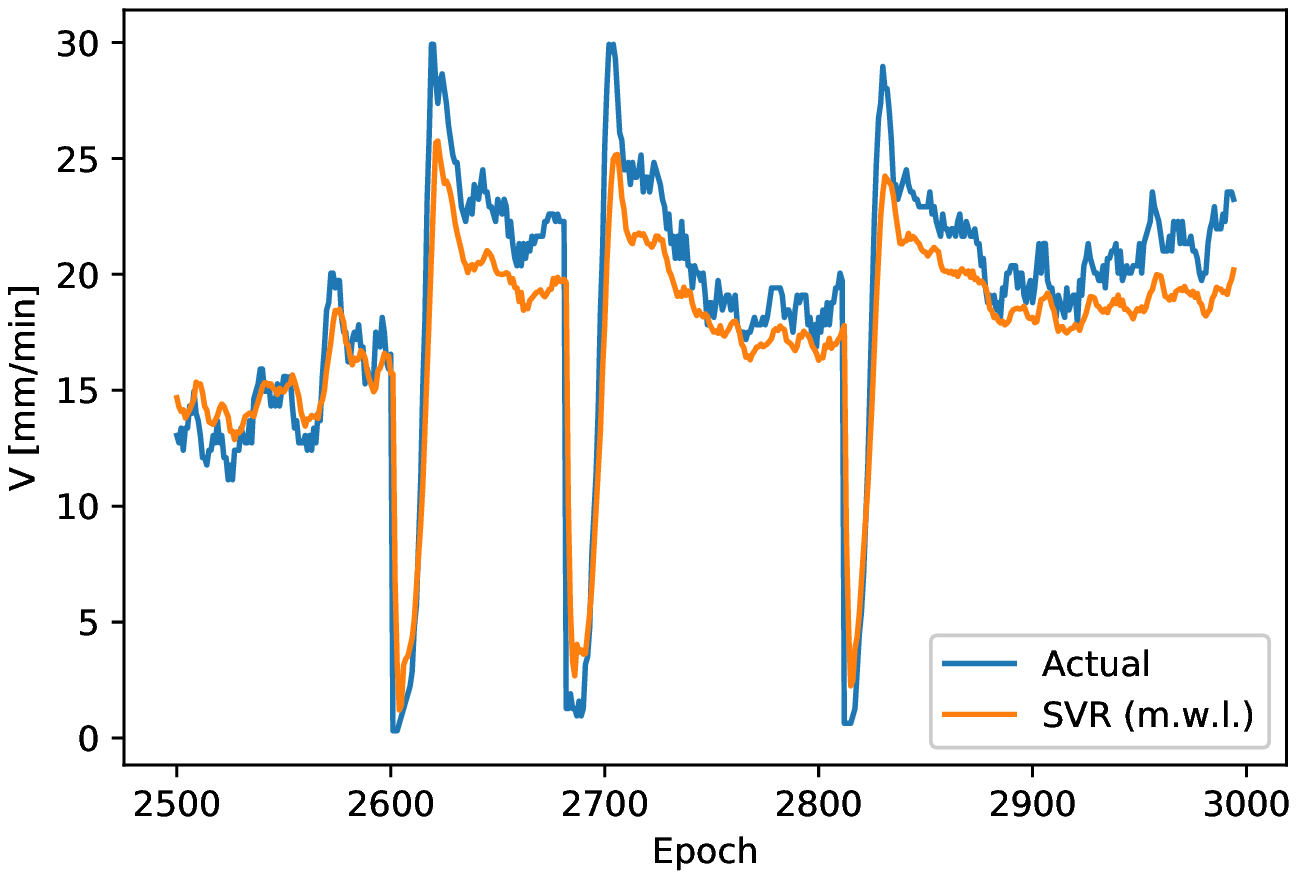}
\end{minipage}
}

\subfigure[RF (s.w/o.l.)]{
\begin{minipage}{0.22\textwidth}
\includegraphics[width=1.2\textwidth]{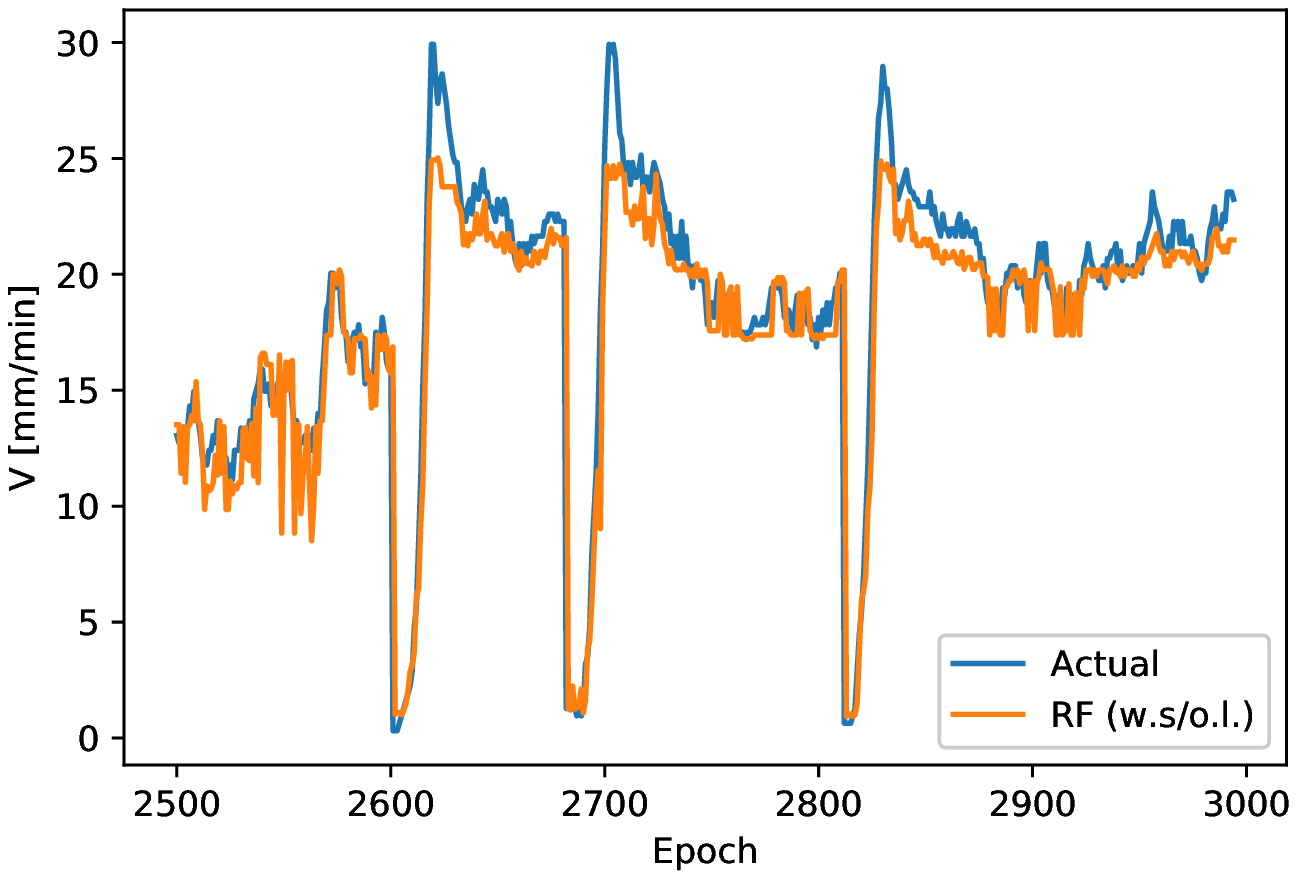}
\end{minipage}
}
\subfigure[RF (s.w.l.)]{
\begin{minipage}{0.22\textwidth}
\includegraphics[width=1.2\textwidth]{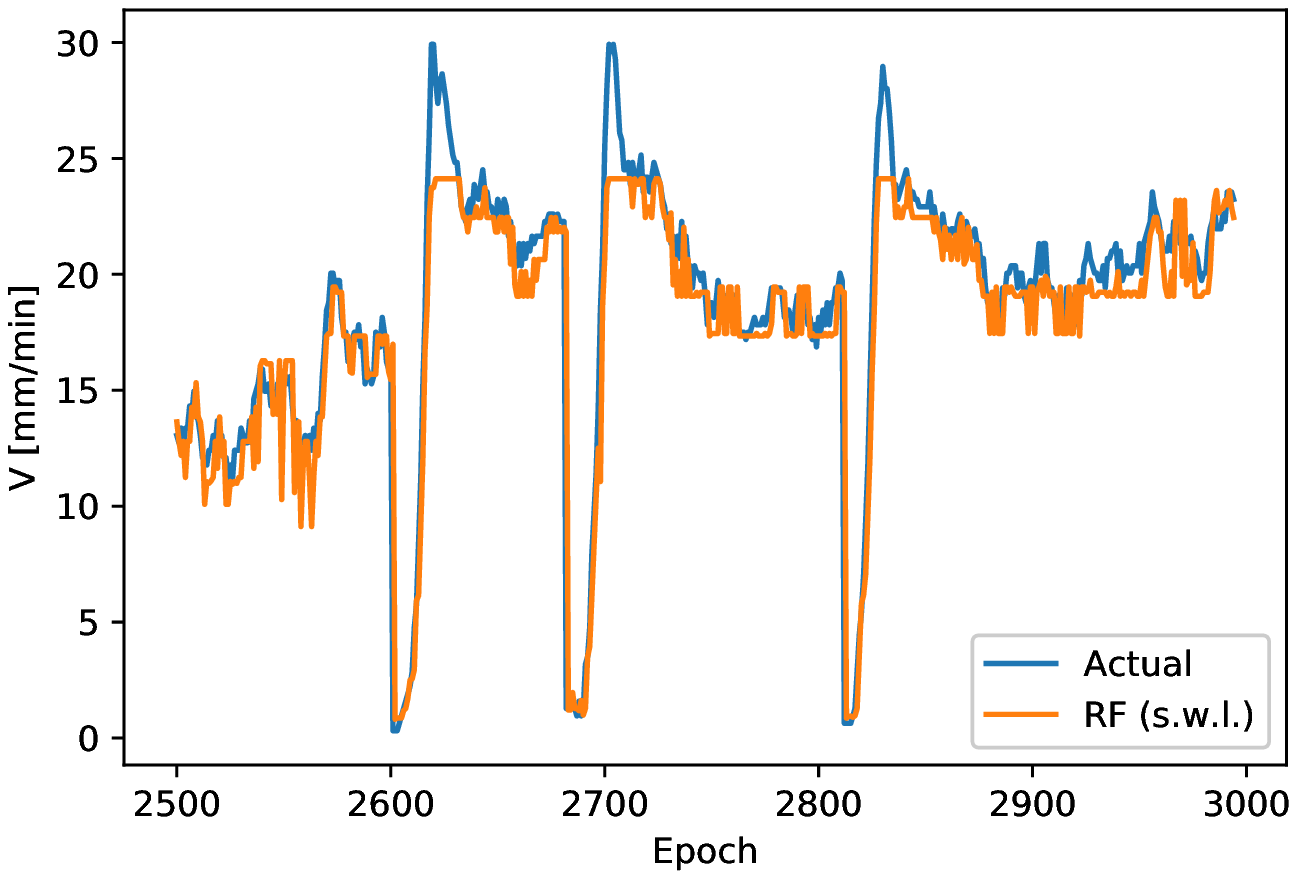}
\end{minipage}
}
\subfigure[RF (m.w/o.l.)]{
\begin{minipage}{0.22\textwidth}
\includegraphics[width=1.2\textwidth]{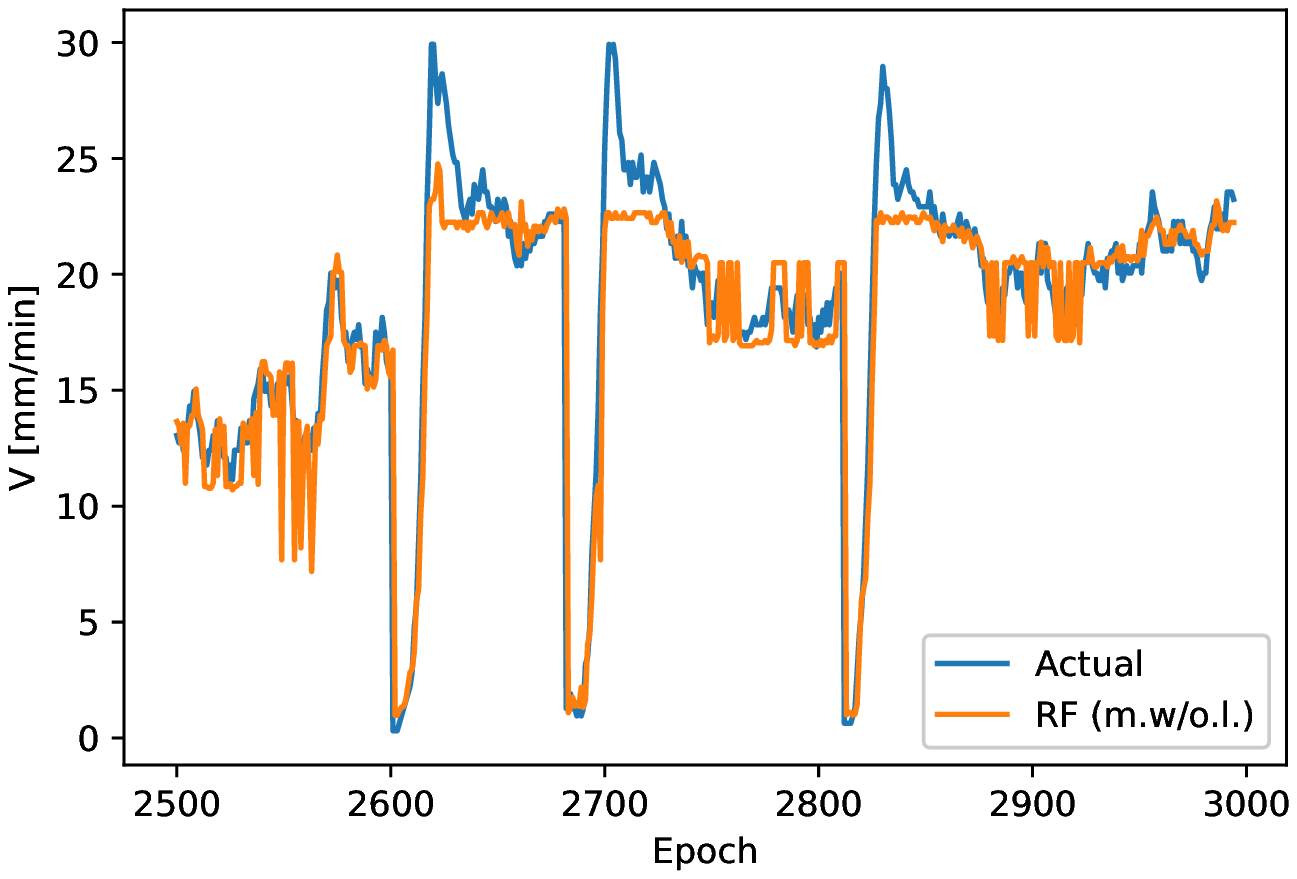}
\end{minipage}
}
\subfigure[RF (m.w.l.)]{
\begin{minipage}{0.22\textwidth}
\includegraphics[width=1.2\textwidth]{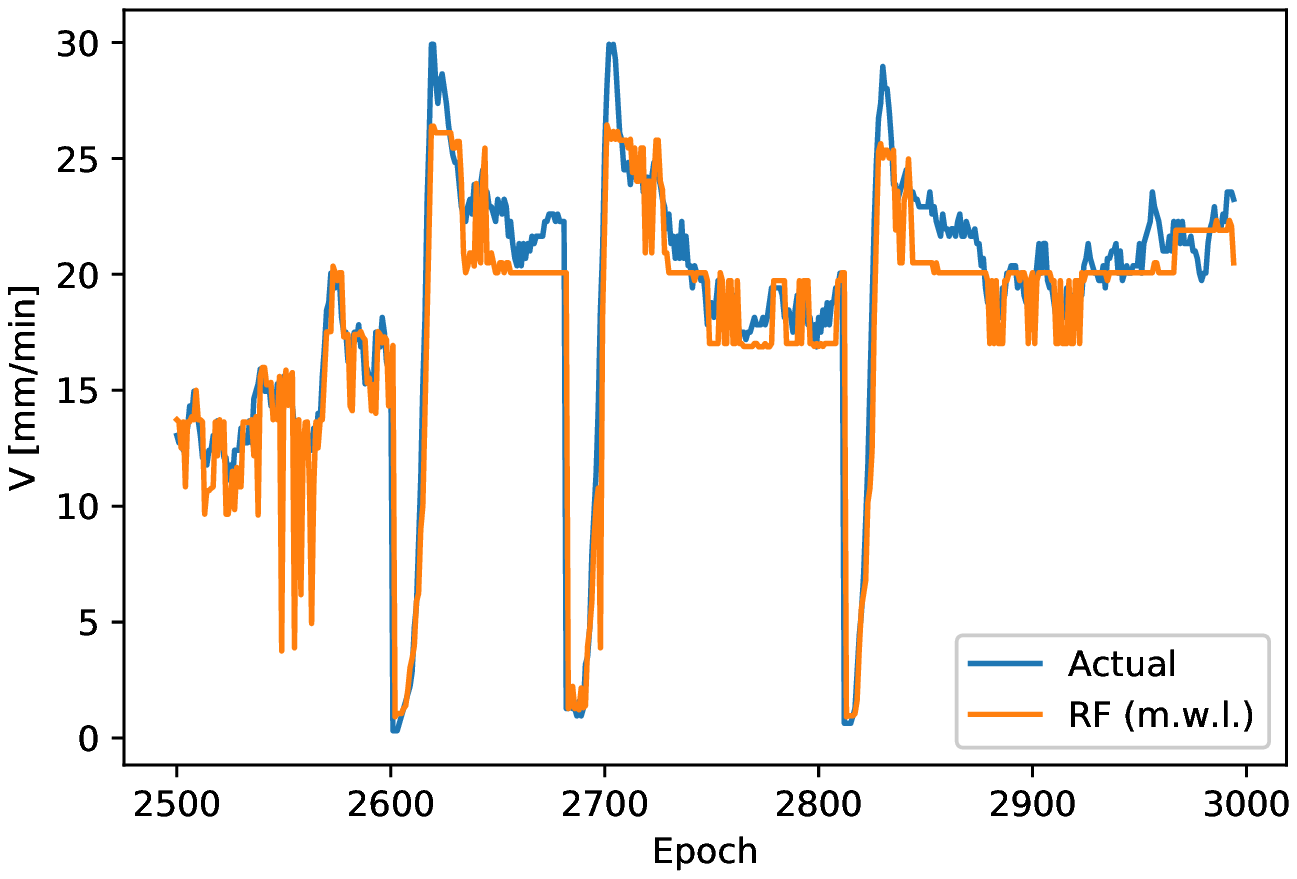}
\end{minipage}
}

\subfigure[FNN (s.w/o.l.)]{
\begin{minipage}{0.22\textwidth}
\includegraphics[width=1.2\textwidth]{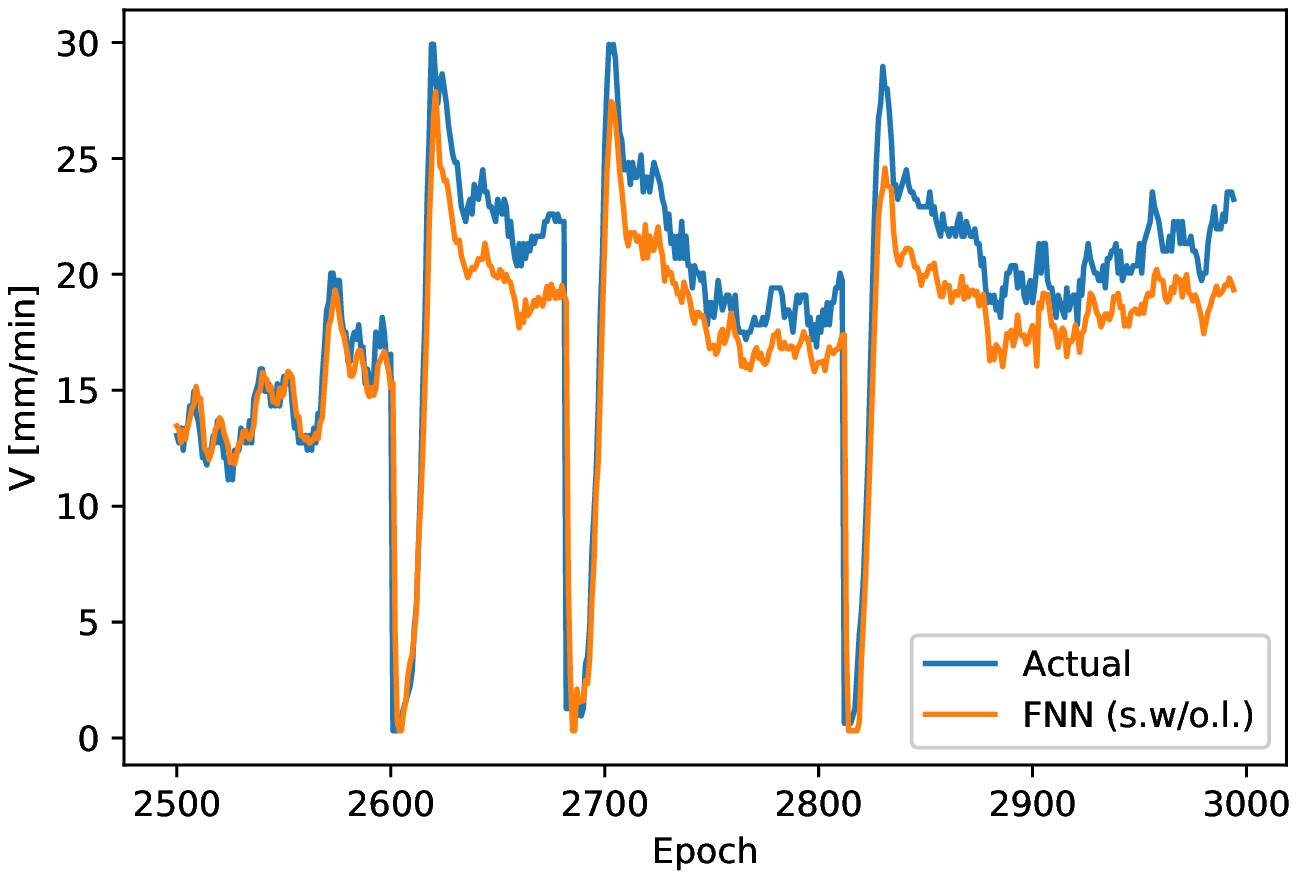}
\end{minipage}
}
\subfigure[FNN (s.w.l.)]{
\begin{minipage}{0.22\textwidth}
\includegraphics[width=1.2\textwidth]{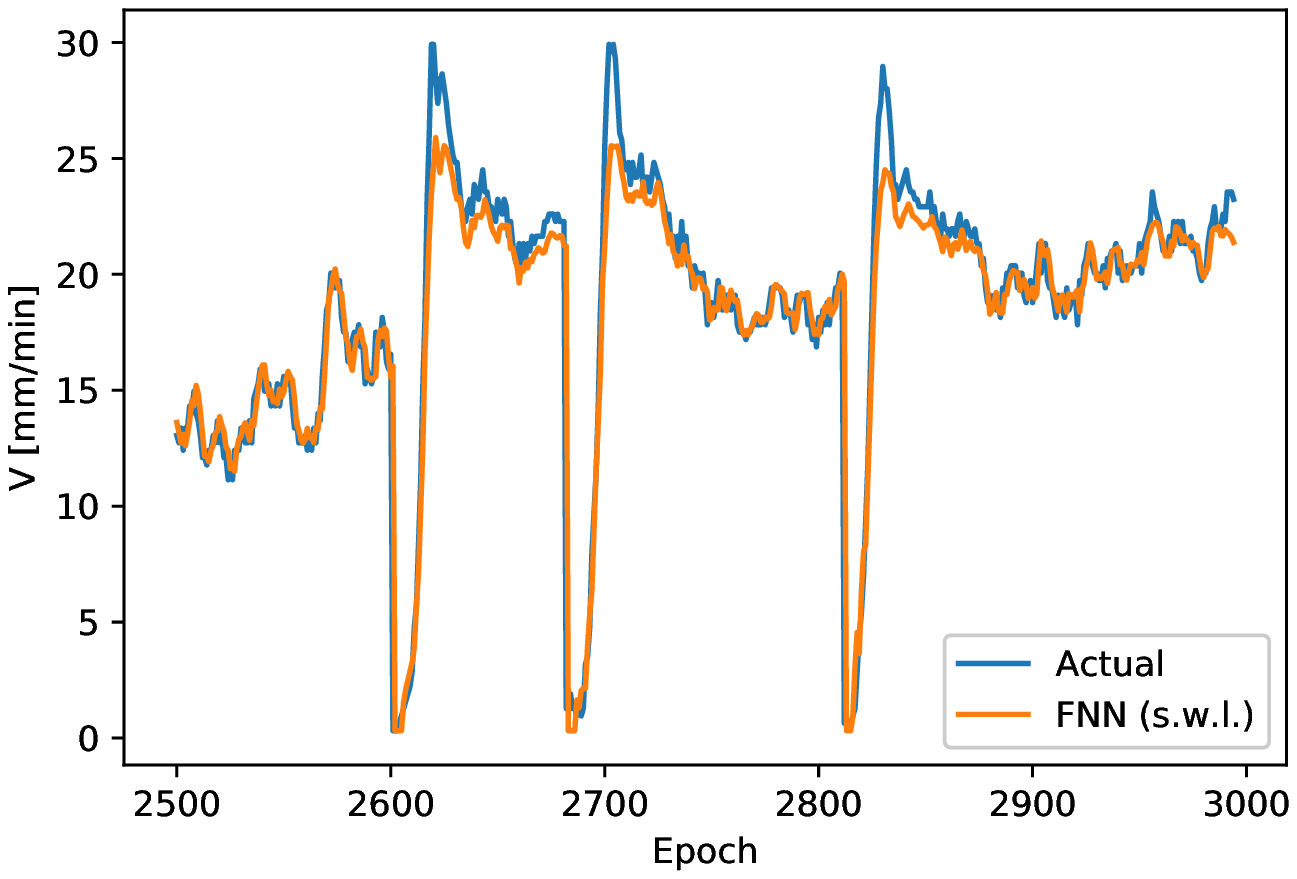}
\end{minipage}
}
\subfigure[FNN (m.w/o.l.)]{
\begin{minipage}{0.22\textwidth}
\includegraphics[width=1.2\textwidth]{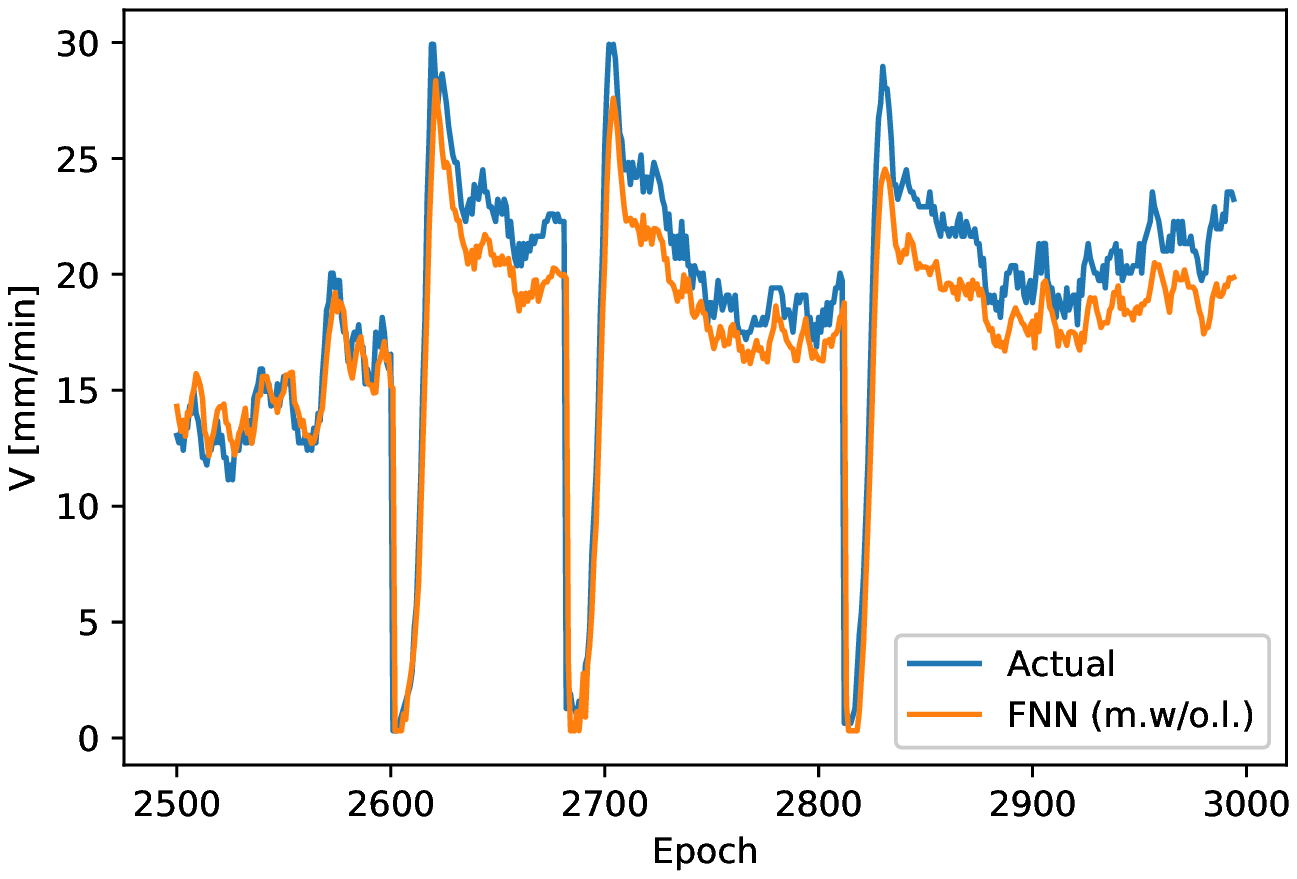}
\end{minipage}
}
\subfigure[FNN (m.w.l.)]{
\begin{minipage}{0.22\textwidth}
\includegraphics[width=1.2\textwidth]{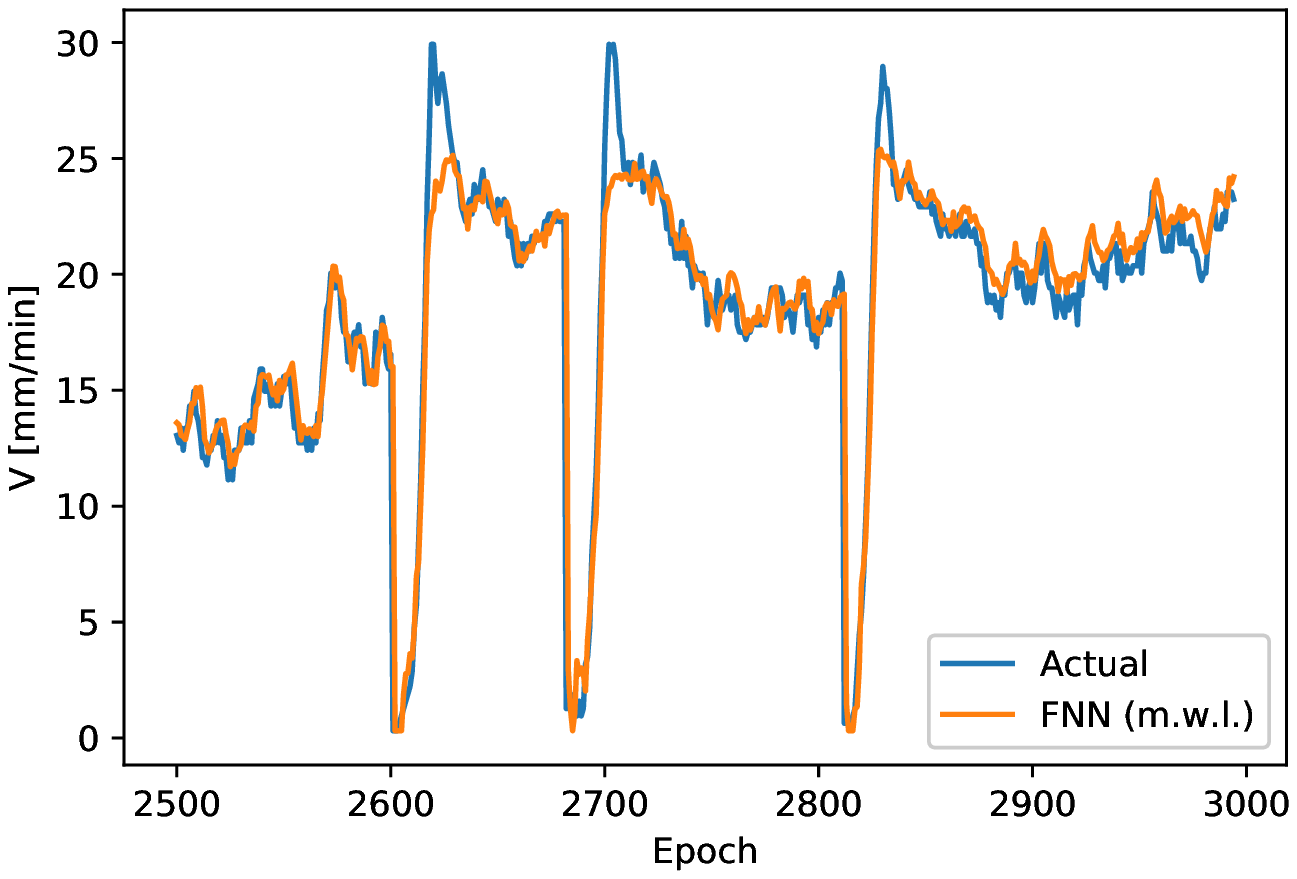}
\end{minipage}
}

\subfigure[RNN (s.w/o.l.)]{
\begin{minipage}{0.22\textwidth}
\includegraphics[width=1.2\textwidth]{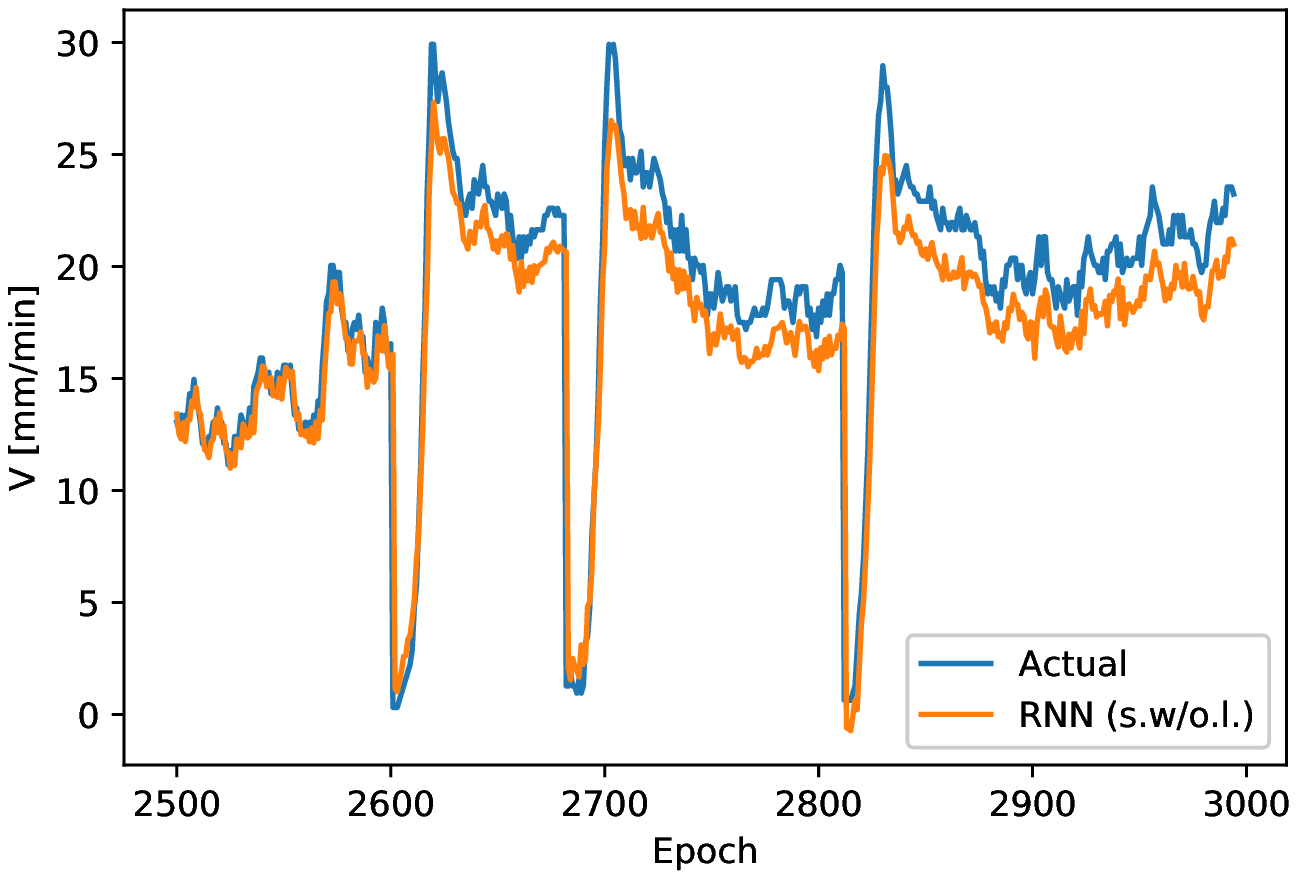}
\end{minipage}
}
\subfigure[RNN (s.w.l.)]{
\begin{minipage}{0.22\textwidth}
\includegraphics[width=1.2\textwidth]{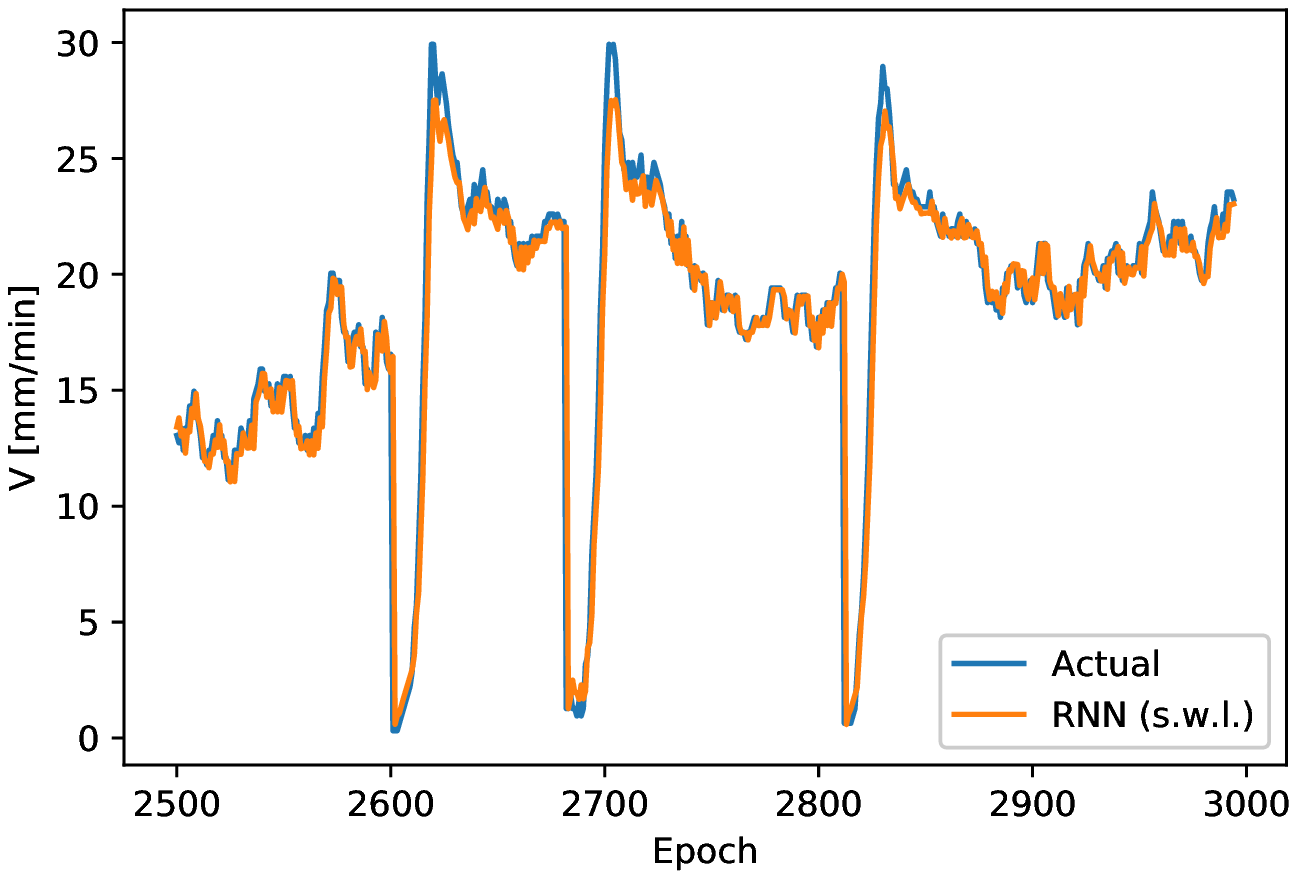}
\end{minipage}
}
\subfigure[RNN (m.w/o.l.)]{
\begin{minipage}{0.22\textwidth}
\includegraphics[width=1.2\textwidth]{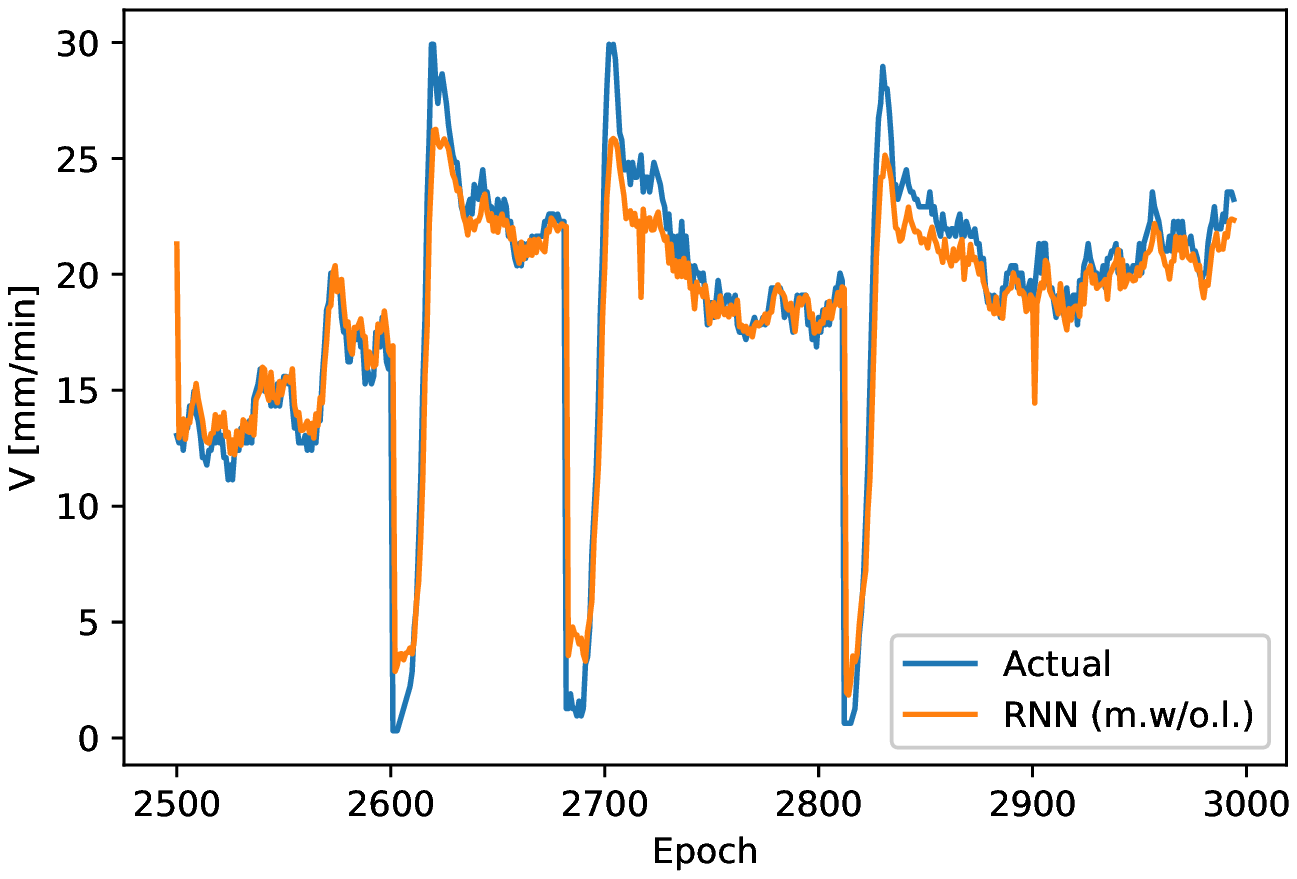}
\end{minipage}
}
\subfigure[RNN (m.w.l.)]{
\begin{minipage}{0.22\textwidth}
\includegraphics[width=1.2\textwidth]{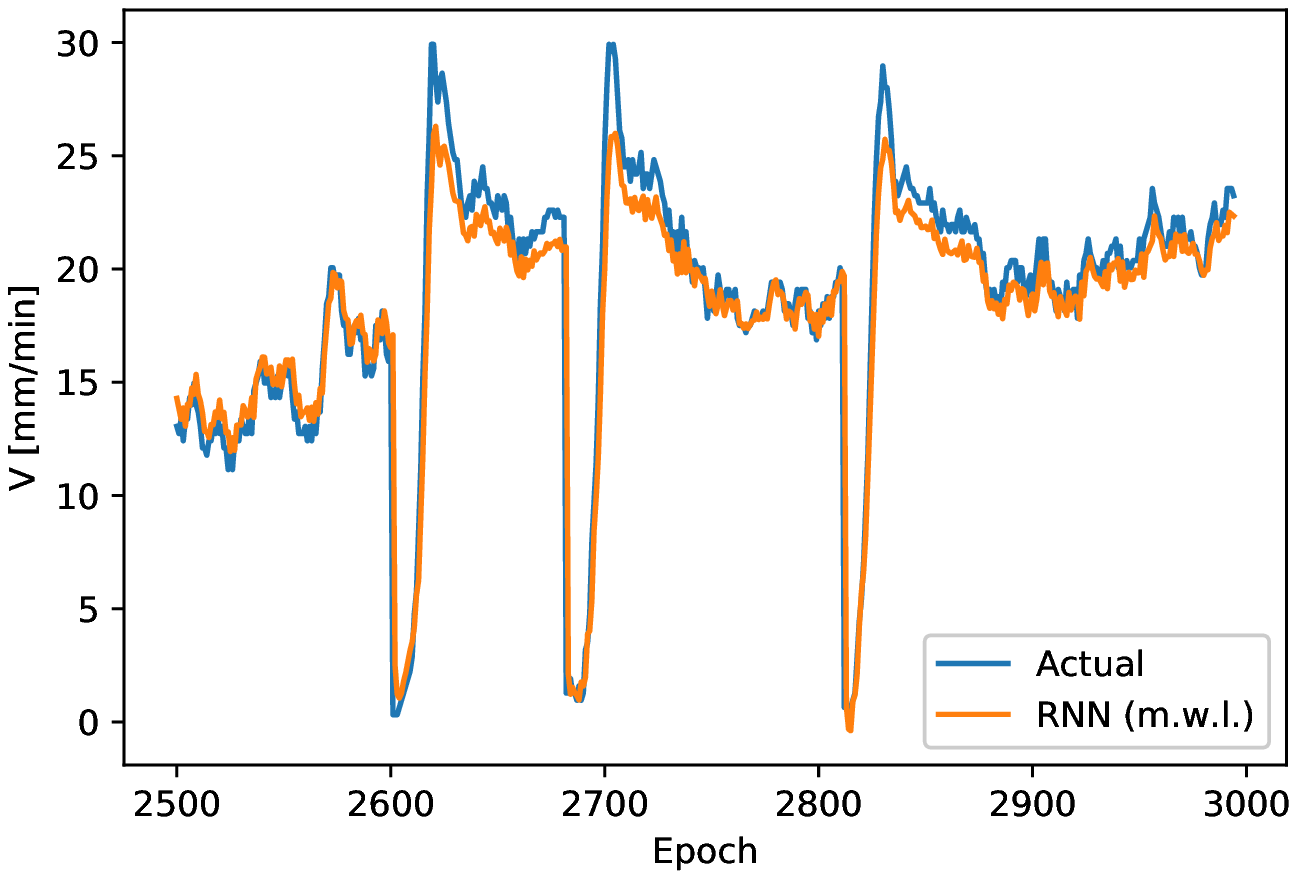}
\end{minipage}
}

\subfigure[GRU (s.w/o.l.)]{
\begin{minipage}{0.22\textwidth}
\includegraphics[width=1.2\textwidth]{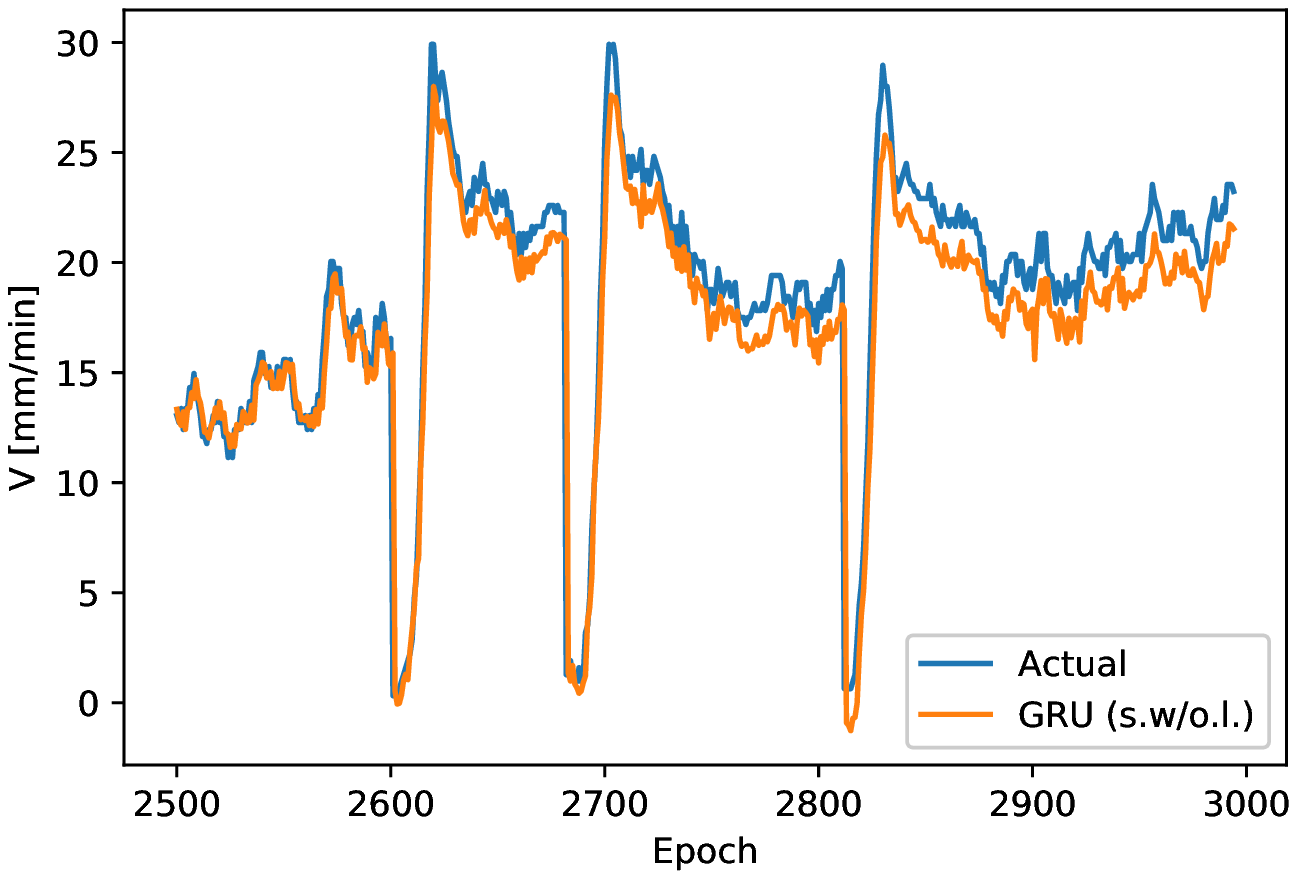}
\end{minipage}
}
\subfigure[GRU (s.w.l.)]{
\begin{minipage}{0.22\textwidth}
\includegraphics[width=1.2\textwidth]{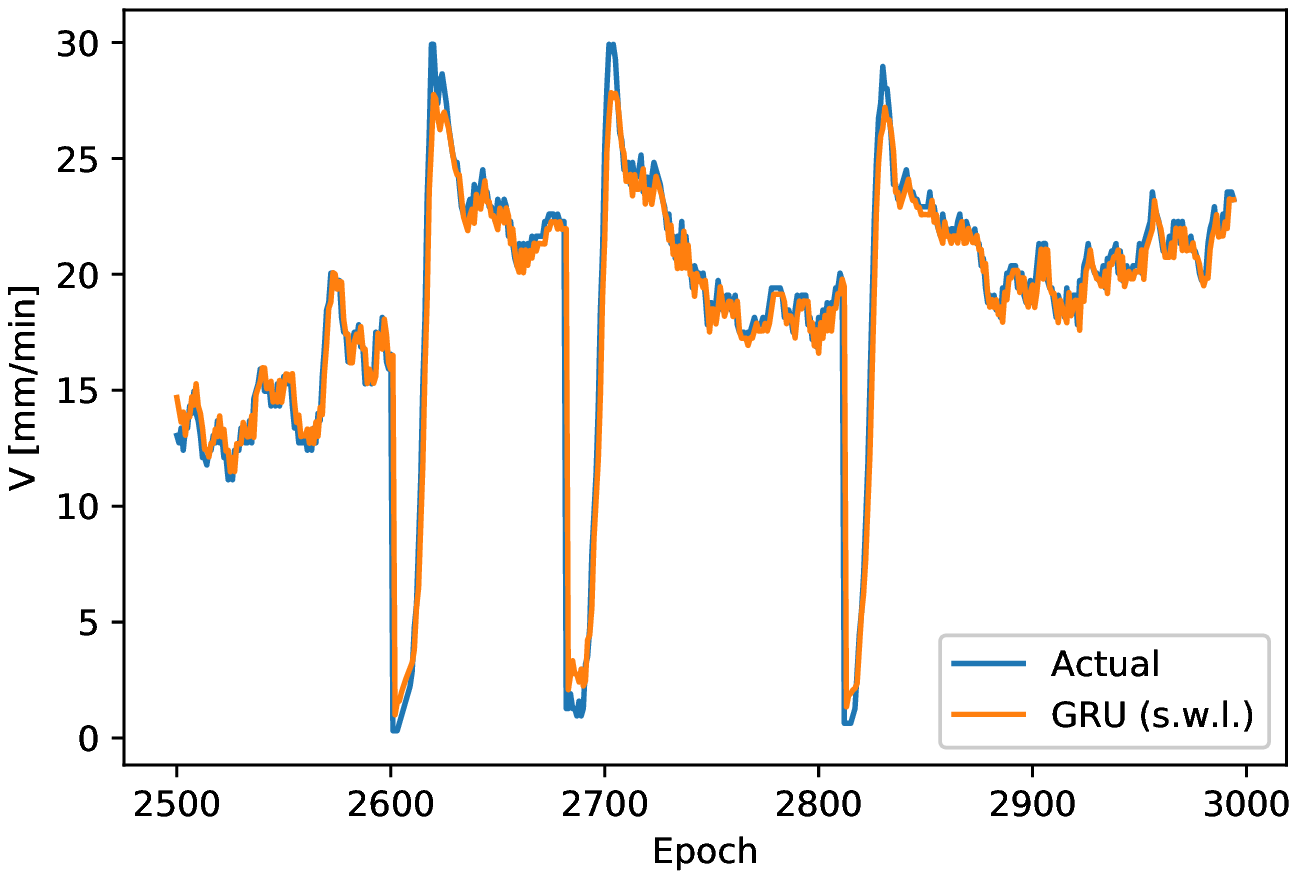}
\end{minipage}
}
\subfigure[GRU (m.w/o.l.)]{
\begin{minipage}{0.22\textwidth}
\includegraphics[width=1.2\textwidth]{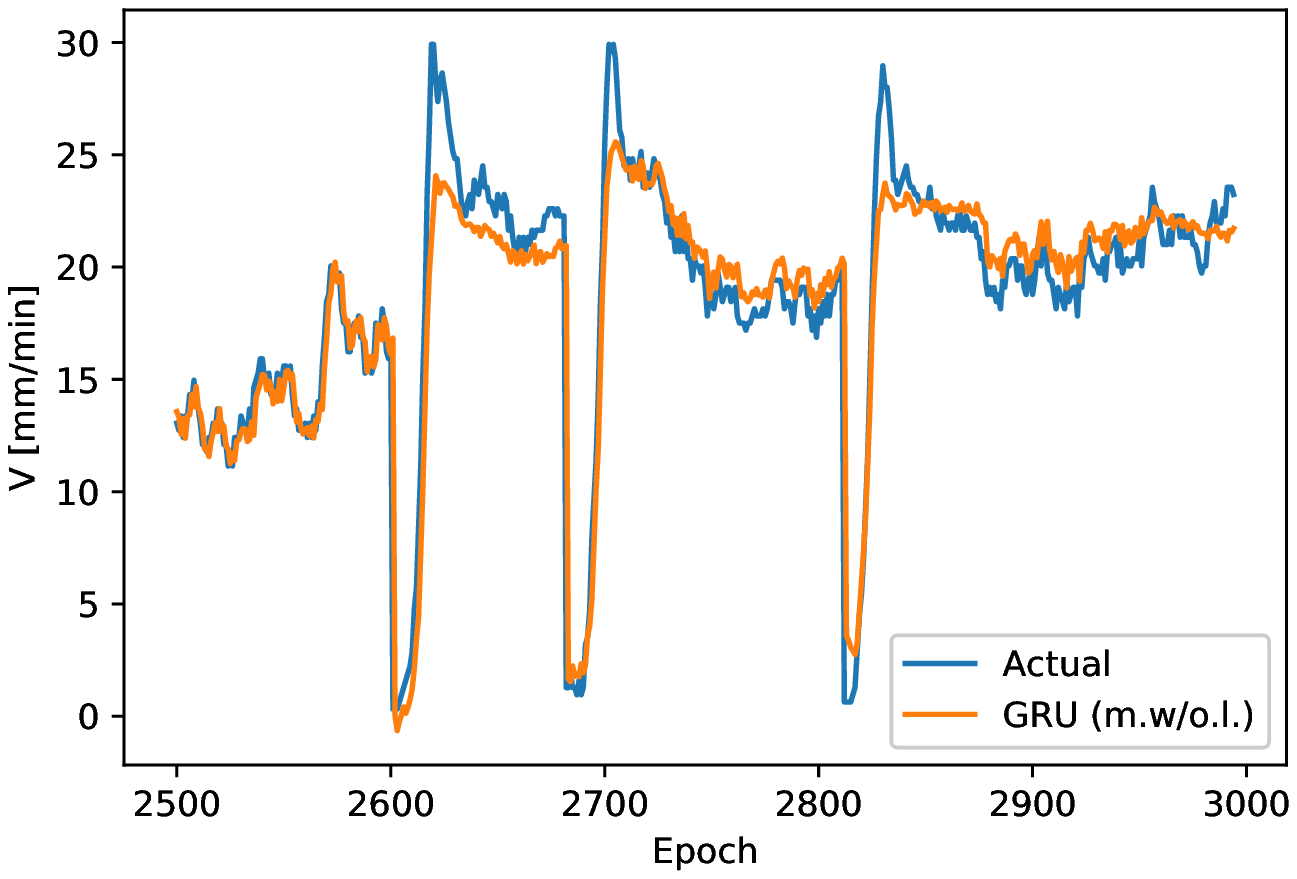}
\end{minipage}
}
\subfigure[GRU (m.w.l.)]{
\begin{minipage}{0.22\textwidth}
\includegraphics[width=1.2\textwidth]{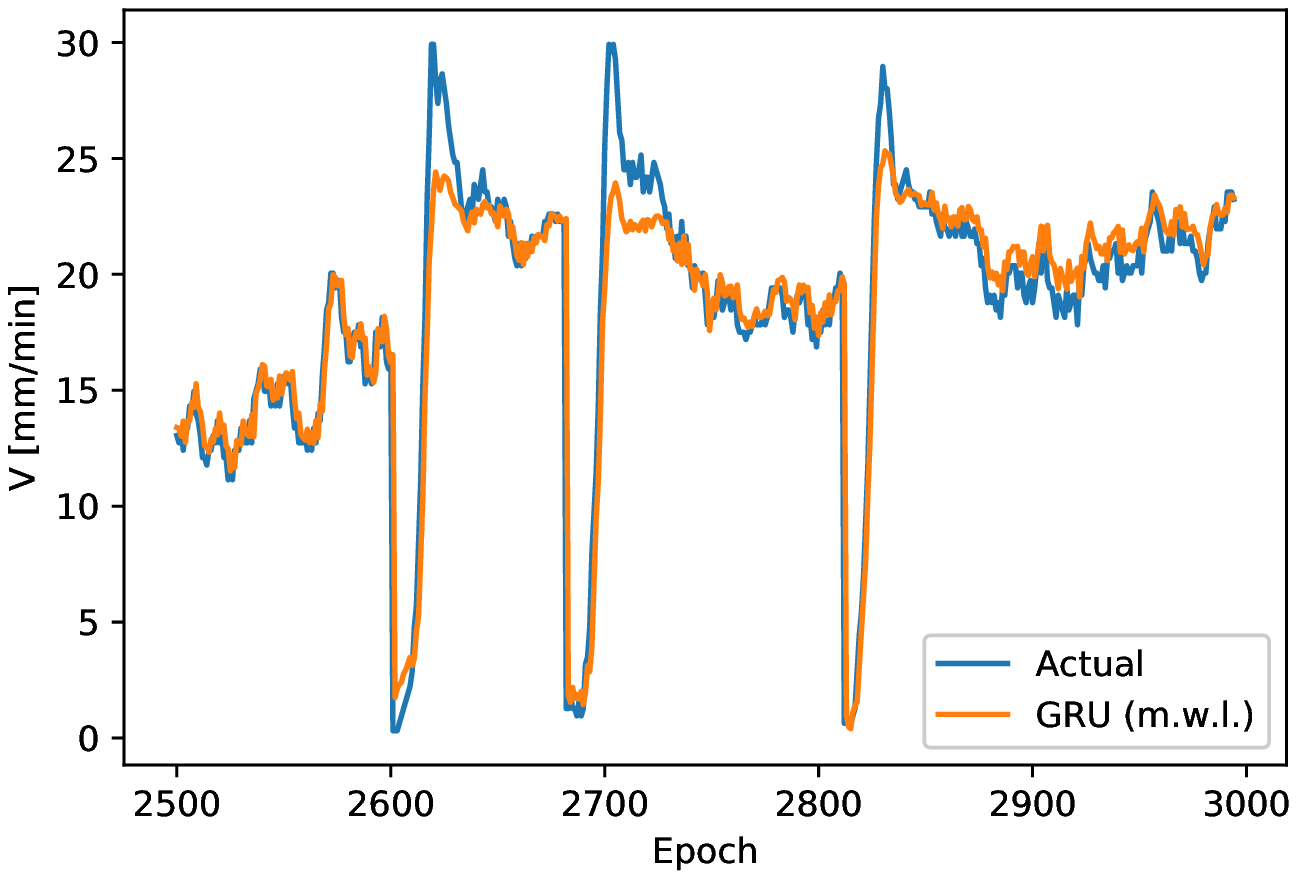}
\end{minipage}
}

\subfigure[LSTM (s.w/o.l.)]{
\begin{minipage}{0.22\textwidth}
\includegraphics[width=1.2\textwidth]{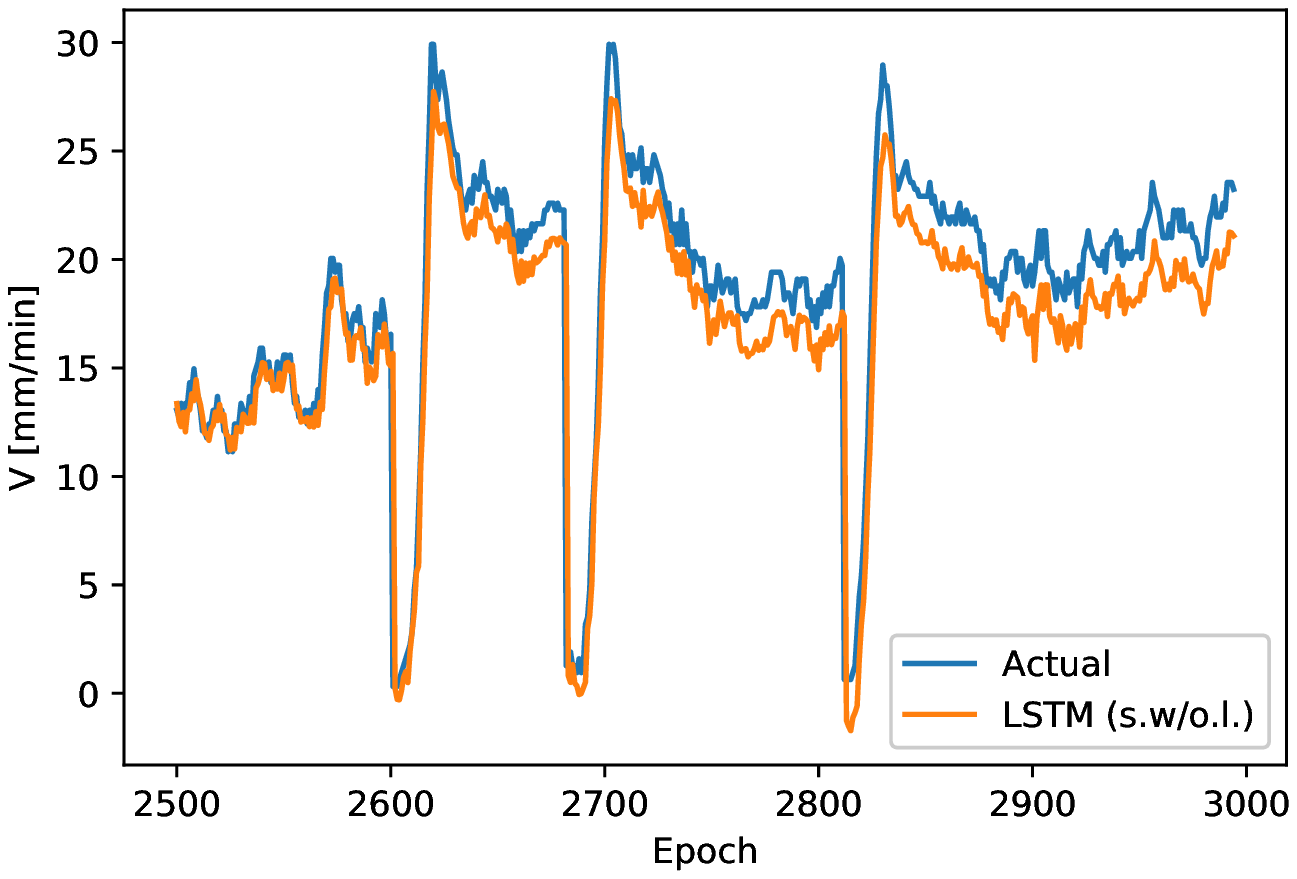}
\end{minipage}
}
\subfigure[LSTM (s.w.l.)]{
\begin{minipage}{0.22\textwidth}
\includegraphics[width=1.2\textwidth]{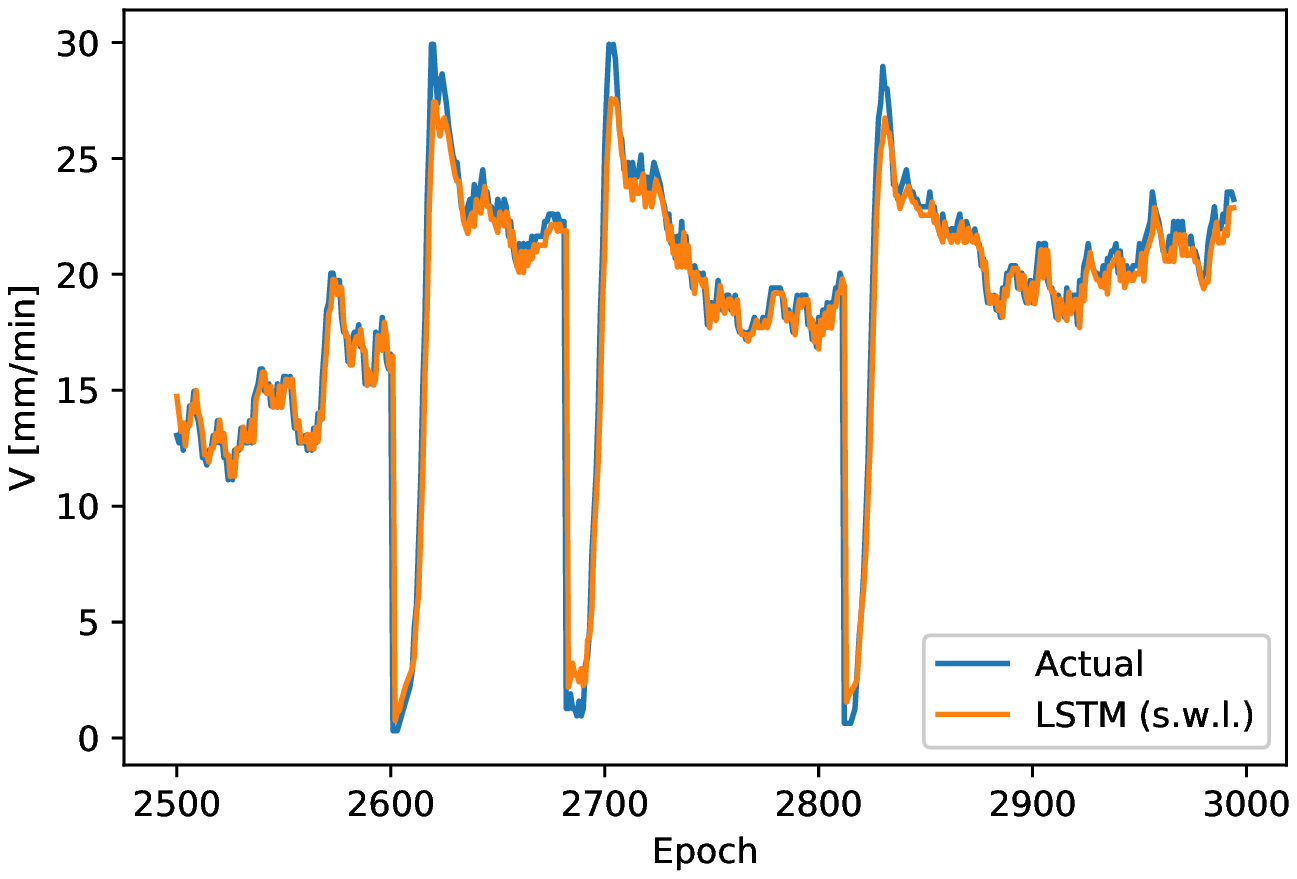}
\end{minipage}
}
\subfigure[LSTM (m.w/o.l.)]{
\begin{minipage}{0.22\textwidth}
\includegraphics[width=1.2\textwidth]{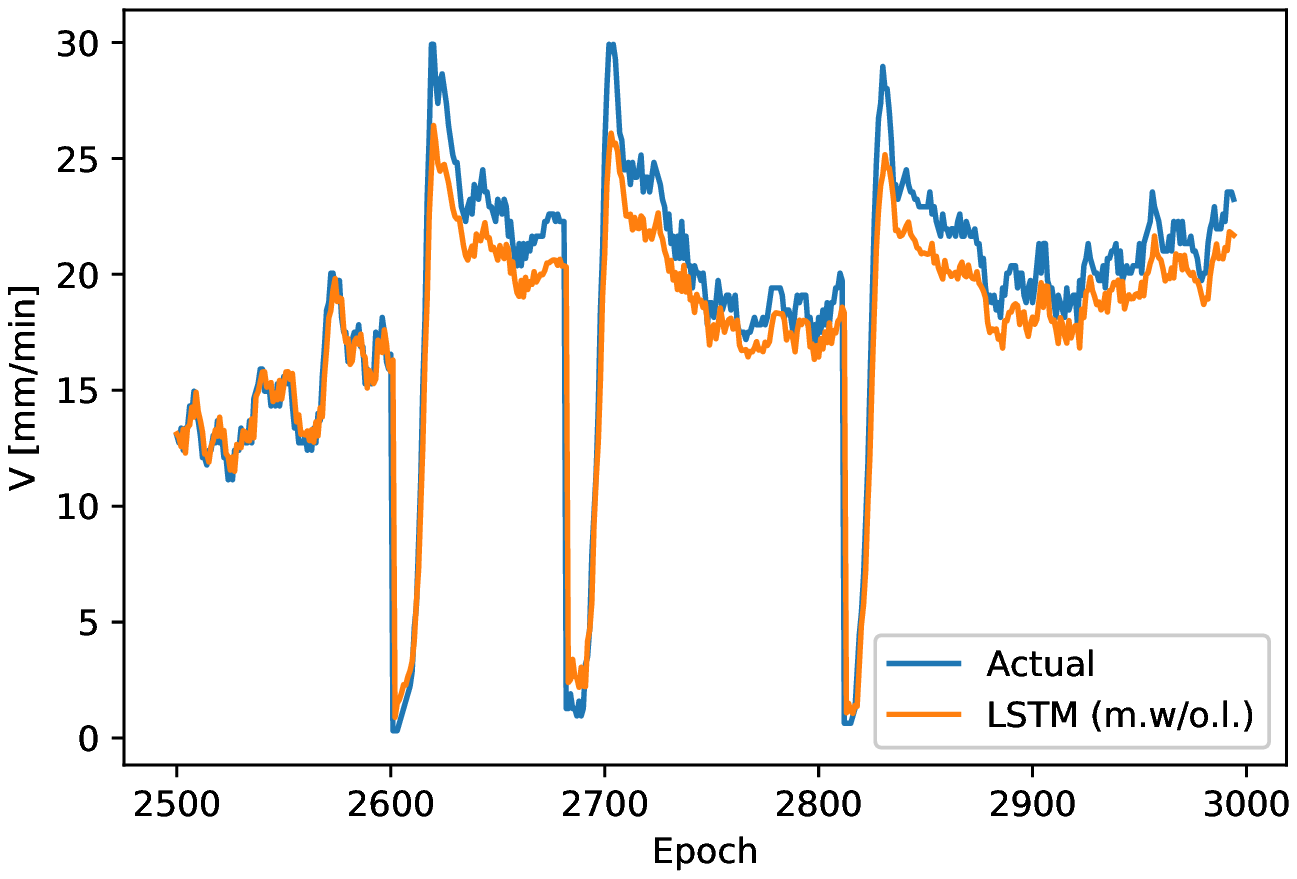}
\end{minipage}
}
\subfigure[LSTM (m.w.l.)]{
\begin{minipage}{0.22\textwidth}
\includegraphics[width=1.2\textwidth]{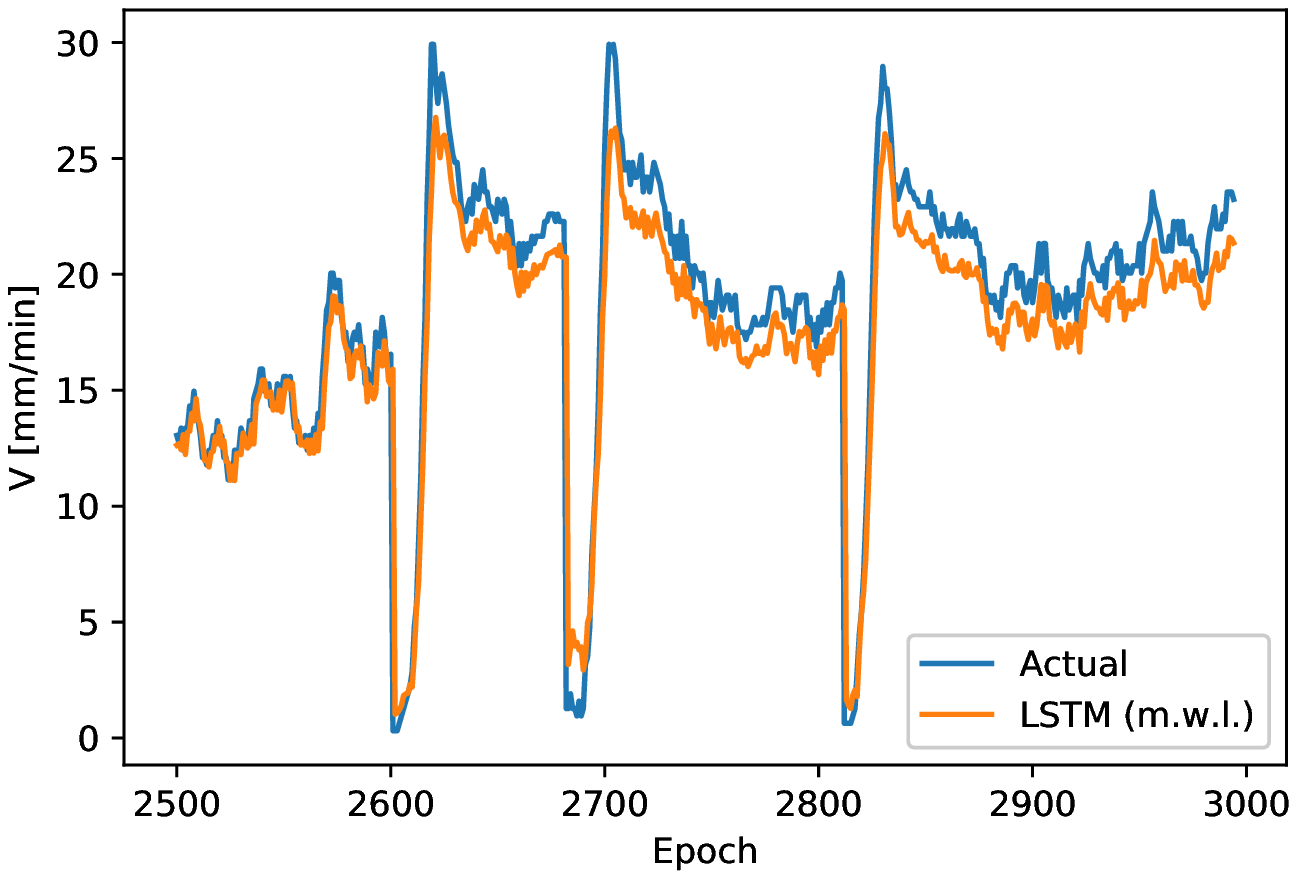}
\end{minipage}
}
\caption{Forecast Results of ``Advance Rate"} \label{fig:ar}
\end{figure}

\begin{figure}[htbp]
\centering
\subfigure[SVR (s.w/o.l.)]{
\begin{minipage}{0.22\textwidth}
\includegraphics[width=1.2\textwidth]{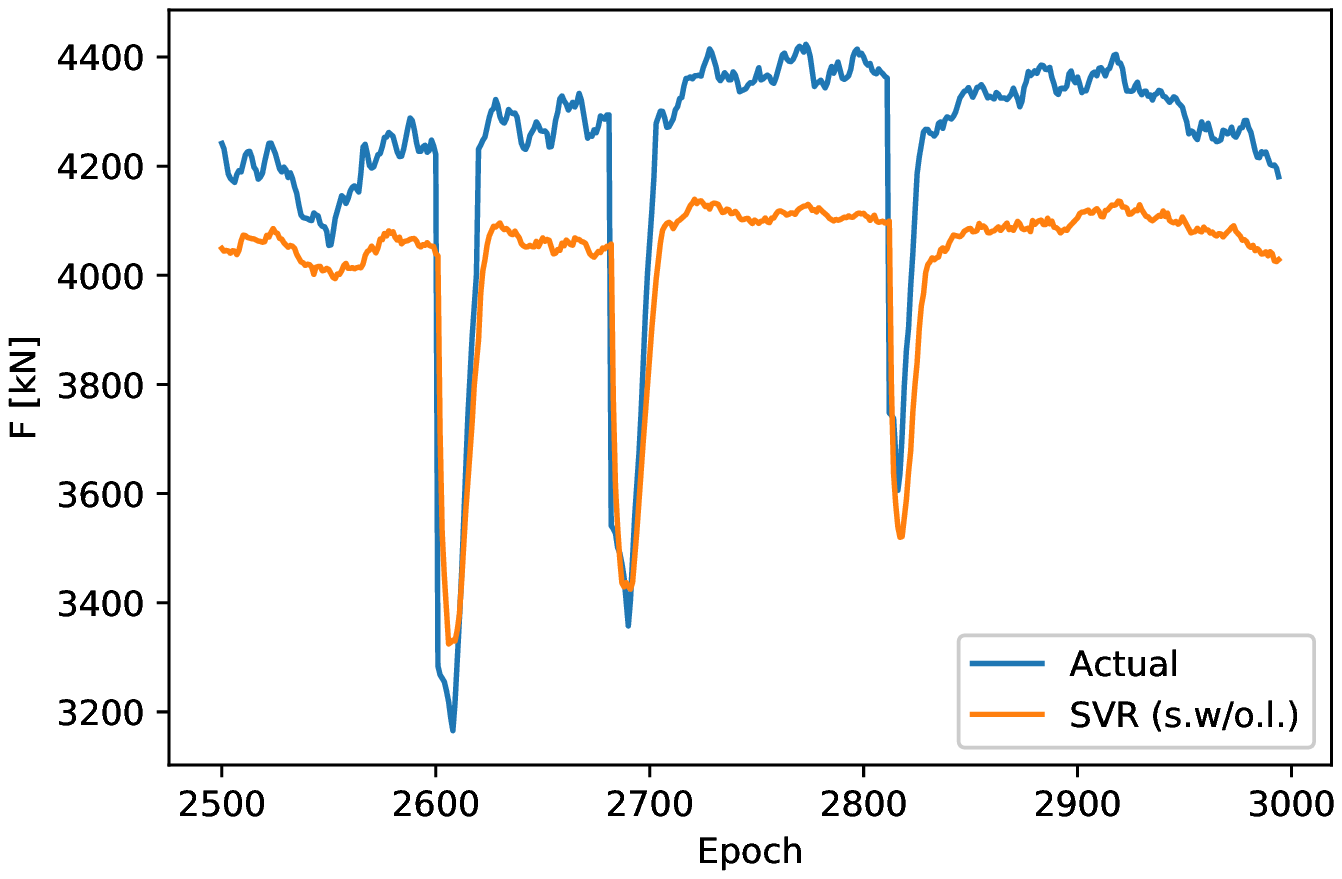}
\end{minipage}
}
\subfigure[SVR (s.w.l.)]{
\begin{minipage}{0.22\textwidth}
\includegraphics[width=1.2\textwidth]{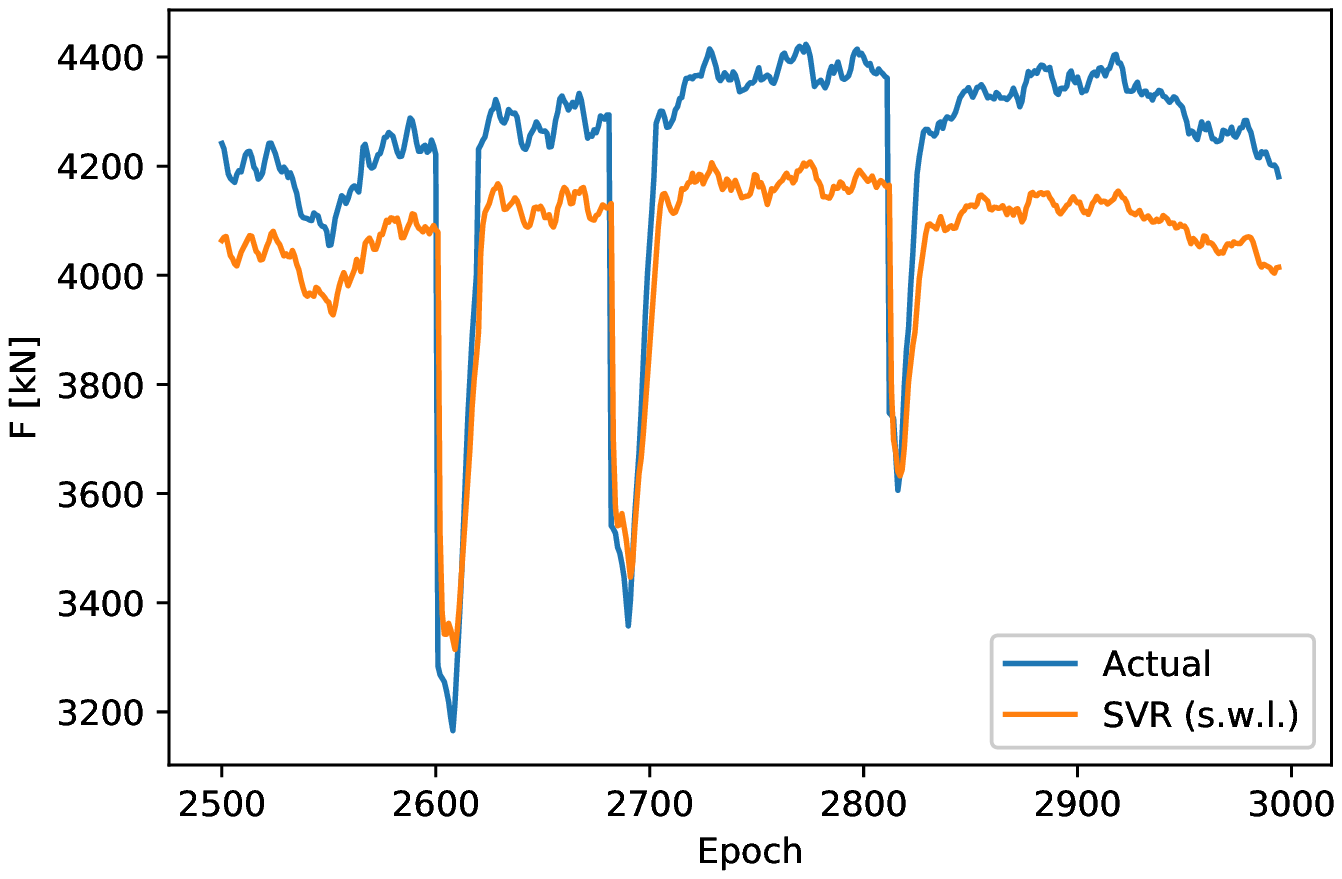}
\end{minipage}
}
\subfigure[SVR (m.w/o.l.)]{
\begin{minipage}{0.22\textwidth}
\includegraphics[width=1.2\textwidth]{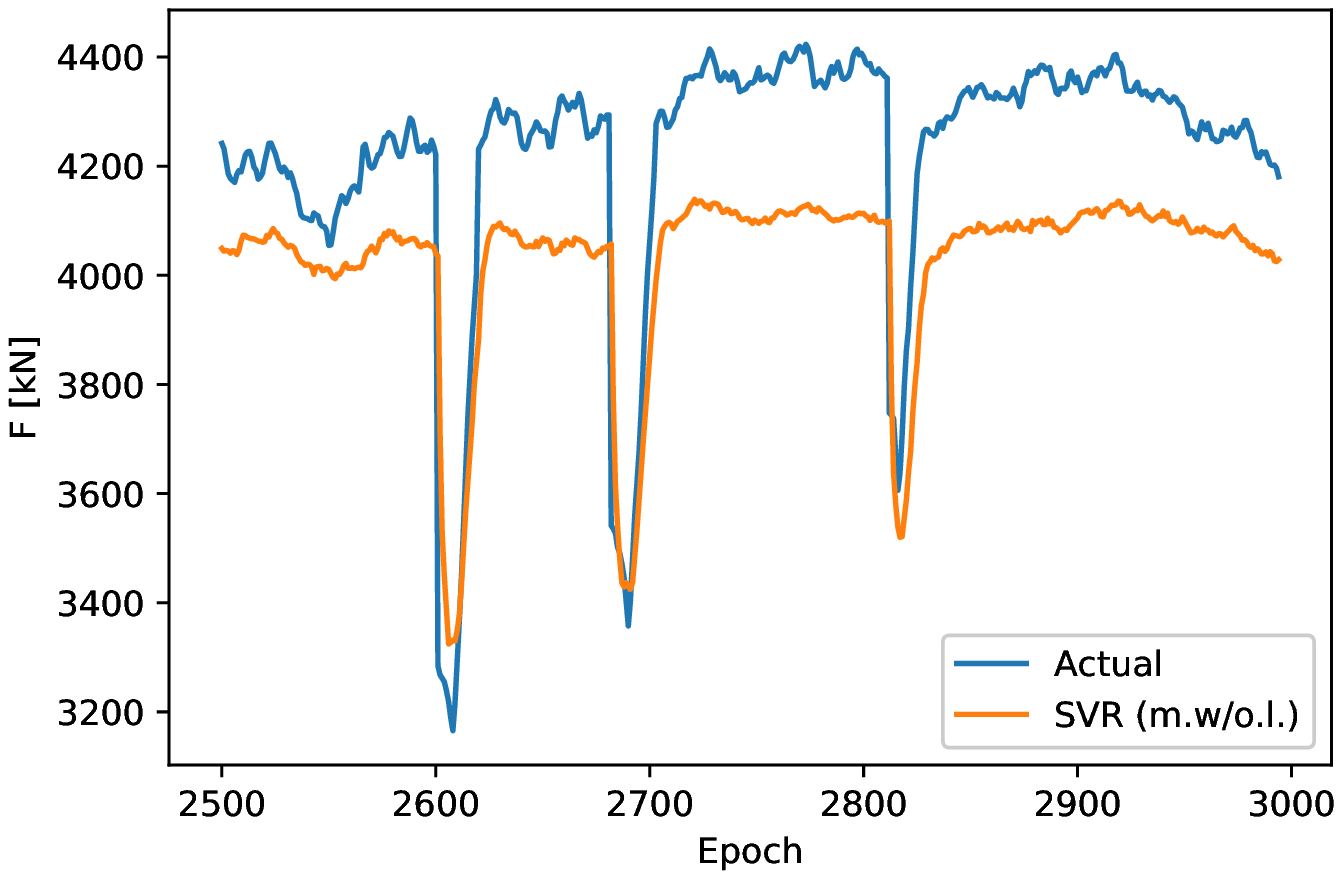}
\end{minipage}
}
\subfigure[SVR (m.w.l.)]{
\begin{minipage}{0.22\textwidth}
\includegraphics[width=1.2\textwidth]{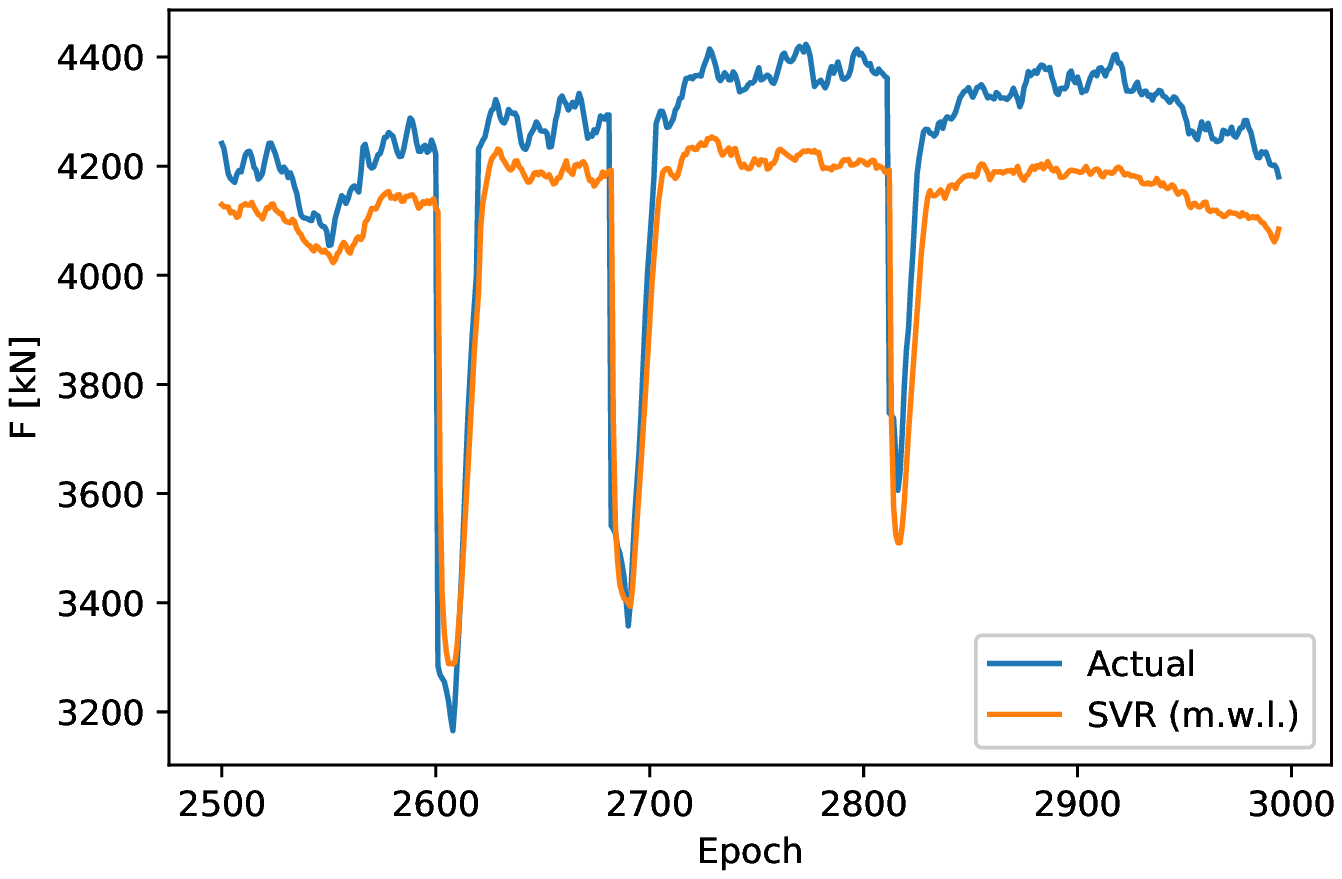}
\end{minipage}
}

\subfigure[RF (s.w/o.l.)]{
\begin{minipage}{0.22\textwidth}
\includegraphics[width=1.2\textwidth]{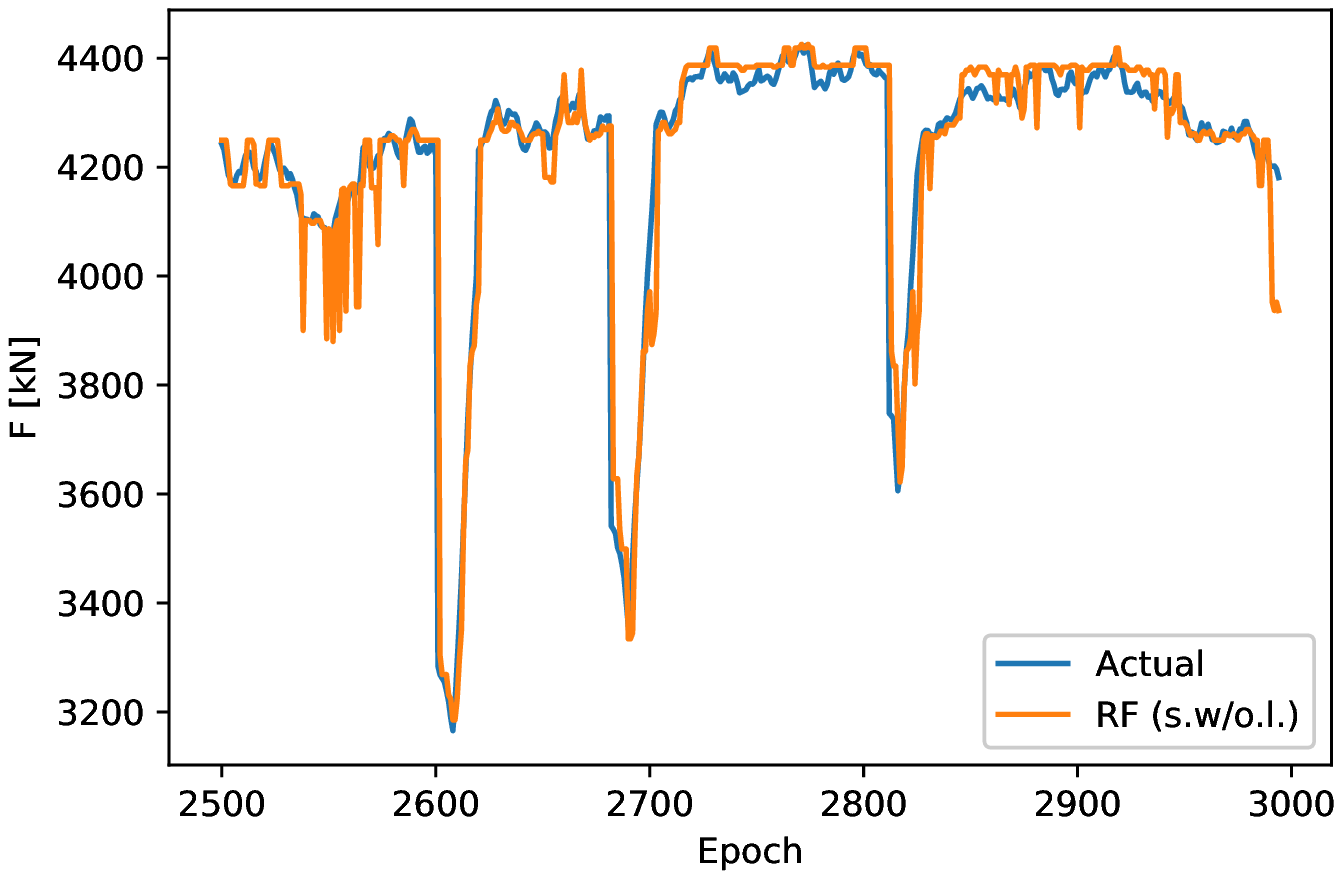}
\end{minipage}
}
\subfigure[RF (s.w.l.)]{
\begin{minipage}{0.22\textwidth}
\includegraphics[width=1.2\textwidth]{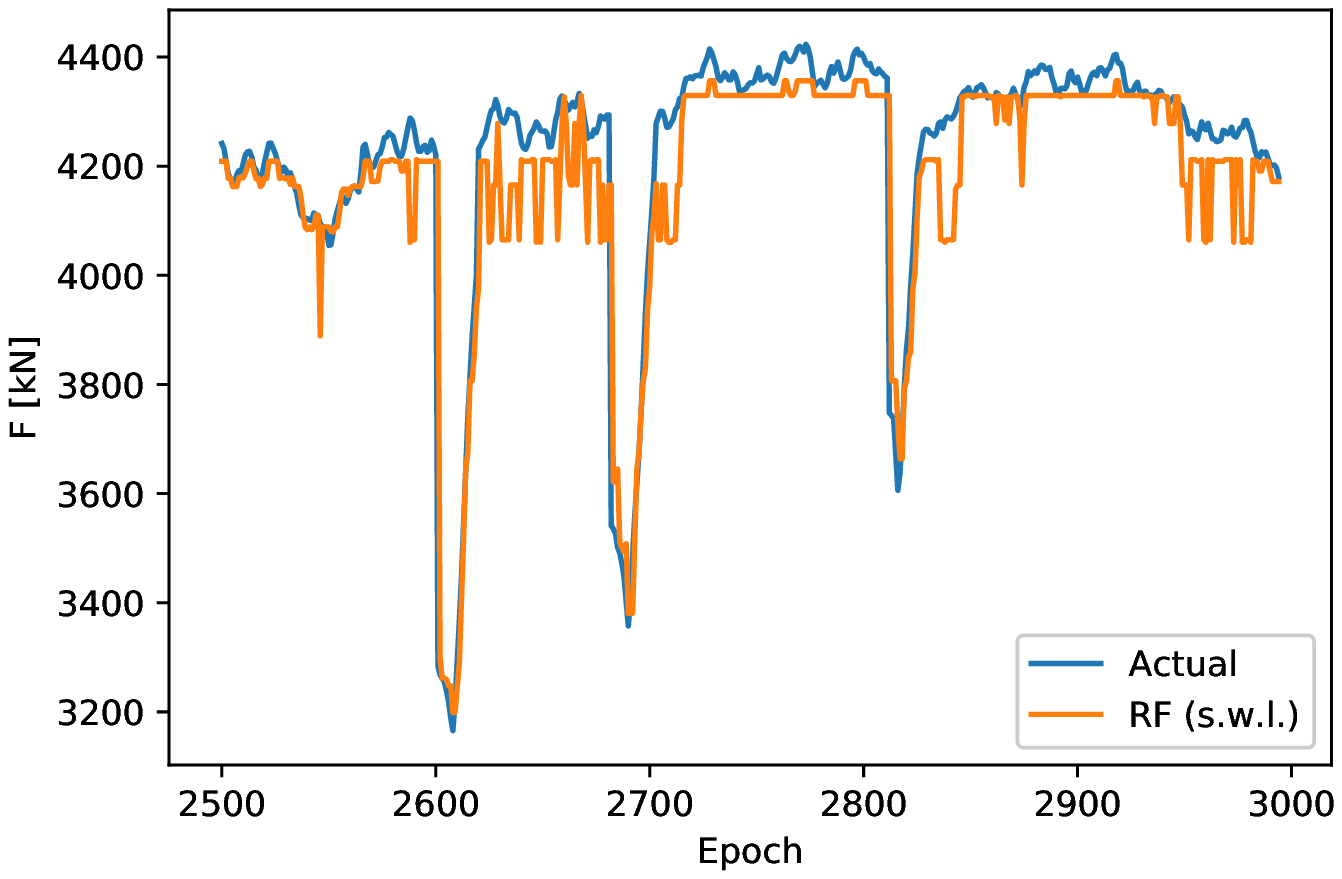}
\end{minipage}
}
\subfigure[RF (m.w/o.l.)]{
\begin{minipage}{0.22\textwidth}
\includegraphics[width=1.2\textwidth]{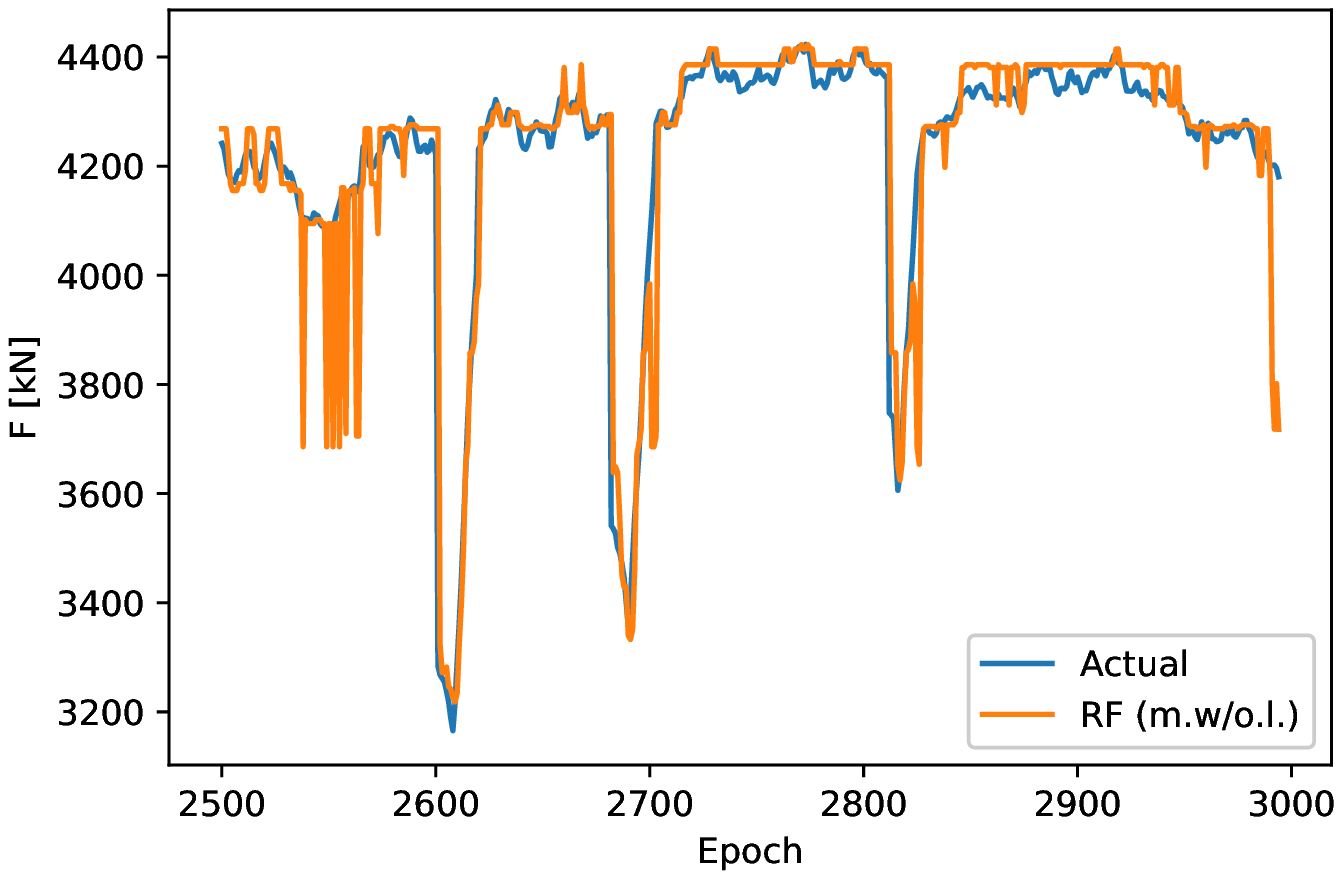}
\end{minipage}
}
\subfigure[RF (m.w.l.)]{
\begin{minipage}{0.22\textwidth}
\includegraphics[width=1.2\textwidth]{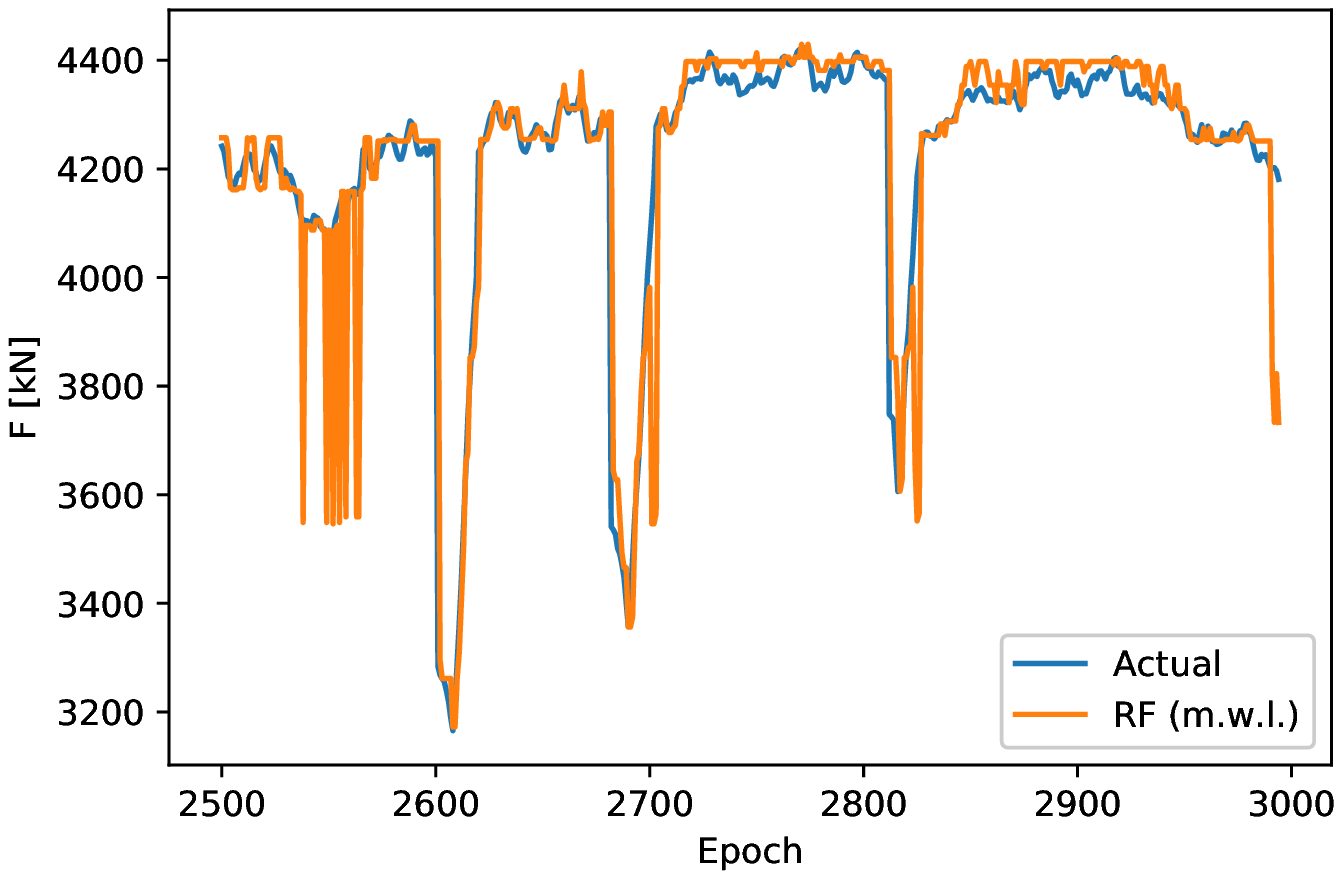}
\end{minipage}
}

\subfigure[FNN (s.w/o.l.)]{
\begin{minipage}{0.22\textwidth}
\includegraphics[width=1.2\textwidth]{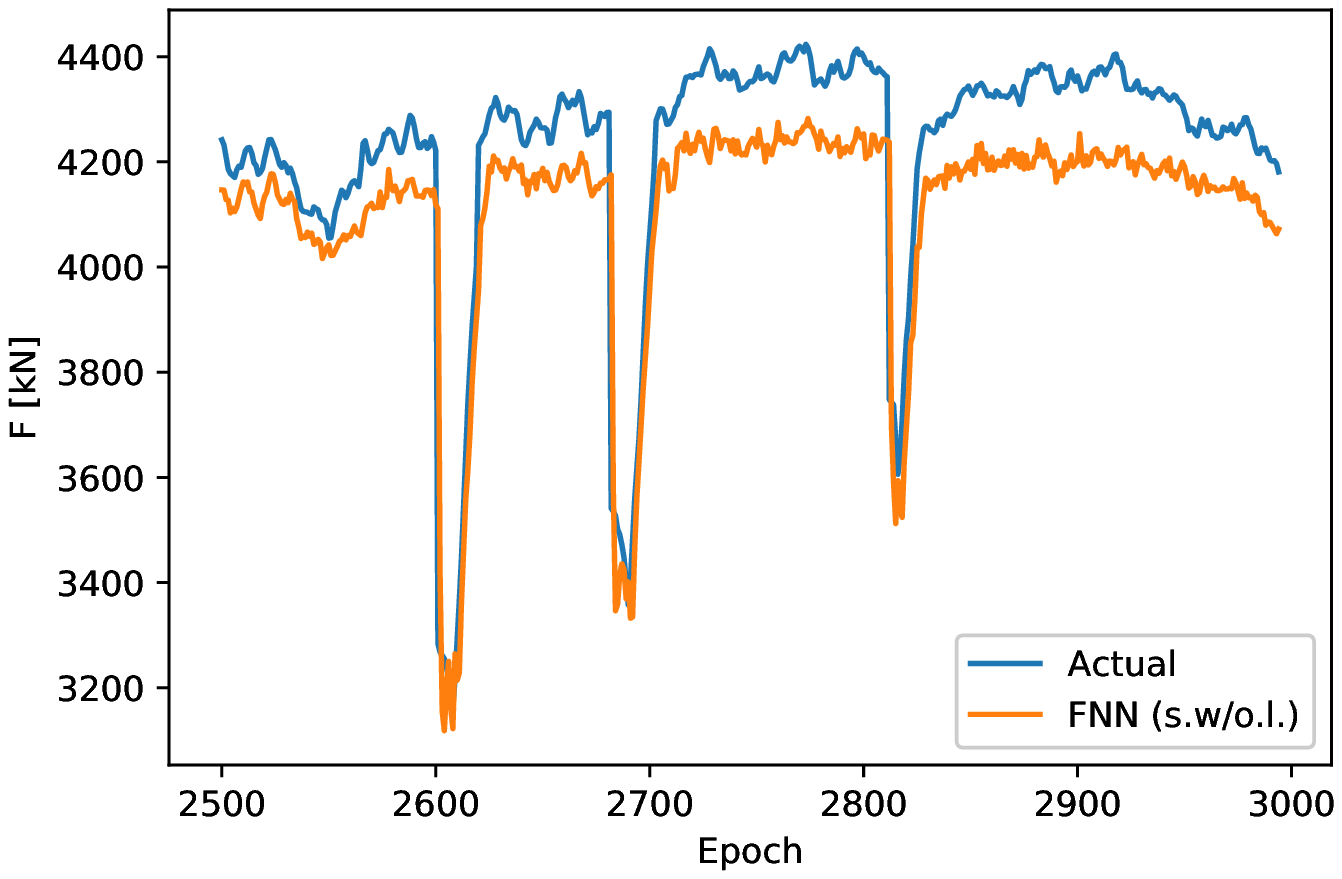}
\end{minipage}
}
\subfigure[FNN (s.w.l.)]{
\begin{minipage}{0.22\textwidth}
\includegraphics[width=1.2\textwidth]{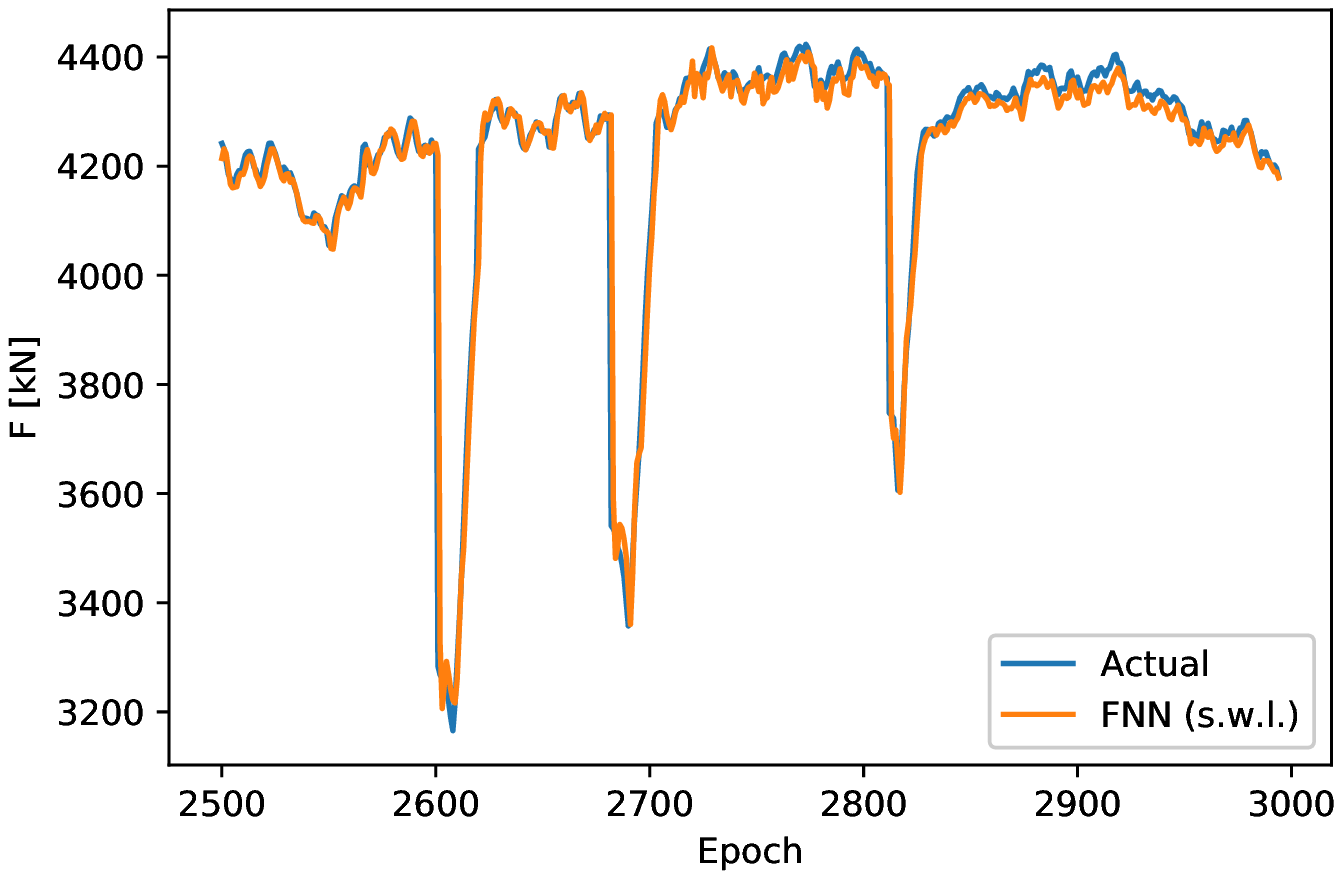}
\end{minipage}
}
\subfigure[FNN (m.w/o.l.)]{
\begin{minipage}{0.22\textwidth}
\includegraphics[width=1.2\textwidth]{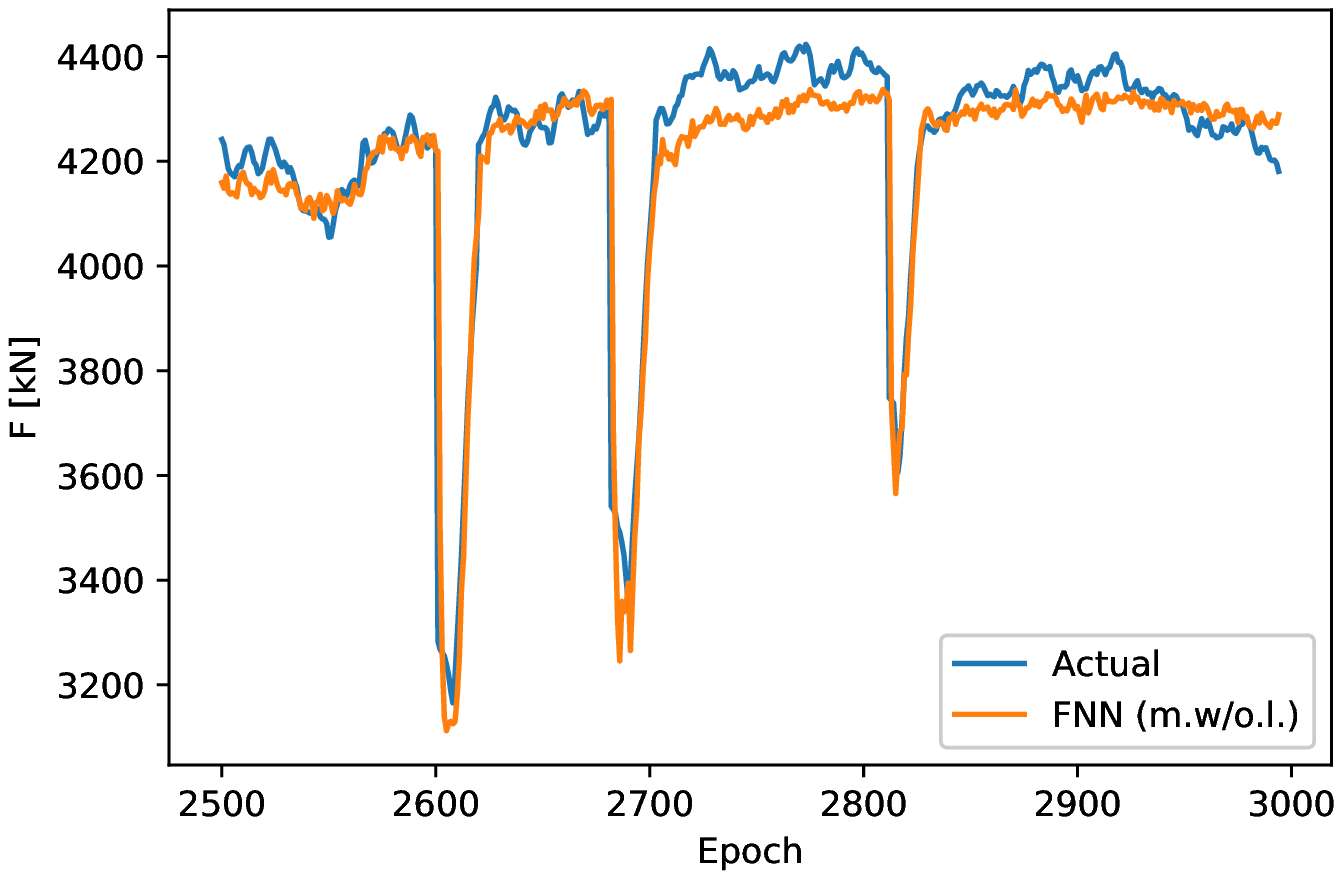}
\end{minipage}
}
\subfigure[FNN (m.w.l.)]{
\begin{minipage}{0.22\textwidth}
\includegraphics[width=1.2\textwidth]{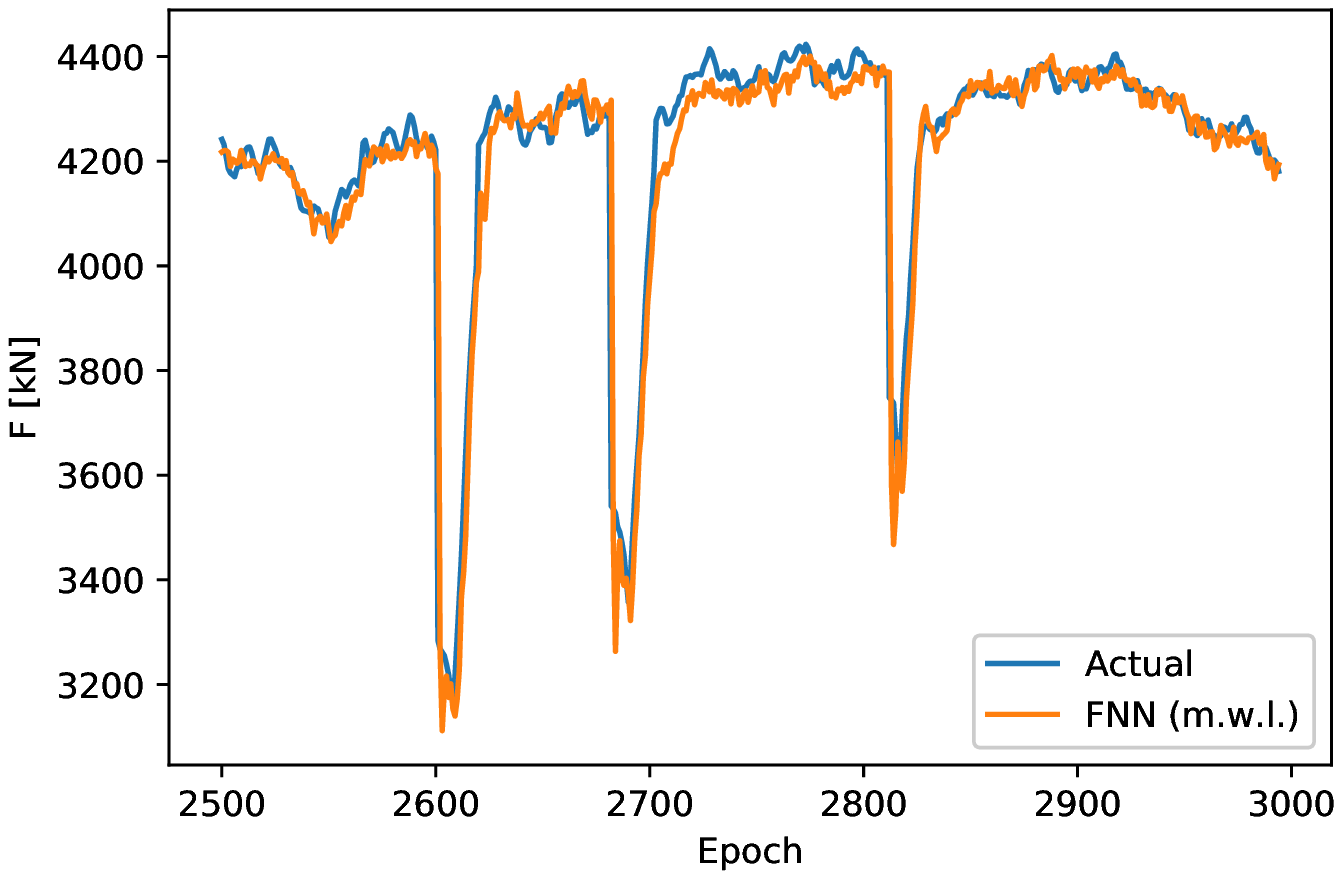}
\end{minipage}
}

\subfigure[RNN (s.w/o.l.)]{
\begin{minipage}{0.22\textwidth}
\includegraphics[width=1.2\textwidth]{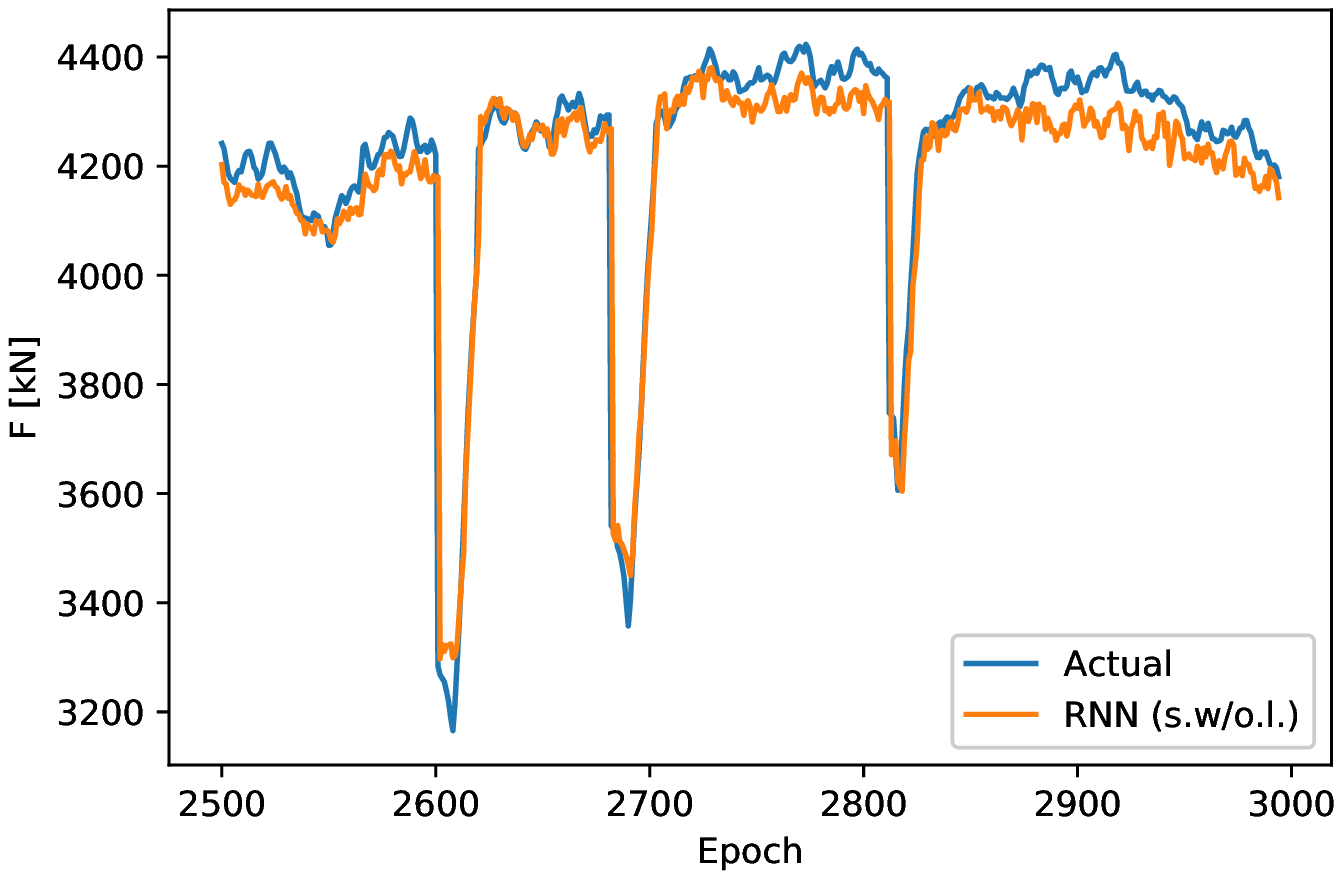}
\end{minipage}
}
\subfigure[RNN (s.w.l.)]{
\begin{minipage}{0.22\textwidth}
\includegraphics[width=1.2\textwidth]{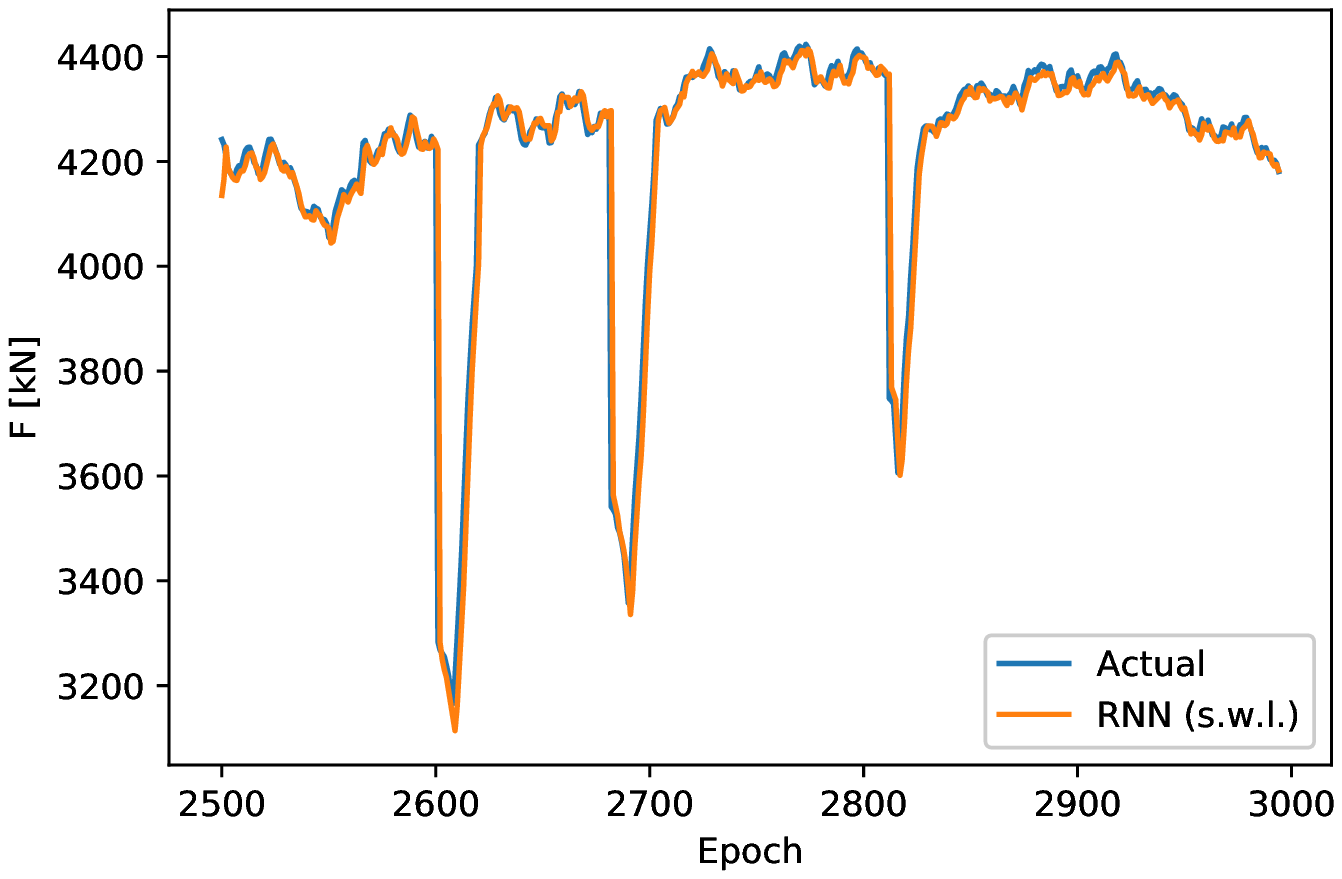}
\end{minipage}
}
\subfigure[RNN (m.w/o.l.)]{
\begin{minipage}{0.22\textwidth}
\includegraphics[width=1.2\textwidth]{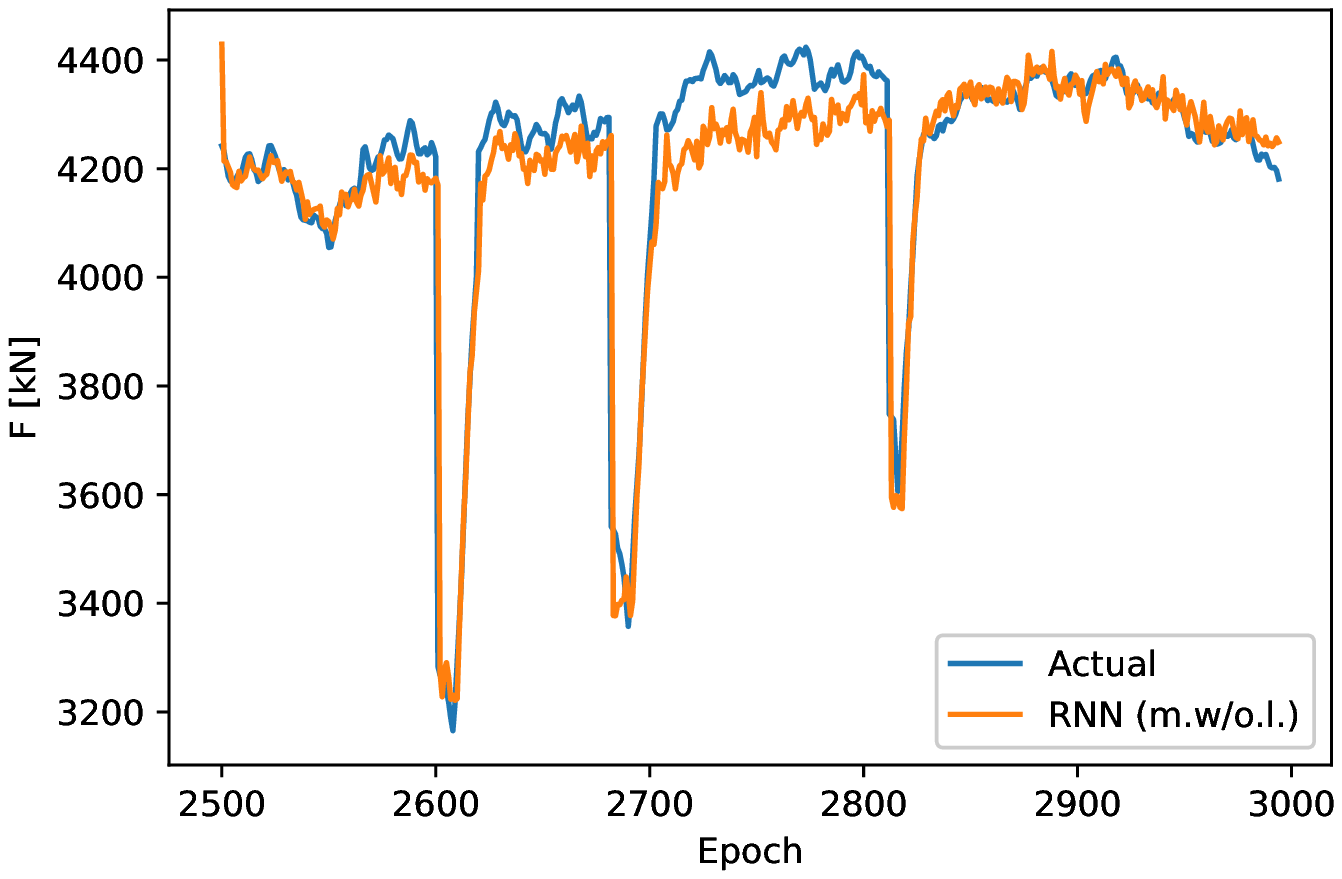}
\end{minipage}
}
\subfigure[RNN (m.w.l.)]{
\begin{minipage}{0.22\textwidth}
\includegraphics[width=1.2\textwidth]{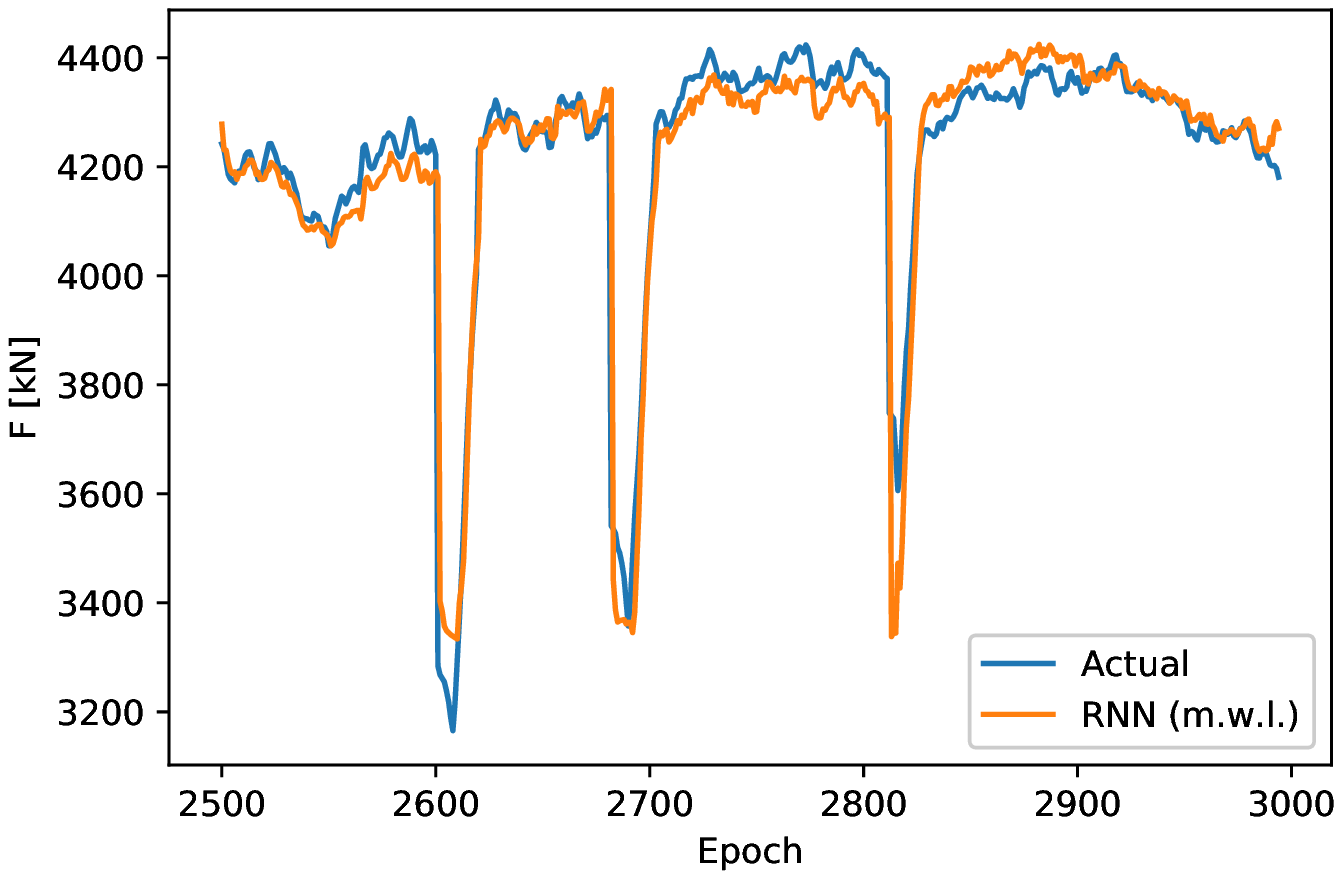}
\end{minipage}
}

\subfigure[GRU (s.w/o.l.)]{
\begin{minipage}{0.22\textwidth}
\includegraphics[width=1.2\textwidth]{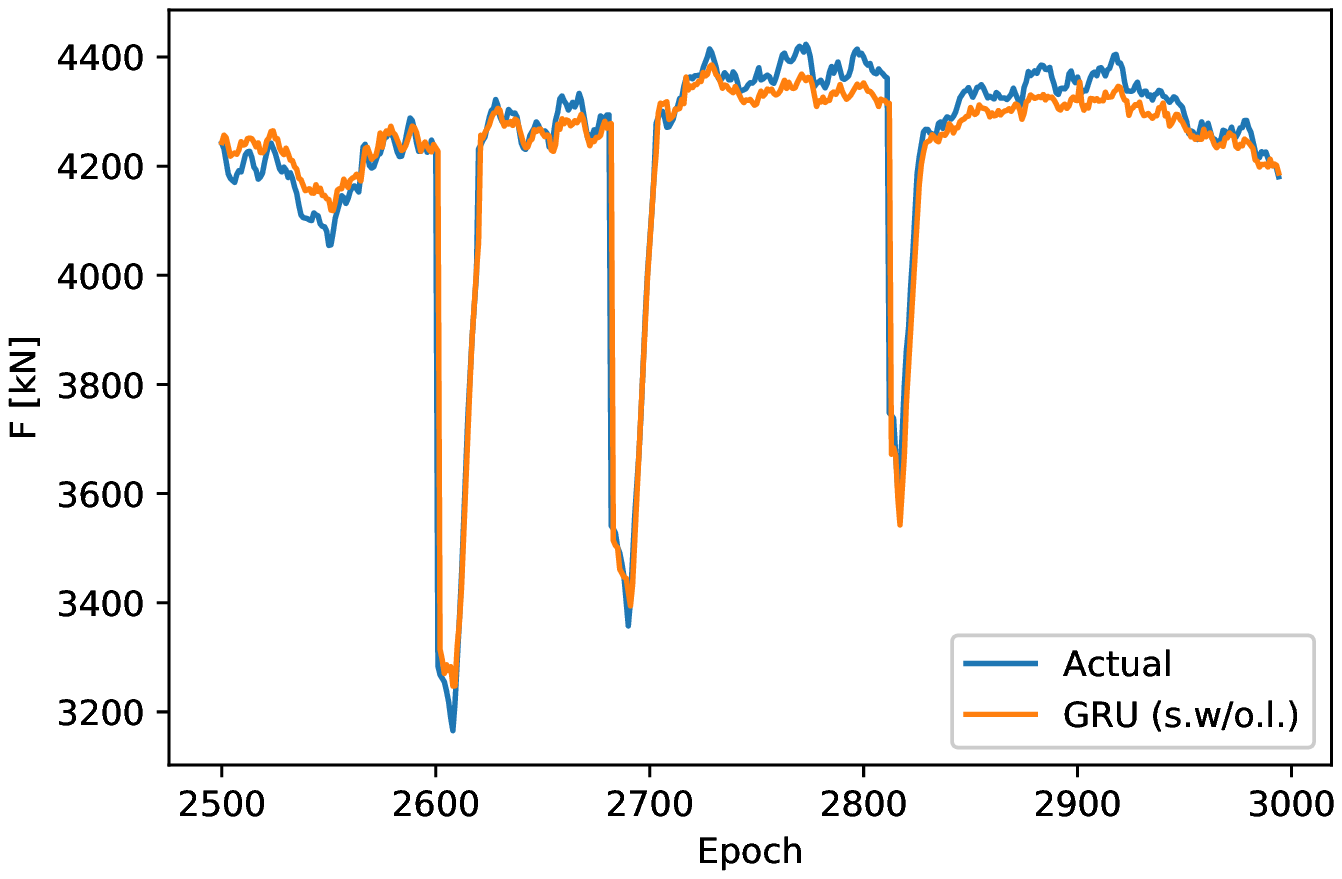}
\end{minipage}
}
\subfigure[GRU (s.w.l.)]{
\begin{minipage}{0.22\textwidth}
\includegraphics[width=1.2\textwidth]{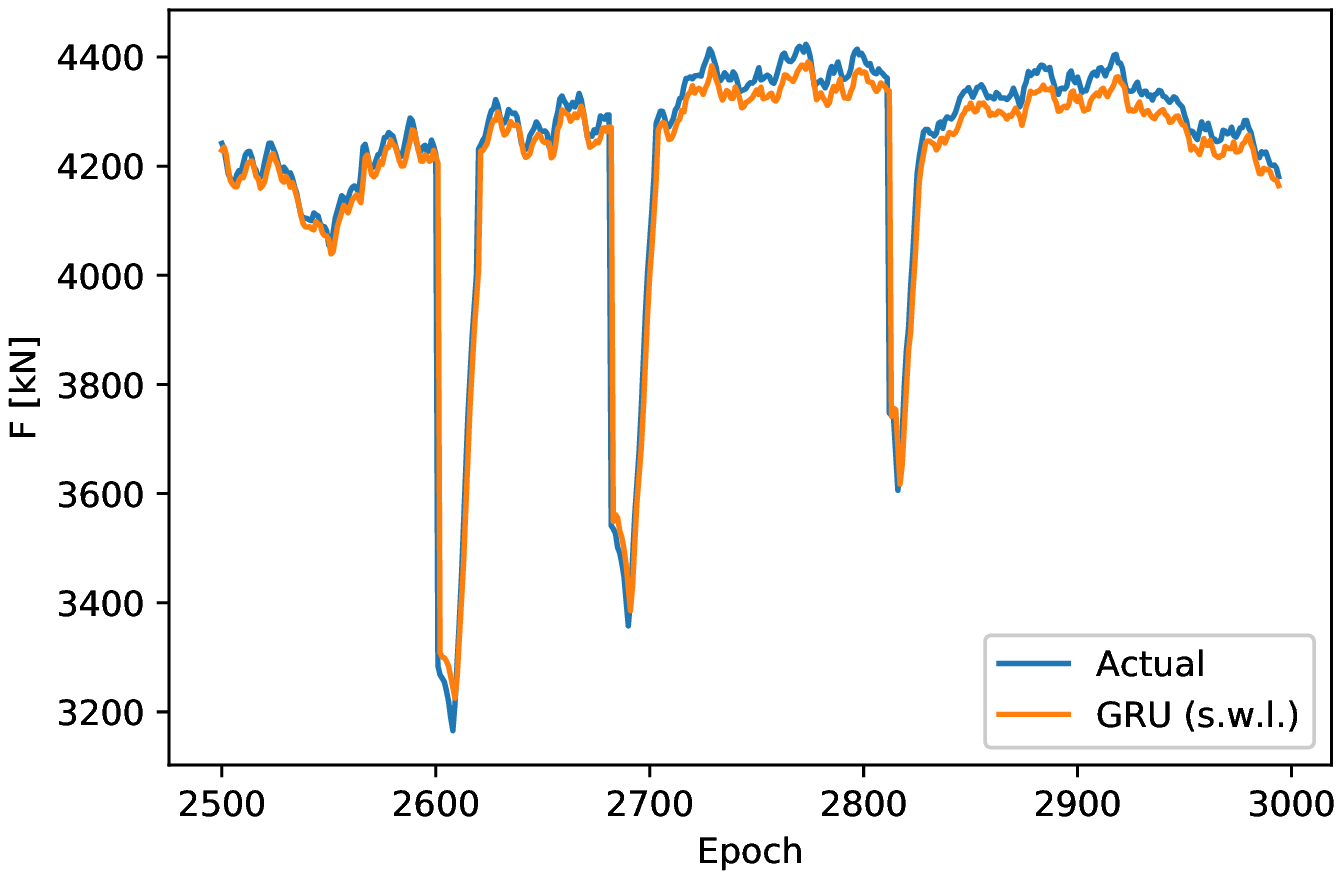}
\end{minipage}
}
\subfigure[GRU (m.w/o.l.)]{
\begin{minipage}{0.22\textwidth}
\includegraphics[width=1.2\textwidth]{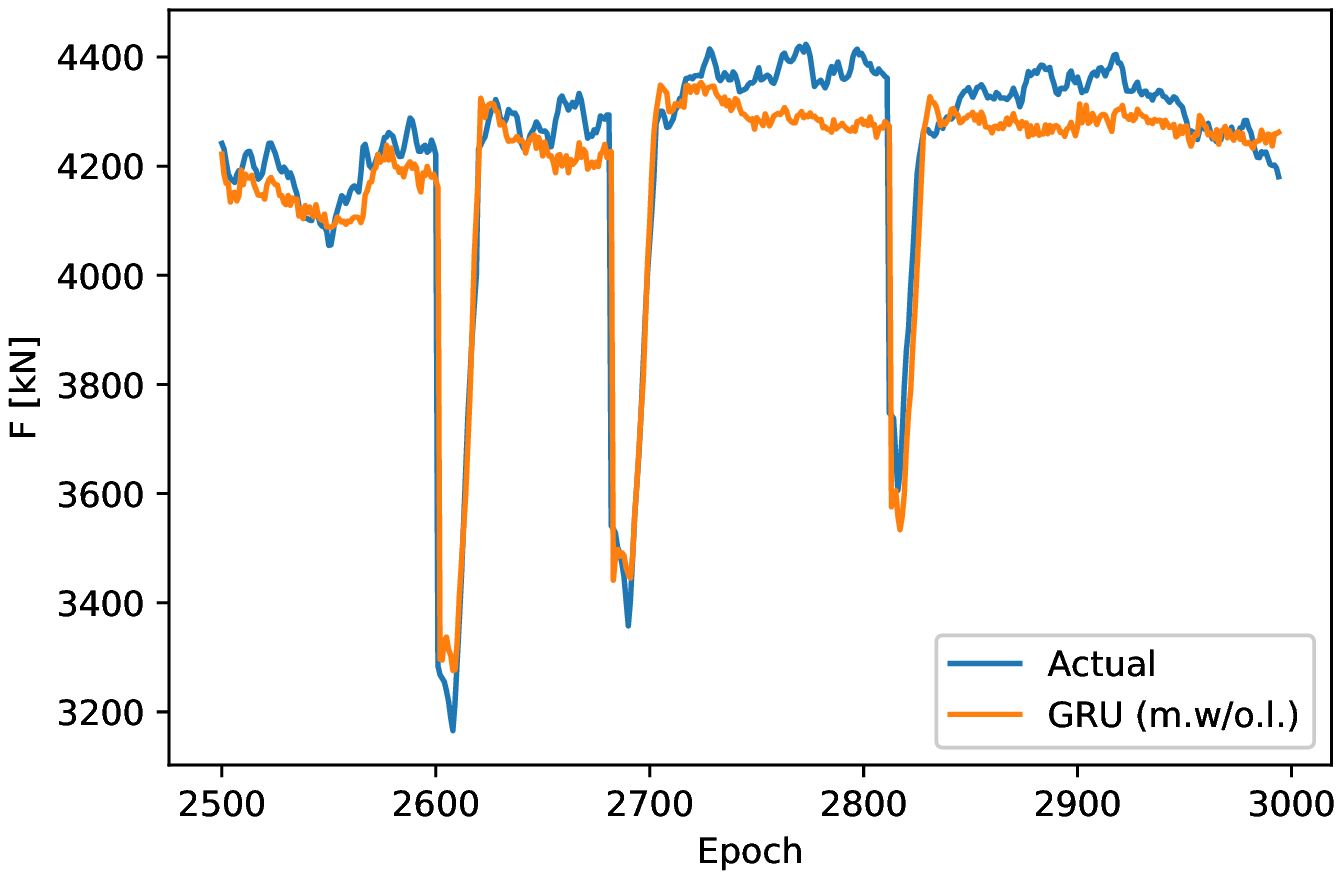}
\end{minipage}
}
\subfigure[GRU (m.w.l.)]{
\begin{minipage}{0.22\textwidth}
\includegraphics[width=1.2\textwidth]{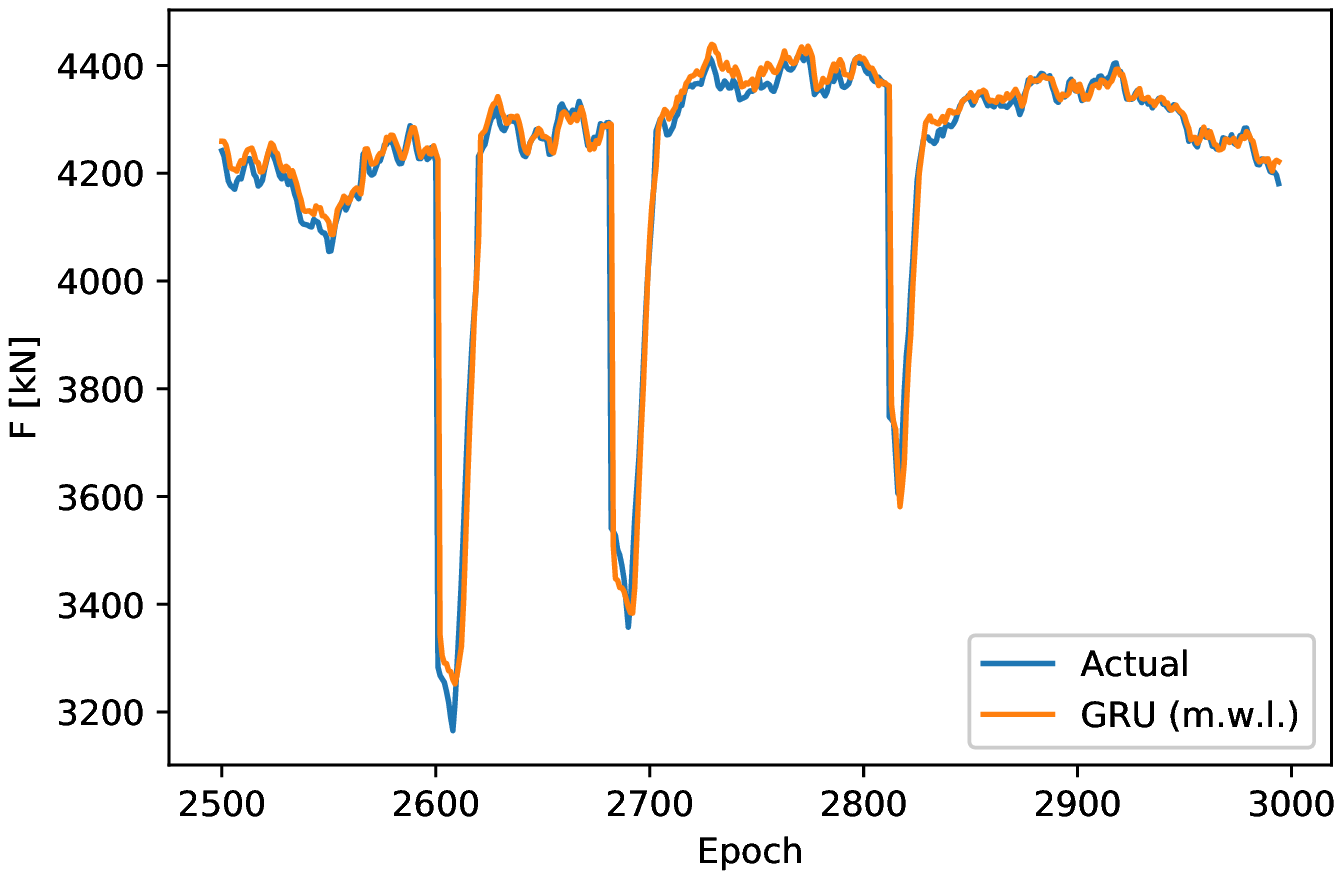}
\end{minipage}
}

\subfigure[LSTM (s.w/o.l.)]{
\begin{minipage}{0.22\textwidth}
\includegraphics[width=1.2\textwidth]{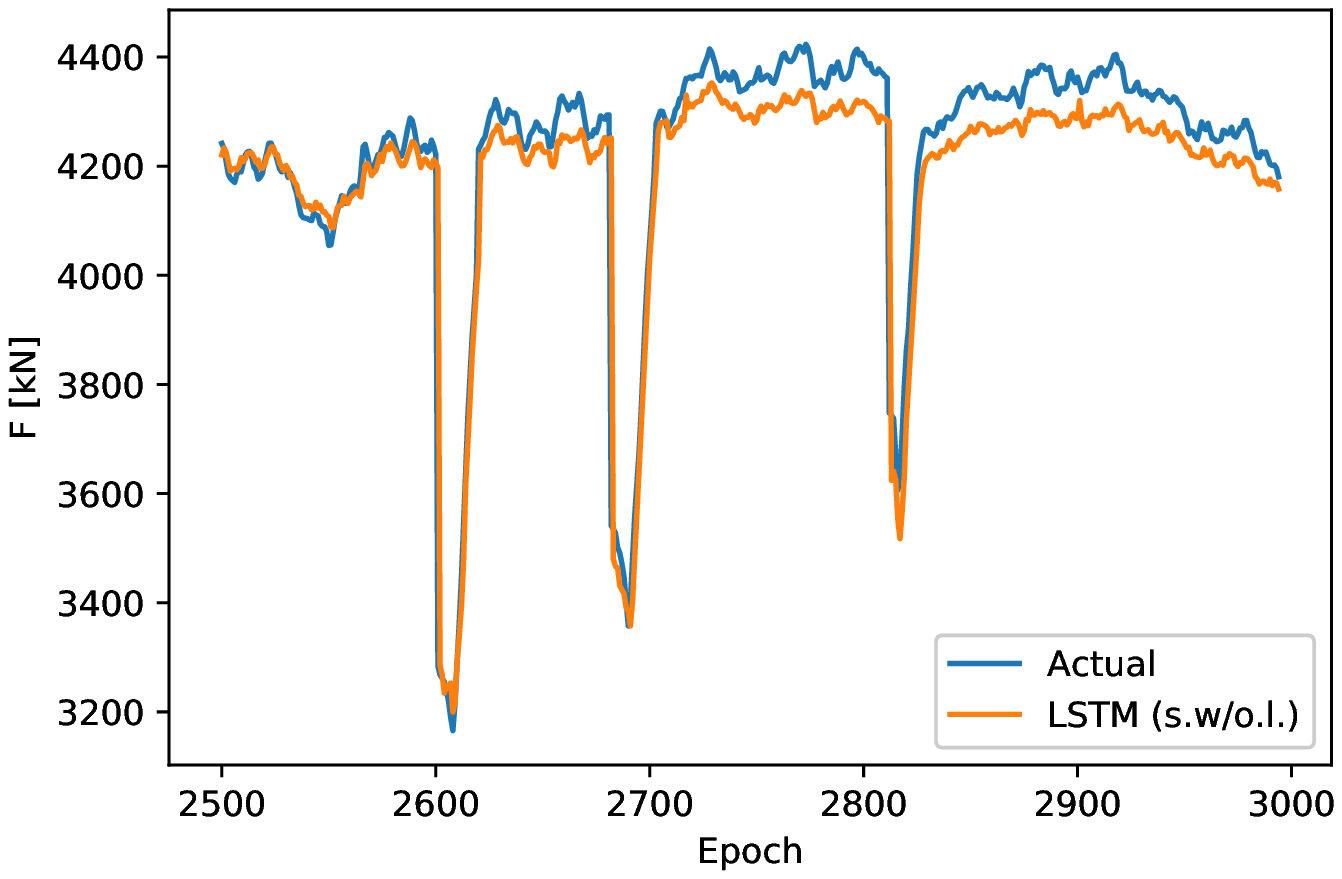}
\end{minipage}
}
\subfigure[LSTM (s.w.l.)]{
\begin{minipage}{0.22\textwidth}
\includegraphics[width=1.2\textwidth]{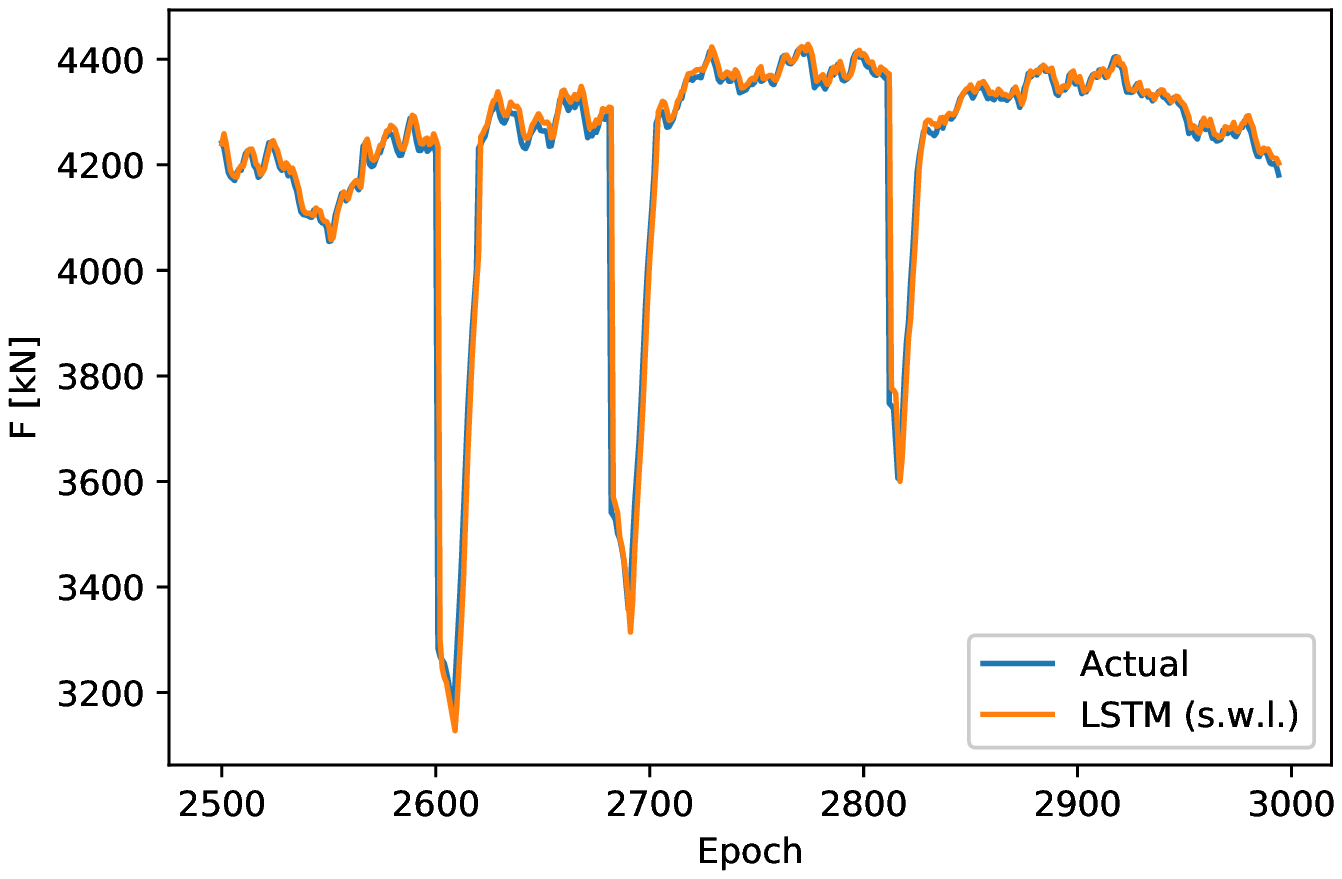}
\end{minipage}
}
\subfigure[LSTM (m.w/o.l.)]{
\begin{minipage}{0.22\textwidth}
\includegraphics[width=1.2\textwidth]{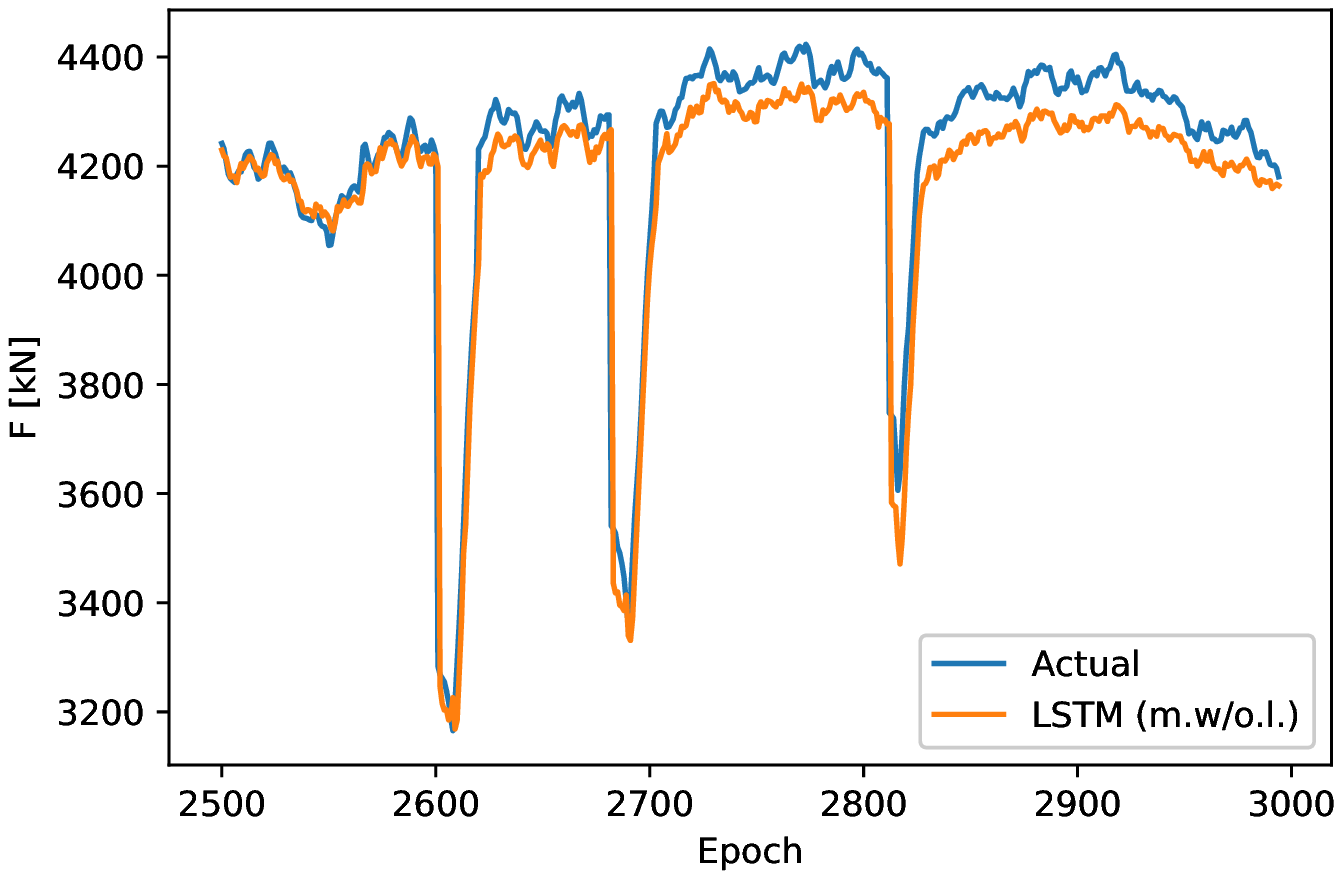}
\end{minipage}
}
\subfigure[LSTM (m.w.l.)]{
\begin{minipage}{0.22\textwidth}
\includegraphics[width=1.2\textwidth]{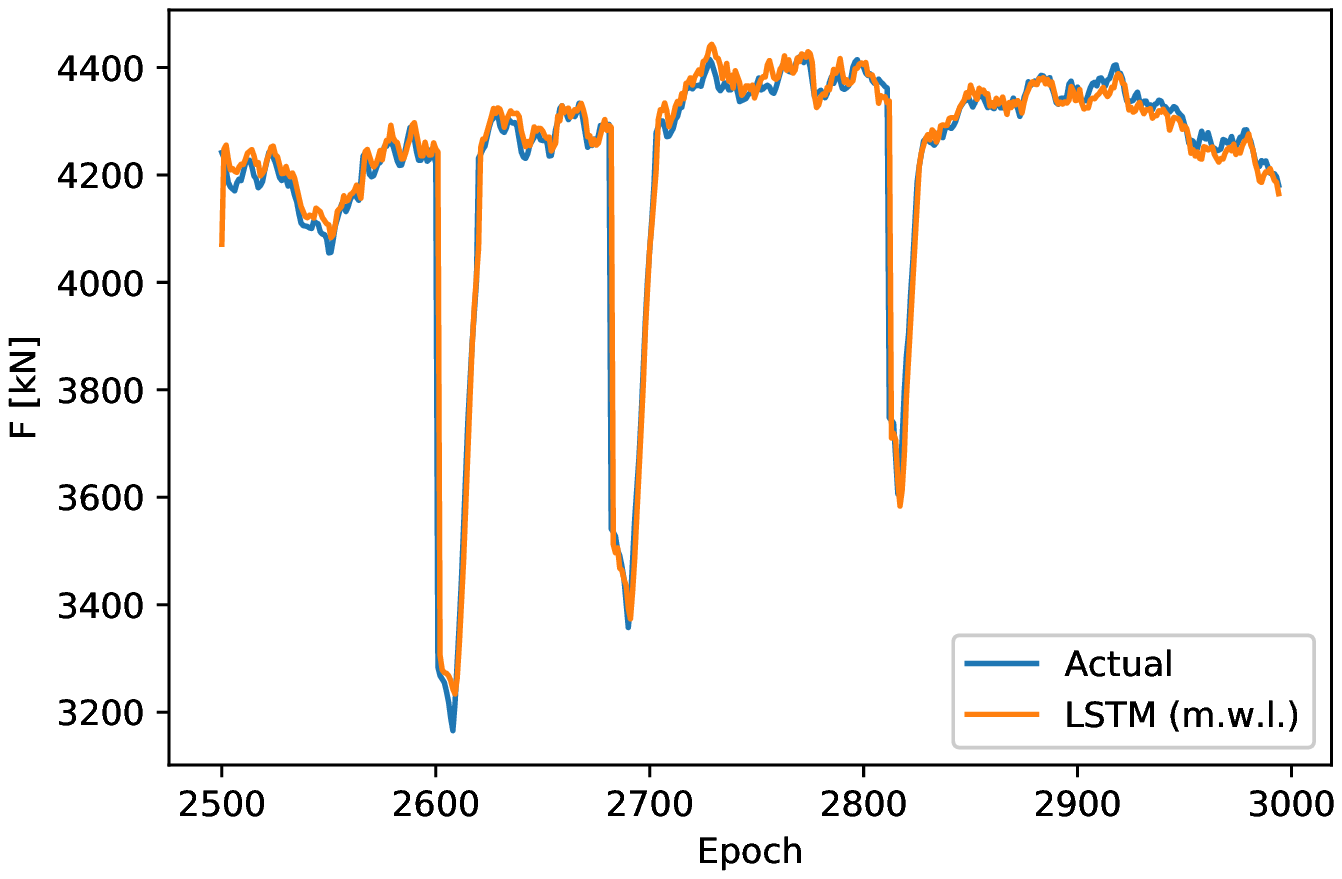}
\end{minipage}
}
\caption{Forecast Results of ``Thrust"} \label{fig:thrust}
\end{figure}

As shown in Fig. \ref{fig:torque} -  Fig. \ref{fig:thrust}, the deep-learning models (RNN, LSTM and GRU) outperform the shallow-learning models (SVR, RF and FNN) for the forecast task in most cases, and these deep models can accurately capture the dynamic variation of the three parameters ``Torque", ``Advance Rate" and ``Thrust" during the next period. Especially, in most cases, GRU performs better than RNN and LSTM, which is because the relatively simpler gate structure is enough to afford the forecast task and meanwhile avoids the overfitting.


As listed in Tab. \ref{tab:results}, the performance of single-output models is better than that of multiple-output models in most cases because the former has a simpler structure that leads to a higher training efficiency. However, to achieve the forecast of the three
parameters, we need to build three single-output models respectively. In contrast, one multiple-output model can simultaneously provide the forecast results of the three TBM parameters and thus has a higher operating efficiency because of a smaller amount of model parameters.

\begin{table}[htbp]
\centering
\caption{Results of Lasso-based Feature Extraction}
{\scriptsize \begin{tabular}{|@{}r |c|c|c|}
\hline
\diagbox[width=25em,trim=l]{\ Features}{\ Coefficient}{Parameters}&Torque & Advance Rate& Thrust\\
\hline
Rotation speed of cutter (r/min)&29.912 &4.178 &\\
\hline
Cutter power (kw) &28.871& &\\
\hline
Pressure of chamber at top left (bar)&-9.126& &\\
\hline
Pressure of chamber at bottom right (bar)&1.451& &\\
\hline
Temperature of oil tank ($^{\circ}$C)&0.813& &\\
\hline
Pressure of shield tail seal at left front (bar) & &48.521&\\
\hline
Rolling angle ($^{\circ}$)& &-29.760&\\
\hline
Propelling pressure (bar)& &0.075&\\
\hline
Propelling pressure of C group (bar)& &0.041&30.267\\
\hline
Pressure of screw pump (bar)& &0.023&\\
\hline
Pressure of shield tail seal at top left back (bar)& & &36.896\\
\hline
Propelling pressure of A group (bar)& & &29.833\\
\hline
Propelling pressure of B group (bar)& & &26.638\\
\hline
Propelling pressure of D group (bar)& & &5.219\\
\hline
Displacement of A group of thrust cylinders (mm)& & &-0.001\\
\hline
\end{tabular}}\label{tab:lasso}
\end{table}

In general, compared with the single-output models, the performance loss of the multiple-output models is due to their complex structures. To overcome this limitation, we use the Lasso-based feature extraction to eliminate the redundant features from the inputs so as to decrease the model complexity.
The effectiveness of the Lasso-based feature extraction can be measured by the quantity called ``Performance Gain" (or briefly denoted as ``PERF Gain"):
\begin{equation*}
\mathop{\mbox{PERF Gain}}\limits_{\mbox{(s.w/o.l. vs. s.w.l.)}} = \frac{\mathop{\mbox{RMSE}}_{\mbox{s.w/o.l.}} - \mathop{\mbox{RMSE}}_{\mbox{s.w.l.}} } { \mathop{\mbox{RMSE}}_{\mbox{s.w/o.l.}}  };
\end{equation*}
and
\begin{equation*}
\mathop{\mbox{PERF Gain}}\limits_{\mbox{(m.w/o.l. vs. m.w.l.)}} = \frac{\mathop{\mbox{RMSE}}_{\mbox{m.w/o.l.}} - \mathop{\mbox{RMSE}}_{\mbox{m.w.l.}} } { \mathop{\mbox{RMSE}}_{\mbox{m.w/o.l.}}  },
\end{equation*}
where ``RMSE" can also be changed to ``MAPE" accordingly. In Tab. \ref{tab:lasso}, we have listed the three groups of features whose Lasso coefficients have the large magnitude and deem that they are strongly related with the forecast task. Meanwhile, we omit the features with the coefficients close to {\it zero}. We find that the essential features of the three TBM operation parameters differ from each other except for two individual features. This fact suggests that the three parameters have difference inherent characteristics. The experimental results given in Tab. \ref{tab:results} point out that the introduction of Lasso-based feature extraction brings a significant performance gain in most cases, and thus support the validity of this method.

\begin{table}[htbp]\caption{RMSE (MAPE) of Models for Real-time TBM Multi-parameters Forecast}
\centering
\resizebox{\textwidth}{!}{
  \begin{tabular}{|c|@{}c||c|c|c||c|c|c|}
  \hline
    & \diagbox[width=10em,trim=l]{\ Models}{\ Results}{Settings} &s.w/o.l.& s.w.l.&$\mathop{\mbox{PERF Gain}}\limits_{\mbox{(s.w/o.l. vs. s.w.l.)}}$&m.w/o.l.& m.w.l.&$\mathop{\mbox{PERF Gain}}\limits_{\mbox{(m.w/o.l. vs. m.w.l.)}}$\\
  \hline
  \multirow{6}{*}{\begin{sideways}Torque\end{sideways}} &SVR&98.204(13.541\%)&
    \cellcolor{gray}{36.510(3.999\%)}&62.882\%(70.467\%)&98.204(13.541\%)& 66.564(8.976\%)&32.219\%(33.712\%)\\
  \cline{2-8}
   &RF &94.795(8.471\%)&30.499(3.221\%)&67.826\%(61.976\%)&77.331(7.959\%)&\cellcolor{gray}{{\bf 29.224}({\bf 2.972\%})}&62.209\%(62.659\%)\\
  \cline{2-8}
   & FNN &41.857(5.144\%)&\cellcolor{gray}{31.265(2.919\%)}&25.305\%(43.254\%)&93.226(11.672\%)&49.240
   (6.362\%)&47.182\%(45.493\%)\\
  \cline{2-8}
  & RNN &40.719(4.555\%)&\cellcolor{gray}{28.472({\bf 2.281\%})}&30.077\%(49.923\%)&61.797(8.229\%)&56.020(7.283\%)&9.348\%(11.496\%)\\
  \cline{2-8}
  & GRU &{\bf 35.476}({\bf 3.974\%})&\cellcolor{gray}{{\bf 28.112}(2.593\%)}& 20.758\%(34.751\%)&64.407(8.596\%)& 55.912(7.328\%)&13.190\%(14.751\%)\\
  \cline{2-8}
  & LSTM &38.239(4.266\%)&\cellcolor{gray}{29.143(2.821\%)}&23.787\%(33.872\%)&{\bf 39.688}({\bf 4.710\%})&54.931(7.517\%)&-38.407\%(-59.597\%)\\
  \hline
  \hline
  \multirow{6}{*}{\begin{sideways}Advance Rate\end{sideways}}
 & SVR&3.613(54.953\%)&\cellcolor{gray}{2.512(38.643\%)}&30.473\%(29.680\%)&3.613(54.953\%)&2.687
    (43.608\%)&25.630\%(20.645\%)\\
  \cline{2-8}
   & RF &2.215({\bf 28.719\%})&\cellcolor{gray}{2.149(27.627\%)}&2.980\%(3.802\%)&2.405({\bf 29.062\%})&2.404(29.299\%)&0.042\%(-0.815\%)\\
  \cline{2-8}
  & FNN &2.586(34.065\%)&\cellcolor{gray}{1.861(25.646\%)}&28.036\%(24.715\%)&{2.424}(29.629\%)&{\bf 1.943}(27.001\%)&19.843\%(8.870\%)\\
  \cline{2-8}
   & RNN &2.367(31.959\%)&\cellcolor{gray}{1.786({\bf 25.972\%})}&24.546\%(18.733\%)&{\bf 2.121}(36.967\%)&1.971({\bf 28.685\%})&7.072\%(22.404\%)\\
  \cline{2-8}
   & GRU &{\bf 2.124}(29.405\%)&\cellcolor{gray}{{\bf 1.749}(28.723\%)}&17.655\%(2.319\%)&2.173(31.787\%)&2.137(29.780\%)&1.657\%(6.314\%)\\
  \cline{2-8}
   & LSTM &2.338(31.997\%)&\cellcolor{gray}{1.775(28.111\%)}&24.080\%(12.145\%)&2.152
   (30.102\%)&2.189(31.483\%)&-1.719\%(-4.588\%)\\
  \hline
  \hline
  \multirow{6}{*}{\begin{sideways}Thrust\end{sideways}}
  &SVR&221.442(4.956\%)&{190.923(4.294\%)}&13.782\%(13.358\%)&221.442(4.956\%)&\cellcolor{gray}{ 145.007}({3.109\%})&34.517\%(37.268\%)\\
  \cline{2-8}
   &  RF &\cellcolor{gray}{85.340(0.956\%)}&99.354(1.434\%)&-16.421\%(-50.000\%)&110.099
   (1.137\%)&125.190(1.185\%)&-13.707\%(-4.222\%)\\
  \cline{2-8}
   &  FNN &132.889(2.854\%)&\cellcolor{gray}{{\bf 65.777}({0.590\%})}&50.502\%({79.327\%})&83.898(1.262\%)&76.749(0.928\%)&8.521\%(26.466\%)\\
  \cline{2-8}
   & RNN &78.607(1.187\%)&\cellcolor{gray}{68.987}({0.588\%})&12.238\%(50.463\%)& {\bf 83.702}({\bf 1.232\%})&82.603(1.101\%)&1.313\%(10.633\%)\\
  \cline{2-8}
   & GRU &{\bf 69.903}({\bf 0.844\%})&69.119(0.855\%)&1.122\%(-1.303\%)&86.797(1.428\%)&\cellcolor{gray}{{67.414}
   ({0.621}\%)}&22.331\%(56.513\%)\\
  \cline{2-8}
   & LSTM &80.748(1.302\%)&\cellcolor{gray}{67.223({\bf 0.546\%})}&16.750\%(58.065\%)&86.724(1.458\%)&{\bf 66.159}({\bf 0.599\%})&23.713\%(58.916\%)\\
  \hline
  \end{tabular}\label{tab:results}}
\end{table}


\section{Conclusion}

Although TBMs have been widely used in various tunnel construction projects, there still remain some challenges in their machinery design and construction safety. Many research interests are inspired to explore the TBMs' risk evaluation, fault diagnostics and prognostics. One key to these problems is the real-time forecast of the TBM load parameters (including torque, advance rate and thrust), which requires that the forecast models should instantaneously provide the accurate future values of these parameters during the next period as long as the current operational data are collected. There have been many works on the forecast of TBM operational parameters, while it is still challenging to realize a true sense of ``real-time" forecast.

In this paper, we use the ML methods to develop the models for the real-time forecast of the TBM load parameters. In particular, we adopt the shallow-learning models (including SVR, RF and FNN) and the deep-learning models (RNN, LSTM, GRU) as the forecast models, respectively. Different from the non-sequential learning problems which require that models should provide the estimate of the current real value when the current data are collected, the inputs of these real-time forecast models are a length of time series data containing the previous and the current data, and their outputs are the estimate of the future real values of the load parameters. Because of the loop structure, we find that these deep models outperform the shallow models in most case.

Moreover, we consider the forecast task in two kinds of modes: single-output, which allows the models generate the forecast result of one parameter each a time, and multiple-output, which requires that the models should simultaneously provide the forecast results of the three parameters. Although the experimental results show that the single-output models perform better than multiple-output models,
three individual single-output models should be develop for the three TBM load parameters. In contrast, one multiple-output model can simultaneously provide the forecast of the three parameters, and thus it has a higher operating efficiency.

In addition, we also use the least absolute shrinkage and selection (Lasso) method to extract essential features of the three parameters according to the magnitude of their Lasso coefficients. Based on the extracted features, we then develop the real-time forecast models for the TBM load parameters. The experimental results show that this manner brings a significant performance gain in most cases. In our future works, we will consider the application of the proposed ML-based models in TBM health monitoring and fault prognostics.

\appendix

\section{Important Features in TBM Operational Data}

In the appendix, we list some important features appearing in the in-situ TBM operational data.

\begin{table}[htbp]\caption{Some Features in TBM Operation Data \cite{gao2019recurrent}}
\centering
\makebox[\linewidth][c]{
\resizebox{\textwidth}{!}{ \begin{tabular}{| r | r |}
\hline
Parameter (Unit) & Parameter (Unit)\\
\hline
Temperature of oil tank ($^{\circ}$C)& Temperature of gear oil ($^{\circ}$C)\\
Rotation speed of cutter(r/min)& Cutter power (kw)\\
Propelling pressure (bar)& Propelling pressure of A group (bar)\\
Propelling pressure of B group (bar)& Propelling pressure of C group (bar)\\
Propelling pressure of D group (bar)& Pressure of equipment bridge (bar)\\
Pressure of articulation system (bar)& Pressure of Shield tail seal at top right front (bar)\\
Pressure of Shield tail seal at right front (bar)& Pressure of Shield tail seal at bottom left front (bar)\\
Pressure of Shield tail seal at top right back (bar)& Pressure of Shield tail seal at right back (bar)\\
Pressure of Shield tail seal at bottom left back (bar)& Pressure of Shield tail seal at left front (bar)\\
Pressure of Shield tail seal at top left front (bar)& Pressure of Shield tail seal at left back (bar)\\
Pressure of Shield tail seal at top left back (bar)& Pressure of Shield tail seal at bottom right back (bar)\\
Rolling angle ($^{\circ}$)& Pressure of screw pump at back (bar)\\
Pressure of chamber at top left (bar)& Pressure of chamber at bottom left (bar)\\
Pressure of chamber at bottom right (bar)& Bentonite pressure (bar)\\
Temperature of screw conveyor ($^{\circ}$C)& Pitch angle ($^{\circ}$)\\
Thrust of cutterhead (kN)& Advance rate (mm/min)\\
Torque of cutterhead (kNm)& Displacement of A group of thrust cylinders (mm)\\
Displacement of B group of thrust cylinders (mm)& Displacement of C group of thrust cylinders (mm)\\
Displacement of D group of thrust cylinders (mm)& Displacement of articulated system at top right (mm)\\
Displacement of articulated system at bottom left (mm)& Displacement of articulated system at top left (mm)\\
Displacement of articulated system at bottom right (mm)& Bentonite pressure of shield shell (bar)\\
Pressure of screw conveyor at front (bar)& Pressure of screw pump (bar)\\
\hline
\end{tabular}}}\label{tab:list}
\end{table}



\newpage

\section*{Acknowledgment}
This work is supported by the National Key R\&D Program of China (Grant No. 2018YFB1700704 \& 2018YFB1702502) and the National Natural Science Foundation of China (Grant No. 11401076 \& 61473328).


\section*{}
\newpage

\newpage

\newpage

\newpage

\newpage

\end{document}